\PassOptionsToPackage{table}{xcolor}
\documentclass[10pt,twocolumn,letterpaper]{article}
\usepackage{amsmath,amsfonts}
\usepackage{algorithmic}
\usepackage{algorithm}
\usepackage{array}
\usepackage[caption=false,font=normalsize,labelfont=sf,textfont=sf]{subfig}
\usepackage{textcomp}
\usepackage{stfloats}
\usepackage{graphicx}
\usepackage{cite}

 \usepackage[pagenumbers]{cvpr} 

\usepackage{tikz,pgfplots}
\usepackage{color}
\usepackage{amssymb}
\definecolor{voxelorange}{RGB}{255,109,4}
\definecolor{voxelorangedark}{RGB}{127.5,54.5,2}
\definecolor{orangegray}{RGB}{237.5,164.5,112}
\definecolor{iccvblue}{rgb}{0.21,0.49,0.74}
\usepackage[pagebackref,breaklinks,colorlinks,allcolors=voxelorangedark]{hyperref}

\usepackage{flushend}

\usepackage{amsmath,amsfonts,bm}

\def\eqref#1{equation~\ref{#1}}

\def\1{\bm{1}}

\def\rvx{{\mathbf{x}}}

\def\vb{{\bm{b}}}

\def\vy{{\bm{y}}}

\DeclareMathAlphabet{\mathsfit}{\encodingdefault}{\sfdefault}{m}{sl}
\SetMathAlphabet{\mathsfit}{bold}{\encodingdefault}{\sfdefault}{bx}{n}

\def\sN{{\mathbb{N}}}

\def\sR{{\mathbb{R}}}
\def\sS{{\mathbb{S}}}
\def\sT{{\mathbb{T}}}

\newcommand{\E}{\mathbb{E}}

\DeclareMathOperator*{\argmax}{arg\,max}
\DeclareMathOperator*{\argmin}{arg\,min}

\definecolor{tableheader}{RGB}{228,224,221} 
\definecolor{tablered}{RGB}{255,230,230}
\definecolor{tablegreen}{RGB}{230,240,230}

\pdfoutput=1

\begin{document}

\title{Auto-Labeling Data for Object Detection}

\author{Brent A. Griffin$^\text{\scriptsize 1}$  \qquad Manushree Gangwar$^\text{\scriptsize 1}$ \qquad Jacob Sela$^\text{\scriptsize 1}$ \qquad Jason J. Corso$^\text{\scriptsize 1,2}$ \\
	$^\text{\scriptsize 1~}$Voxel51 \qquad $^\text{\scriptsize 2~}$University of Michigan \\
	{\tt\small \{brent,manushree,jacob,jason\}@voxel51.com}
}
\maketitle

\begin{abstract}

Great labels make great models.
However, traditional labeling approaches for tasks like object detection have substantial costs at scale.
Furthermore, alternatives to fully-supervised object detection either lose functionality or require larger models with prohibitive computational costs for inference at scale.
To that end, this paper addresses the problem of training standard object detection models without \textit{any} ground truth labels.
Instead, we configure previously-trained vision-language foundation models to generate application-specific pseudo ``ground truth" labels.
These auto-generated labels directly integrate with existing model training frameworks, and we subsequently train lightweight detection models that are computationally efficient.
In this way, we avoid the costs of traditional labeling, leverage the knowledge of vision-language models, and keep the efficiency of lightweight models for practical application.
We perform exhaustive experiments across multiple labeling configurations, downstream inference models, and datasets to establish best practices and set an extensive auto-labeling benchmark. 
From our results, we find that our approach is a viable alternative to standard labeling in that it maintains competitive performance on multiple datasets and substantially reduces labeling time and costs.
	
\end{abstract}

\section{Introduction}

Since the introduction of modern deep learning \cite{alexnet}, groundbreaking visual AI models like ViT and DALL-E increasingly rely on massive datasets for training \cite{dosovitskiy2021an,ramesh2022hierarchical}.
In fact, the computational cost to train a single state-of-the-art deep learning model in various fields doubles every 3.4 months due to increasingly large models and datasets \cite{openai18, Zhao_2023_WACV}.
Similarly, object detection, i.e., predicting bounding boxes and category labels for objects in an RGB image or video frames, has seen remarkable methodological advances \cite{ren2016faster,Zhao_2024_CVPR,carion2020end,NEURIPS2024_c34ddd05} largely thanks to an abundance of annotated training and evaluation data in high-quality datasets \cite{EvEtAl10,coco,Gupta_2019_CVPR,kuznetsova2020open}.
Furthermore, we can improve detection performance for various applications by training on expansive data sources from the web or deployed robot and AV systems \cite{Gr23}.
However, traditional image labeling approaches have substantial costs at scale, with annotation taking anywhere between 7 seconds per bounding box to 1.5 hours for full semantic segmentation \cite{JaGr13,Cordts_2016_CVPR}.

There have been tremendous efforts to mitigate annotation costs to enable data on demand and training at scale.
For object detection, weakly superivsed detectors learn from low-cost but coarse annotations like image-level labels \cite{Bilen_2016_CVPR,seo2022object},
semi-supervised detectors learn on a combination of labeled \textit{and} unlabeled data \cite{Zhang_2023_CVPR,Zhou_2021_CVPR},
and, remarkably, unsupervised detectors learn without \textit{any} labeled data \cite{Wang_2023_CVPR,Ishtiak_2023_CVPR}.
Another approach to achieve training at scale is through the use of a vision-language model (VLM). 
After pre-training on large-scale image-text pair datasets \cite{zhou2020unified}, VLMs have demonstrated success in downstream tasks like image classification \cite{clip}, object detection \cite{gu2022openvocabulary}, and segmentation \cite{shi2024the}.

Our current work is inspired by the success of this previous object detection and VLM work.
However, there are a number of trade offs associated with these approaches that we aim to address.
First, weakly- and semi-supervised methods reduce the cost and amount of annotation needed, but they still require labeled data.
Second, unsupervised object detectors train without labeled data, but they are unable to identify application-specific object classes.
On the other hand, VLMs are pre-trained and can detect numerous classes specified via a text-prompt.
However, due to the large size of the combined general purpose image and langauge models, VLMs are computationally cost prohibitive for many conventional detection applications, e.g., real-time inference on robot hardware or processing massive data.
Furthermore, VLM performance is highly sensitive to changes in configuration and application.

To that end, this paper addresses the problem of training conventional object detection models without \textit{any} ground truth labels.
Instead, we use previously-trained VLMs as foundation models that, given an application-specific text prompt, generate pseudo ground truth labels for previously unlabeled data.
We call this process {\bf A}uto-{\bf L}abeling (AL). 
As we will show, AL is much more time and cost effective than traditional labeling given the correct configuration. 
After auto-labeling a training dataset, we train lightweight detection models that are computationally efficient for conventional detection applications.
In this way, we avoid the time and costs of traditional labeling (e.g., 2.5\textrm{K} hours \& \$46.3\textrm{K} for BDD \cite{Yu_2020_CVPR}), leverage the knowledge of VLMs, and keep the efficiency of lightweight detection models.

\begin{figure*}
	\centering
	\includegraphics[width=0.975\textwidth]{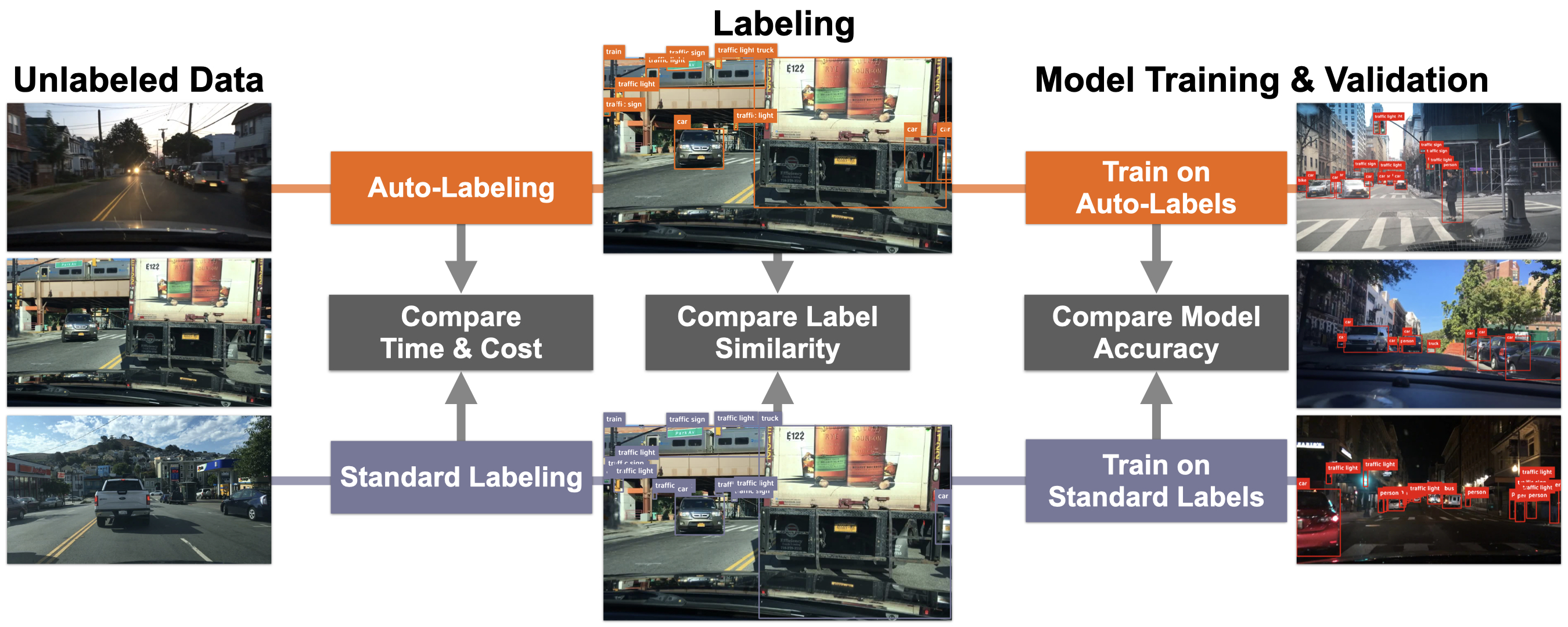}
	\caption{{\bf Auto-Labeling Data for Object Detection Overview}. Visualizations generated using the FiftyOne Library \cite{moore2020fiftyone}.}
\end{figure*}

To our knowledge, the closest existing work to our approach is Nagase \etal \cite{nagase2025annotation}.
In their recent experiments, they use a VLM ensemble of Grounding DINO-T \cite{shilong2024} \& GLIP-Large \cite{Li_2022_CVPR} to label VOC \cite{EvEtAl10} \& COCO \cite{coco}.
They then perform 14 downstream model training experiments and find improved average recall over standard label training on the VOC validation and COCO test sets.
While we celebrate the success of this previous work, we also emphasize that there remain an abundance of open questions in regards to auto-labeling for object detection and beyond.

In our work, we uniquely focus on the selection and configuration of individual vision-language foundation models, which is computationally more efficient than using an ensemble.
We also incorporate more recent advancements in foundation models \cite{Cheng_2024_CVPR,wang2025yoloerealtimeseeing}, and find that we are able to perform auto-labeling 12-300$\times$ faster than with the Grounding DINO-T model and with greater reliability.
We also expand experiments to include more challenging datasets across multiple application domains to better understand the current limits of auto-labeling.
Overall, we conduct experiments across three foundation models with multiple configurations (\cref{tab:fa}), six different downstream inference model architectures (\cref{tab:inf}), and four datasets (\cref{tab:dataset}).
In total, this entails 445 separate model training experiments using 169 unique sets of labels.
From our experiments and subsequent analysis, we explicitly answer the following:
\begin{enumerate}
	\item What costs are associated with auto-labeling? (\cref{sec:cost})
	\item How different is AL from human labels? (\cref{sec:label})
	\item In what scenarios are AL-trained models competitive with those trained on traditional annotation? (\cref{sec:model})
	\item What are best practices for general use? (\cref{sec:bp})
\end{enumerate}
From the results, we find that AL is a viable alternative to standard labeling approaches with substantial time and cost benefits and competitive performance in many scenarios.
Furthermore, this paper provides a practical guide for visual AI developers and an expansive benchmark for future auto-labeling research.

\section{Auto-Labeling Methodology}

\subsection{Problem Formulation}
\label{sec:problem}

Here, we define the problem of {\bf A}uto-{\bf L}abeling (AL) data.
Formally, we are given an \textit{unlabeled} dataset $\sS=\{\rvx_i\}^N_{i=1}$ with $N$ examples drawn i.i.d.~from an underlying distribution $P$, where $\rvx_i$ are the data.
Traditionally, human annotators provide a set of ground truth labels $\vy_i$ for each $\rvx_i$, resulting in labeled dataset $\sS^\text{\tiny L} = \{(\rvx_i, \vy_i)\}^N_{i=1}$, which is subsequently used to train models.
Alternatively, our goal is to reduce annotation time and costs by \textit{automatically} generating pseudo ground truth labels $\vy^{\text{\tiny A}}_i$ for each $\rvx_i$, resulting in AL dataset $\sS^\text{\tiny A} = \{(\rvx_i, \vy^{\text{\tiny A}}_i)\}^N_{i=1}$. 
Notably, we structure $\sS^\text{\tiny A}$ the same as $\sS^\text{\tiny L}$ to enable direct replacement and integration with existing model training frameworks.

We formulate the auto-labeling problem as
\begin{align}
	\argmin_{\sS^\text{A}} \E_{\rvx, \vy \sim P} [ \ell (\rvx, \vy; f_{(\sS^\text{\tiny A})}) ],
	\label{eq:al}
\end{align}
where $\ell$ is the task-specific loss function and $f_{(\sS^\text{\tiny A})}$ is a model trained on $\sS^\text{\tiny A}$.
In plain words, the goal of \cref{eq:al} is to automatically generate labels ($\sS^\text{\tiny A}$) to train a downstream model ($f$) that accurately predicts ground truth labels ($\vy$) for unseen images ($\rvx$) drawn from the expected underlying data distribution ($P$).

\subsection{Auto-Labeling via Foundation Models}
To generate auto-labels $\vy^\text{\tiny A}$, we use existing, off-the-shelf foundation models in a zero-shot manner.
Specifically, we generate object detection labels using
\begin{align}
	f^{\text{\tiny A}}(\rvx_i, \alpha, \sT) := 
	\begin{bmatrix}
		\vb_1, c_1 \\
		\vdots~~~\vdots \\
		\vb_j, c_j \\
		\vdots~~~\vdots \\
		\vb_n, c_n
	\end{bmatrix}
	= \vy^\text{\tiny A}_i,
	\label{eq:fa}
\end{align}
where $f^\text{\tiny A}$ is a previously-trained foundation model; 
$\alpha\in[0,1]$ is a fixed threshold that determines the model confidence required to output object labels; 
$\sT=\{\text{\footnotesize text prompt for class}_i\}^M_{i=1}$ is an ordered set of text prompts that map $M$ class names or descriptions to predicted class indices $c_j\in\sN$;
$\vb_j=~[x_j, y_j, w_j, h_j]\in\sR^4$ is the corresponding bounding box center, width, and height for a predicted object label of class $c_j$ with confidence $>\alpha$;
and $\vy^\text{\tiny A}_i\in\sR^{n\times5}$ is the complete set of object labels for image $\rvx_i$.
Using \cref{eq:fa}, we update our AL formulation from \cref{sec:problem} to explicitly include $f^\text{\tiny A}$ as $\sS^\text{\tiny A} = \{(\rvx_i, f^{\text{\tiny A}}(\rvx_i, \alpha, \sT) )\}^N_{i=1}$.
Notably, $\sT$ is determined on a per-dataset basis, but we test many $f^\text{\tiny A}$, $\alpha$ configurations to understand best AL practices.

\setlength{\tabcolsep}{4pt}
\begin{table}
	\centering
	\caption{ {\bf Foundation Models used for Auto-Labeling}.\\
		All paper experiments and runtimes use an NVIDIA L40S GPU.
	}
	\vspace{-0.75em}
	\scriptsize
	\begin{tabular}{| c | l | r | r | c | c |}
		\hline
		\rowcolor{tableheader} \multicolumn{1}{| c |}{\bf Auto-} & \multicolumn{1}{ c |}{} & \multicolumn{1}{ c |}{\bf Number} & \multicolumn{1}{ c |}{\bf Runtime} & \multicolumn{2}{ c | }{\bf VOC $\text{F}_\text{1}$ Score @}  \\ 
		\rowcolor{tableheader} \multicolumn{1}{| c |}{\bf Labeling} & \multicolumn{1}{ c |}{} & \multicolumn{1}{ c |}{\bf of} & \multicolumn{1}{ c |}{\bf to Label} & \multicolumn{2}{ c | }{\bf Confidence ($\alpha$)}  \\ 
		\rowcolor{tableheader} \multicolumn{1}{| c |}{\bf Model} & \multicolumn{1}{ c |}{\bf Training Datasets} & \multicolumn{1}{ c |}{\bf Params} & \multicolumn{1}{ c |}{\bf VOC (\textrm{s})} & \multicolumn{1}{ c }{\bf ~~~~0.5~~~~} & \multicolumn{1}{ c |}{\bf 0.9} \\ \hline
		YOLOW	&	O365, GQA, Flickr30k	&	72.9	\textrm{M}	&	197.2	&	0.785	&	0.482	\\ \hline
		YOLOE	&	O365, GQA, Flickr30k	&	35.2	\textrm{M}	&	204.9	&	0.761	&	0.433	\\ \hline
		GDINO	&	O365, GoldG, Cap4M	&	172.2	\textrm{M}	&	2,290.3	&	0.759	&	0.034	\\ \hline
	\end{tabular}
	\label{tab:fa}
\end{table}

We evaluate \cref{eq:fa} via precision, recall, and $F_1$ metrics relative to previously annotated data ($\rvx, \vy$) using
\begin{equation}
\begin{aligned}
	\argmax_{f^\text{\tiny A}, \alpha} \E_{\rvx, \vy \sim P} [ F_1 \big(\vy; f^{\text{\tiny A}}(\rvx, \alpha, \sT) \big)], \\
	F_1 = {\footnotesize 2\frac{\text{precision}\cdot\text{recall}}{\text{precision + recall}}} = {\footnotesize \frac{\text{2TP}}{\text{2TP + FP + FN}}},
\end{aligned}
\label{eq:f1}
\end{equation}
where true positives ({\footnotesize TP}) is the number of auto-labels with the correct class label \textit{and} bounding box Intersection over Union $\text{(IoU)}>0.5$ relative to $\vy$, false positives ({\footnotesize FP}) is the number of auto-labels failing the {\footnotesize TP} criteria, and false negatives ({\footnotesize FN}) is the number of $\vy$ labels without a corresponding {\footnotesize TP}.
Notably, $F_1~\in~[0,1]$ is the harmonic mean of precision ({\footnotesize $\frac{\text{TP}}{\text{TP+FP}}~\in~[0,1]$}) and recall ({\footnotesize $\frac{\text{TP}}{\text{TP+FN}} \in [0,1]$}). 
In plain words, {\bf precision} is the frequency of AL being correct, {\bf recall} is the frequency of ground truth objects being correctly auto-labeled, and the {\bf $F_1$ score} emphasizes AL performance across \textit{both} metrics, where an $F_1=1$ indicates perfect precision and recall.
Using \cref{eq:f1}, we directly compare AL to human labels prior to any downstream model training.

We include $f^\text{\tiny A}$ in \cref{eq:f1} to account for choosing the best available foundation model for auto-labeling.
For our experiments, we use the foundation models listed in \cref{tab:fa}.
YOLO-World (YOLOW) \cite{Cheng_2024_CVPR} and YOLOE \cite{wang2025yoloerealtimeseeing} are both pre-trained on Objects365 \cite{Shao_2019_ICCV}, GQA \cite{Hudson_2019_CVPR}, and Flickr30k \cite{young14} and implemented via Ultralytics \cite{jocher2023yolo}.
Grounding DINO-T (GDINO)  \cite{shilong2024} is pre-trained on Objects365, GoldG \cite{Kamath_2021_ICCV}, and Cap4M \cite{Li_2022_CVPR} and implemented via HuggingFace \cite{wolf-etal-2020-transformers}.
Notably, the VOC $F_1$ score dramatically changes with foundation model ($f^\text{\tiny A}$) \textit{and} confidence threshold ($\alpha$).

\subsection{Downstream Inference Model Training}
\label{sec:inf}
Lightweight inference models are commonly used for edge deployments where compute resources or network capabilities are limited.
State-of-the-art inference model performance for application-specific tasks like object detection typically results from supervised learning with annotated labels.
To validate if AL is a viable alternative to traditional annotation, we evaluate AL via downstream model training and subsequent validation performance.
Using \cref{eq:fa} in \cref{eq:al}, we formalize the AL model training task as
\begin{align}
	\argmin_{f^\text{\tiny A}, \alpha} \E_{\rvx, \vy \sim P} [ \ell (\rvx, \vy; f_{(\sS, f^\text{A}, \alpha, \sT)}) ],
	\label{eq:fi}
\end{align}
where downstream inference model $f$ trains on application-specific dataset $\sS$, which is labeled \textit{only} by a foundation model $f^\text{\tiny A}$ with application-specific text prompt $\sT$ and confidence threshold $\alpha$.
Using \cref{eq:fi}, we can compare AL to human labels for actual downstream model training and validation across various applications.
Note that training an inference model $f$ for each $f^\text{\tiny A}, \alpha$ configuration in \cref{eq:fi} requires more time than the label-only evaluation in \cref{eq:f1}.

\setlength{\tabcolsep}{4.5pt}
\begin{table}
	\centering
	\caption{ {\bf Inference Models used for AL-based Training}.
	Baseline performance results from training on human labels.
	Inference runtime uses 80 classes and is averaged over the COCO train set.
	}
	\vspace{-0.75em}
	\scriptsize
	\begin{tabular}{| l | r | c| c | c | c |}
		\hline
		\rowcolor{tableheader} \multicolumn{1}{| c |}{\bf Inference} & \multicolumn{1}{ c |}{\bf \# of} & \multicolumn{1}{ c |}{\bf Inference} & \multicolumn{3}{ c |}{\bf Baseline VOC Validation Performance}  \\ 
		\rowcolor{tableheader} \multicolumn{1}{| c |}{\bf Model} & \multicolumn{1}{ c |}{\bf Params} & \multicolumn{1}{ c |}{\bf Runtime} & \multicolumn{1}{ c }{\bf ~~~mAP50~~~~} & \multicolumn{1}{ c }{\bf ~~~mAP75~~~~} & \multicolumn{1}{ c |}{\bf mAP50-95} \\ \hline
		YOLO11n	&	2.6	\textrm{M}&	0.51 \textrm{ms}	&	0.756	&	0.605	&	0.549	\\ \hline
		YOLO11s	&	9.5	\textrm{M}&	1.03 \textrm{ms} &	0.817	&	0.672	&	0.613	\\ \hline
		YOLO11m	&	20.1 \textrm{M}	&	2.55 \textrm{ms}	&	0.844	&	0.714	&	0.652	\\ \hline
		YOLO11l	&	25.4 \textrm{M}	&	3.13 \textrm{ms}	&	0.855	&	0.736	&	0.673	\\ \hline
		YOLO11x	&	57.0 \textrm{M}	&	5.78 \textrm{ms}	&	0.860	&	0.744	&	0.681	\\ \hline
		RT-DETR	&	33.0 \textrm{M}	&	4.51 \textrm{ms}	&	0.775	&	0.641	&	0.587	\\ \hline
	\end{tabular}
	\label{tab:inf}
\end{table}

For AL model training experiments, we use the inference models listed in \cref{tab:inf}.
We use several YOLO11 variants \cite{yolo11_ultralytics} to understand AL performance across different model sizes as well as transformer-based RT-DETR \cite{Zhao_2024_CVPR} to understand AL performance across different model architectures.
We train all inference models via Ultralytics for 100 epochs without \textit{any} pre-trained weights, ensuring that validation performance is from either pure AL or human labels.

After training, we evaluate inference model performance on the validation set using the \textit{mean} average precision,
$ \text{mAP50} = \frac{1}{M} \sum_{c=1}^{M} \text{AP50}$,
where the \textit{average precision} $\text{AP50}$ is taken over a discretized precision/recall curve with $\text{IoU}>0.5$ for each of the $M$ object classes \cite{EvEtAl10}.

\section{Auto-Labeling Evaluation}
\label{sec:eval}

\subsection{Dataset Selection}
\label{sec:data}

\setlength{\tabcolsep}{3.75pt}
\begin{table}
	\centering
	\caption{{\bf Experiment Datasets}.}
	\vspace{-0.75em}
	\scriptsize
	\begin{tabular}{| c | c | r | r | l | }
		\hline 
\rowcolor{tableheader} & \multicolumn{1}{ c |}{\bf \# of} & \multicolumn{2}{ c |}{\bf \# of Images}  &  \\ 
\rowcolor{tableheader}  \multicolumn{1}{| c |}{\bf Dataset} & \multicolumn{1}{ c |}{\bf Classes} & \multicolumn{1}{ c }{\bf Train} & \multicolumn{1}{ c |}{\bf Val.} & \multicolumn{1}{ c |}{\bf Application} \\ \hline
VOC	&	20	&	16,551	&	4,952	&	Basic object categories for web images	\\ \hline
COCO	&	80	&	118,287	&	5,000	&	Common objects, moderate complexity	\\ \hline
LVIS	&	1,203	&	100,170	&	19,809	&	Large vocabulary, high complexity \\ \hline
BDD	&	10	&	70,000	&	10,000	&	Autonomous driving views \& objects	\\ \hline
	\end{tabular}
\label{tab:dataset}
\end{table}

For Auto-Labeling (AL) experiments and evaluation, we use the datasets listed in \cref{tab:dataset}, which vary in terms of application complexity and domain.
Notably, there is no leakage between these datasets and the AL foundation models \cite{Cheng_2024_CVPR,wang2025yoloerealtimeseeing,shilong2024}.
For the PASCAL {\bf V}isual {\bf O}bject {\bf C}lasses Dataset (VOC) \cite{EvEtAl10}, we combine the original train \& validation splits from the 2007 \& 2012 challenges as a single train set for AL  (16,551 images); after training inference models via AL, we then use the original 2007 test split (4,952 images) for validation.
For the Microsoft {\bf C}ommon {\bf O}bjects in {\bf Co}ntext (COCO) \cite{coco}, {\bf L}arge {\bf V}ocabulary {\bf I}nstance {\bf S}egmentation (LVIS) \cite{Gupta_2019_CVPR}, and {\bf B}erkeley {\bf D}eep{\bf D}rive (BDD) datasets \cite{Yu_2020_CVPR}, we use the standard train splits for AL and downstream inference model training then the standard validation splits for inference model evaluation.

All four datasets were originally labeled by human annotators, which enables us to directly compare AL to human labels \textit{and} compare inference model performance after training on AL or human labels.
For our detection-based comparison, we do not use COCO crowd instances and all segmentation labels are converted to bounding boxes.

\noindent{\bf Remarks on LVIS}.
Two properties unique to LVIS require accommodation.
First, of the 1,203 train split classes, only 1,035 are in the validation split.
Thus, we auto-label all 1,203 classes for training, but downstream inference model mAP evaluation is only on the 1,035 validation classes.

Second, LVIS uses verbose descriptions to differentiate its 1,203 classes.
For example, class 242 is ``chili/chili vegetable/chili pepper/chili pepper vegetable/chilli/chilli vegetable/chilly/chilly vegetable/chile/chile vegetable."
Unfortunately, when prompted by LVIS's 1,203 verbose class descriptions, GDINO runs out of memory due to architectural constraints.
We tested customized splitting of the class descriptions across 30 individual forward-passes (i.e., $\sT_1, \cdots, \sT_{30}$ in \cref{eq:fa}), but this still results in $\approx300\times$ increase in AL time relative to YOLOW \& YOLOE.
For this reason, we omit GDINO from LVIS experiments.

\subsection{Auto-Labeling Costs}
\label{sec:cost}

\setlength{\tabcolsep}{5pt}
\begin{table}
	\centering
	\caption{{\bf Labeling Cost Comparison}.}
	\vspace{-0.75em}
	\scriptsize
	\begin{tabular}{| c | r | r | r | r | c | r |}
		\hline 
		\rowcolor{tableheader} & \multicolumn{1}{ c |}{\bf \# of} & \multicolumn{2}{ c |}{\bf Human Labeling}  & \multicolumn{1}{ c |}{\bf Annotation} & \multicolumn{2}{ c |}{\bf Auto-Labeling} \\ 
		\rowcolor{tableheader}  \multicolumn{1}{| c |}{\bf Dataset} & \multicolumn{1}{ c |}{\bf Classes} & \multicolumn{1}{ c }{\bf Objects} & \multicolumn{1}{ c |}{\bf Hours} & \multicolumn{1}{ c |}{\bf Service Cost} & \multicolumn{1}{ c }{\bf Hours} & \multicolumn{1}{ c |}{\bf Cost} \\ \hline
		VOC	&	20	&	40,058	&	78	&	\$1,442.09	&	0.06	&	\$0.05	\\ \hline
		COCO	&	80	&	849,945	&	1,653&	\$30,598.02	&	0.45	&	\$0.42	\\ \hline
		LVIS	&	1,203	&	1,270,141	&	2,470	&	\$45,725.08	&	0.45	&	\$0.42	\\ \hline
		BDD	&	10	&	1,286,871	&	2,502 &	\$46,327.36	&	0.31	&	\$0.29	\\ \hline \hline
\multicolumn{1}{| c |}{\bf Total} &	\multicolumn{1}{c|}{$-$}	&	3,447,015	& 6,703	 & \$124,092.54	& 1.27	&	\$1.18 \\ \hline
	\end{tabular}
	\label{tab:cost}
\end{table}

\begin{figure*}
	\centering
	\begin{minipage}{0.265\textwidth}
		\begin{tikzpicture}

\definecolor{chocolate2196855}{RGB}{219,68,55}
\definecolor{darkgrey176}{RGB}{176,176,176}
\definecolor{darkorange2551094}{RGB}{255,109,4}
\definecolor{gainsboro229}{RGB}{229,229,229}
\definecolor{lightgrey204}{RGB}{204,204,204}
\definecolor{royalblue66133244}{RGB}{66,133,244}
\definecolor{seagreen1515788}{RGB}{15,157,88}

\definecolor{gainsboro247}{RGB}{253,253,253}
\definecolor{gainsboro229}{RGB}{239,239,239}
\definecolor{mediumpurple152142213}{RGB}{152,142,213}
\definecolor{gray119}{RGB}{119,119,119}
\definecolor{sandybrown25119394}{RGB}{251,193,94}

\definecolor{yolo11x}{RGB}{68,    3,   87} 
\definecolor{yolo11l}{RGB}{52,   71,  113.5} 
\definecolor{yolo11m}{RGB}{36,  139,  140} 
\definecolor{yolo11s}{RGB}{140.5, 184.5,  89.5} 
\definecolor{yolo11n}{RGB}{245,  230,   39} 	

\definecolor{yolow}{RGB}{255,109,4}
\definecolor{yoloe}{RGB}{0,48,73}
\definecolor{gdino}{RGB}{68,137,89} 

\tikzstyle{every node}=[font=\scriptsize]

\begin{axis}[
legend cell align={left},
legend style={
  fill opacity=0,
  draw opacity=0,
  text opacity=1,
  at={(0.0, -0.07)},
  anchor=south west,
  draw=lightgrey204,
  fill=gainsboro247,
},
tick align=outside,
tick pos=left,
x grid style={darkgrey176},
xmajorgrids,
xmin=0, xmax=1,
y grid style={darkgrey176},
ylabel={\bf VOC},
ymajorgrids,
ymin=10, ymax=3000000,
axis background/.style={fill=gainsboro247},
axis line style={gainsboro229},
y grid style={gainsboro229},
x grid style={gainsboro229},
axis background/.style={fill=gainsboro247},
ticklabel style={font=\scriptsize},
xtick style={draw=none},
ytick style={draw=none},
ytick align=inside,
width=5.25cm,
height=3.5cm,
y label style={at={(axis description cs:0.165, 0.5)}},
x label style={at={(axis description cs:0.5, 0.23)}},
yticklabel style = {xshift=0.5ex},
xticklabel style = {yshift=1.25ex},
ymode = log,
ticklabel style={font=\tiny},
xminorticks=true,
minor x tick num=1,
grid=both,
ytick={1, 10, 100, 1000, 10000, 40000, 100000, 1000000},
yticklabels={\tiny $\text{1}$, \tiny $\text{10}$, \tiny $\text{100}$, \tiny $\text{1,000}$, \tiny $\text{10\rm{K}}$, \tiny $\text{\textcolor{gray119}{40\rm{K}}}$, \tiny $\text{100\rm{K}}$, \tiny  $\text{1\rm{M}}$},
title={\bf Number of Objects Labels}, 
title style={at={(0.5,1.125)},anchor=north},
]
\addplot [line width=1pt, gray119, mark=none, dashed, mark options={solid,rotate=180}, mark size=0.5, opacity=0.5, fill opacity=0.8]
table {%
0.0 40058
0.05 40058
0.1 40058
0.15 40058
0.2 40058
0.3 40058
0.4 40058
0.5 40058
0.6 40058
0.7 40058
0.8 40058
0.85 40058
0.9 40058
0.95 40058
1.0 40058
};
\addplot [line width=1.25pt,  gdino, mark=square*, mark options={solid,rotate=180}, mark size=1, opacity=0.75, fill opacity=0.8]
table {%
0.5 38964
};
\addplot [line width=1.25pt,  yoloe, mark=diamond*, mark options={solid,rotate=180}, mark size=1, opacity=0.75, fill opacity=0.8]
table {%
0.5 38964
};
\addplot [line width=1.25pt,  yolow, mark=triangle*, mark options={solid,rotate=0}, mark size=1, opacity=0.75, fill opacity=0.8]
table {%
	0.5 40028
};
\addplot [line width=1.25pt, gdino, mark=none, mark options={solid,rotate=180}, mark size=0.5, opacity=0.75, fill opacity=0.8]
table {%
	0.025 1629225
	0.05 963967
	0.1 405639
	0.15 226365
	0.2 147976
	0.3 80098
	0.4 51763
	0.5 37800
	0.6 28023
	0.7 18200
	0.8 8507
	0.85 4115
	0.9 705
	0.95 0
	0.975 0
};
\addplot [line width=1.25pt,  yoloe, mark=none, mark options={solid,rotate=180}, mark size=0.5, opacity=0.75, fill opacity=0.8]
table {%

0.025 171947
0.05 119779
0.1 85589
0.15 71052
0.2 62349
0.3 51595
0.4 44629
0.5 38964
0.6 33714
0.7 28215
0.8 21411
0.85 17038
0.9 11465
0.95 3744
0.975 184
};
\addplot [line width=1.25pt, yolow, mark=none, mark options={solid,rotate=180}, mark size=0.5, opacity=0.75, fill opacity=0.8]
table {%
	0.025 141956
	0.05 103497
	0.1 78333
	0.15 67006
	0.2 60026
	0.3 51303
	0.4 45124
	0.5 40028
	0.6 35238
	0.7 30221
	0.8 23872
	0.85 19453
	0.9 13130
	0.95 3336
	0.975 101
};
\addplot [line width=1.25pt, gdino, dashed, mark=none, mark options={solid,rotate=180}, mark size=0.5, opacity=0.5, fill opacity=0.8]
table {%
	0.9 705
	0.9 0.1
};
\end{axis}

\end{tikzpicture}
	\end{minipage}%
	\begin{minipage}{0.245\textwidth}
		\begin{tikzpicture}

\definecolor{chocolate2196855}{RGB}{219,68,55}
\definecolor{darkgrey176}{RGB}{176,176,176}
\definecolor{darkorange2551094}{RGB}{255,109,4}
\definecolor{gainsboro229}{RGB}{229,229,229}
\definecolor{lightgrey204}{RGB}{204,204,204}
\definecolor{royalblue66133244}{RGB}{66,133,244}
\definecolor{seagreen1515788}{RGB}{15,157,88}

\definecolor{gainsboro247}{RGB}{253,253,253}
\definecolor{gainsboro229}{RGB}{239,239,239}
\definecolor{mediumpurple152142213}{RGB}{152,142,213}
\definecolor{gray119}{RGB}{119,119,119}
\definecolor{sandybrown25119394}{RGB}{251,193,94}

\definecolor{yolo11x}{RGB}{68,    3,   87} 
\definecolor{yolo11l}{RGB}{52,   71,  113.5} 
\definecolor{yolo11m}{RGB}{36,  139,  140} 
\definecolor{yolo11s}{RGB}{140.5, 184.5,  89.5} 
\definecolor{yolo11n}{RGB}{245,  230,   39} 	

\definecolor{yolow}{RGB}{255,109,4}
\definecolor{yoloe}{RGB}{0,48,73}
\definecolor{gdino}{RGB}{68,137,89} 

\tikzstyle{every node}=[font=\scriptsize]

\begin{axis}[
legend cell align={left},
legend style={
  fill opacity=0,
  draw opacity=0,
  text opacity=1,
  at={(0.075,0.075)},
  anchor=south west,
  draw=lightgrey204,
  fill=gainsboro247
},
tick align=outside,
tick pos=left,
x grid style={darkgrey176},
xmajorgrids,
xmin=0, xmax=1,
y grid style={darkgrey176},
ymajorgrids,
ymin=0, ymax=1,
axis background/.style={fill=gainsboro247},
axis line style={gainsboro229},
y grid style={gainsboro229},
x grid style={gainsboro229},
axis background/.style={fill=gainsboro247},
ticklabel style={font=\scriptsize},
xtick style={draw=none},
ytick style={draw=none},
ytick align=inside,
width=5.25cm,
height=3.5cm,
x label style={at={(axis description cs:1.05, 0.15)}},
yticklabel style = {xshift=0.5ex},
xticklabel style = {yshift=1.25ex},
ticklabel style={font=\tiny},
xminorticks=true,
minor x tick num=1,
yminorticks=true,
minor y tick num=1,
grid=both,
title={\bf Precision},
title style={at={(0.5,1.125)},anchor=north},
]
\addplot [line width=1.25pt, gdino, mark=none, mark options={solid,rotate=180}, mark size=0.5, opacity=0.75, fill opacity=0.8]
table {%
	0.025 0.014207368534119
	0.05 0.0285860408084509
	0.1 0.0797014093812479
	0.15 0.154538024871336
	0.2 0.243309725901498
	0.3 0.451134859796749
	0.4 0.65484998937465
	0.5 0.781349206349206
	0.6 0.868036969632088
	0.7 0.930164835164835
	0.8 0.969789585047608
	0.85 0.986877278250304
	0.9 0.997163120567376
};
\addplot [line width=1.25pt, gdino, dashed, mark=none, mark options={solid,rotate=180}, mark size=0.5, opacity=0.5, fill opacity=0.8]
table {%
	0.9 0.997163120567376
	0.9 0
};
\addplot [line width=1.25pt,  yoloe, mark=none, mark options={solid,rotate=180}, mark size=0.5, opacity=0.75, fill opacity=0.8]
table {%
	0.025 0.222208005955323
	0.05 0.314621093847836
	0.1 0.429260769491407
	0.15 0.505151156899172
	0.2 0.562879917881602
	0.3 0.64899699583293
	0.4 0.714916310022631
	0.5 0.771455702699928
	0.6 0.824138340155425
	0.7 0.87598794967216
	0.8 0.92667320536173
	0.85 0.947822514379622
	0.9 0.972961186218927
	0.95 0.992521367521368
	0.975 1
};
\addplot [line width=1.25pt, yolow, mark=none, mark options={solid,rotate=180}, mark size=0.5, opacity=0.75, fill opacity=0.8]
table {%
	0.025 0.269428555327003
	0.05 0.364319738736389
	0.1 0.472393499546807
	0.15 0.543339402441572
	0.2 0.596224969180022
	0.3 0.671968500867396
	0.4 0.733999645421505
	0.5 0.785600079944039
	0.6 0.837391452409331
	0.7 0.886767479567188
	0.8 0.936871648793566
	0.85 0.958823831799722
	0.9 0.976999238385377
	0.95 0.99220623501199
	0.975 1
};
\end{axis}

\end{tikzpicture}
	\end{minipage}%
	\begin{minipage}{0.245\textwidth}
		\begin{tikzpicture}

\definecolor{chocolate2196855}{RGB}{219,68,55}
\definecolor{darkgrey176}{RGB}{176,176,176}
\definecolor{darkorange2551094}{RGB}{255,109,4}
\definecolor{gainsboro229}{RGB}{229,229,229}
\definecolor{lightgrey204}{RGB}{204,204,204}
\definecolor{royalblue66133244}{RGB}{66,133,244}
\definecolor{seagreen1515788}{RGB}{15,157,88}

\definecolor{gainsboro247}{RGB}{253,253,253}
\definecolor{gainsboro229}{RGB}{239,239,239}
\definecolor{mediumpurple152142213}{RGB}{152,142,213}
\definecolor{gray119}{RGB}{119,119,119}
\definecolor{sandybrown25119394}{RGB}{251,193,94}

\definecolor{yolo11x}{RGB}{68,    3,   87} 
\definecolor{yolo11l}{RGB}{52,   71,  113.5} 
\definecolor{yolo11m}{RGB}{36,  139,  140} 
\definecolor{yolo11s}{RGB}{140.5, 184.5,  89.5} 
\definecolor{yolo11n}{RGB}{245,  230,   39} 	

\definecolor{yolow}{RGB}{255,109,4}
\definecolor{yoloe}{RGB}{0,48,73}
\definecolor{gdino}{RGB}{68,137,89} 

\tikzstyle{every node}=[font=\scriptsize]

\begin{axis}[
legend cell align={left},
legend style={
  fill opacity=0,
  draw opacity=0,
  text opacity=1,
  at={(0.025,0.025)},
  anchor=south west,
  draw=lightgrey204,
  fill=gainsboro247
},
tick align=outside,
tick pos=left,
x grid style={darkgrey176},
xmajorgrids,
xmin=0, xmax=1,
y grid style={darkgrey176},
ymajorgrids,
ymin=0, ymax=1,
axis background/.style={fill=gainsboro247},
axis line style={gainsboro229},
y grid style={gainsboro229},
x grid style={gainsboro229},
axis background/.style={fill=gainsboro247},
ticklabel style={font=\scriptsize},
xtick style={draw=none},
ytick style={draw=none},
ytick align=inside,
width=5.25cm,
height=3.5cm,
y label style={at={(axis description cs:0.19, 0.5)}},
yticklabel style = {xshift=0.5ex},
xticklabel style = {yshift=1.25ex},
ticklabel style={font=\tiny},
xminorticks=true,
minor x tick num=1,
grid=both,
yminorticks=true,
minor y tick num=1,
title={\bf Recall},
title style={at={(0.5,1.125)},anchor=north},
x label style={at={(axis description cs:0.5, 0.15)}},
]
\addplot [line width=1.25pt, gdino, mark=none, mark options={solid,rotate=180}, mark size=0.5, opacity=0.75, fill opacity=0.8]
table {%
	0.025 0.577837136152579
	0.05 0.687902541315093
	0.1 0.807079734385142
	0.15 0.87328373857906
	0.2 0.898796744720156
	0.3 0.902067002845873
	0.4 0.846198012881322
	0.5 0.737305906435668
	0.6 0.607244495481552
	0.7 0.422612212292176
	0.8 0.205951370512756
	0.85 0.101378001897249
	0.9 0.0175495531479355
	0.95 0
	0.975 0
};
\addplot [line width=1.25pt,  yoloe, mark=none, mark options={solid,rotate=180}, mark size=0.5, opacity=0.75, fill opacity=0.8]
table {%
	0.025 0.95381696540017
	0.05 0.940760896699785
	0.1 0.917170103350142
	0.15 0.89600079884168
	0.2 0.876104648260023
	0.3 0.835912926256927
	0.4 0.79649508213091
	0.5 0.750386938938539
	0.6 0.693619252084478
	0.7 0.617005342253732
	0.8 0.495306805132558
	0.85 0.403140446352788
	0.9 0.278471216735733
	0.95 0.0930650556692795
	0.975 0.00459333965749663
};
\addplot [line width=1.25pt, yolow, mark=none, mark options={solid,rotate=180}, mark size=0.5, opacity=0.75, fill opacity=0.8]
table {%
	0.025 0.954790553697139
	0.05 0.941285136552
	0.1 0.923760547206551
	0.15 0.908857157122173
	0.2 0.893429527185581
	0.3 0.860602126915972
	0.4 0.82682610215188
	0.5 0.785011732987169
	0.6 0.736631883768536
	0.7 0.669004942832892
	0.8 0.558315442608218
	0.85 0.465624843976234
	0.9 0.320235658295472
	0.95 0.0827050776374257
	0.975 0.00252134405112587
};
\end{axis}

\end{tikzpicture}
	\end{minipage}%
	\begin{minipage}{0.245\textwidth}
		\begin{tikzpicture}

\definecolor{chocolate2196855}{RGB}{219,68,55}
\definecolor{darkgrey176}{RGB}{176,176,176}
\definecolor{darkorange2551094}{RGB}{255,109,4}
\definecolor{gainsboro229}{RGB}{229,229,229}
\definecolor{lightgrey204}{RGB}{204,204,204}
\definecolor{royalblue66133244}{RGB}{66,133,244}
\definecolor{seagreen1515788}{RGB}{15,157,88}

\definecolor{gainsboro247}{RGB}{253,253,253}
\definecolor{gainsboro229}{RGB}{239,239,239}
\definecolor{mediumpurple152142213}{RGB}{152,142,213}
\definecolor{gray119}{RGB}{119,119,119}
\definecolor{sandybrown25119394}{RGB}{251,193,94}

\definecolor{yolo11x}{RGB}{68,    3,   87} 
\definecolor{yolo11l}{RGB}{52,   71,  113.5} 
\definecolor{yolo11m}{RGB}{36,  139,  140} 
\definecolor{yolo11s}{RGB}{140.5, 184.5,  89.5} 
\definecolor{yolo11n}{RGB}{245,  230,   39} 	

\definecolor{yolow}{RGB}{255,109,4}
\definecolor{yoloe}{RGB}{0,48,73}
\definecolor{gdino}{RGB}{68,137,89} 

\tikzstyle{every node}=[font=\scriptsize]

\begin{axis}[
legend cell align={left},
legend style={
  fill opacity=0,
  draw opacity=0,
  text opacity=1,
  at={(0.075,0.075)},
  anchor=south west,
  draw=lightgrey204,
  fill=gainsboro247
},
tick align=outside,
tick pos=left,
x grid style={darkgrey176},
xmajorgrids,
xmin=0, xmax=1,
y grid style={darkgrey176},
ymajorgrids,
ymin=0, ymax=1,
axis background/.style={fill=gainsboro247},
axis line style={gainsboro229},
y grid style={gainsboro229},
x grid style={gainsboro229},
axis background/.style={fill=gainsboro247},
ticklabel style={font=\tiny},
xtick style={draw=none},
ytick style={draw=none},
ytick align=inside,
width=5.25cm,
height=3.5cm,
y label style={at={(axis description cs:0.19, 0.5)}},
x label style={at={(axis description cs:0.5, 0.15)}},
yticklabel style = {xshift=0.5ex},
xticklabel style = {yshift=1.25ex},
xminorticks=true,
minor x tick num=1,
grid=both,
yminorticks=true,
minor y tick num=1,
title={\bf ${\text{F}}_\text{1}$ Score},
title style={at={(0.5,1.125)},anchor=north},
]
\addplot [line width=1.25pt,  gdino, mark=square*, mark options={solid,rotate=180}, mark size=1, opacity=0.75, fill opacity=0.8]
table {%
	0.5 0.758688895168127
};
\addplot [line width=1.25pt,  yoloe, mark=diamond*, mark options={solid,rotate=180}, mark size=1, opacity=0.75, fill opacity=0.8]
table {%
	0.5 0.760775480246007
};
\addplot [line width=1.25pt,  yolow, mark=triangle*, mark options={solid,rotate=0}, mark size=1, opacity=0.75, fill opacity=0.8]
table {%
	0.5 0.785305796269011
};
\addplot [line width=1.25pt, gdino, mark=none, mark options={solid,rotate=180}, mark size=0.5, opacity=0.75, fill opacity=0.8]
table {%
	0.025 0.0277328649486037
	0.05 0.0548910634695351
	0.1 0.145076139170782
	0.15 0.262604955277885
	0.2 0.382952019315656
	0.3 0.601468091481075
	0.4 0.73832783350214
	0.5 0.758688895168127
	0.6 0.714589973707789
	0.7 0.581173401077964
	0.8 0.339750849377123
	0.85 0.183867973649062
	0.9 0.0344920638814611
	0.95 0
	0.975 0
};
\addplot [line width=1.25pt,  yoloe, mark=none, mark options={solid,rotate=180}, mark size=0.5, opacity=0.75, fill opacity=0.8]
table {%
	0.025 0.360444329143181
	0.05 0.471542884313394
	0.1 0.584813007871258
	0.15 0.646062460624606
	0.2 0.6854023650727
	0.3 0.730690757531123
	0.4 0.753504079728884
	0.5 0.760775480246007
	0.6 0.753266822100526
	0.7 0.724034391340647
	0.8 0.645561177178741
	0.85 0.565678856662463
	0.9 0.433010500164975
	0.95 0.16967261768869
	0.975 0.00914467471795636
};
\addplot [line width=1.25pt, yolow, mark=none, mark options={solid,rotate=180}, mark size=0.5, opacity=0.75, fill opacity=0.8]
table {%
	0.025 0.420264375267837
	0.05 0.525317822437393
	0.1 0.625115084761511
	0.15 0.680097885376971
	0.2 0.715179249430478
	0.3 0.754676503103075
	0.4 0.777652555704257
	0.5 0.785305796269011
	0.6 0.783786655333617
	0.7 0.762646025128417
	0.8 0.699671515720319
	0.85 0.626842096419149
	0.9 0.482364443107468
	0.95 0.152555652855233
	0.975 0.00503000572723424
};
\end{axis}

\end{tikzpicture}
	\end{minipage}%
	\vspace{-1.65em}
	\\
	\begin{minipage}{0.265\textwidth}
		\begin{tikzpicture}

\definecolor{chocolate2196855}{RGB}{219,68,55}
\definecolor{darkgrey176}{RGB}{176,176,176}
\definecolor{darkorange2551094}{RGB}{255,109,4}
\definecolor{gainsboro229}{RGB}{229,229,229}
\definecolor{lightgrey204}{RGB}{204,204,204}
\definecolor{royalblue66133244}{RGB}{66,133,244}
\definecolor{seagreen1515788}{RGB}{15,157,88}

\definecolor{gainsboro247}{RGB}{253,253,253}
\definecolor{gainsboro229}{RGB}{239,239,239}
\definecolor{gainsboro245}{RGB}{247,247,247}
\definecolor{mediumpurple152142213}{RGB}{152,142,213}
\definecolor{gray119}{RGB}{119,119,119}
\definecolor{sandybrown25119394}{RGB}{251,193,94}

\definecolor{yolo11x}{RGB}{68,    3,   87} 
\definecolor{yolo11l}{RGB}{52,   71,  113.5} 
\definecolor{yolo11m}{RGB}{36,  139,  140} 
\definecolor{yolo11s}{RGB}{140.5, 184.5,  89.5} 
\definecolor{yolo11n}{RGB}{245,  230,   39} 	

\definecolor{yolow}{RGB}{255,109,4}
\definecolor{yoloe}{RGB}{0,48,73}
\definecolor{gdino}{RGB}{68,137,89} 

\tikzstyle{every node}=[font=\scriptsize]

\begin{axis}[
legend cell align={left},
legend style={
  fill opacity=0,
  draw opacity=0,
  text opacity=1,
  at={(0.075, -0.015)},
  anchor=south west,
  draw=lightgrey204,
  fill=gainsboro247
},
tick align=outside,
tick pos=left,
x grid style={darkgrey176},
xlabel={\textcolor{white}{A}},
xmajorgrids,
xmin=0, xmax=1,
y grid style={darkgrey176},
ylabel={\bf COCO},
ymajorgrids,
ymin=10, ymax=25000000,
axis background/.style={fill=gainsboro247},
axis line style={gainsboro229},
y grid style={gainsboro229},
x grid style={gainsboro229},
axis background/.style={fill=gainsboro247},
ticklabel style={font=\scriptsize},
xtick style={draw=none},
ytick style={draw=none},
ytick align=inside,
width=5.25cm,
height=3.5cm,
title style={at={(0.5,0.925)},anchor=south},
y label style={at={(axis description cs:0.165, 0.5)}},
x label style={at={(axis description cs:0.5, 1.365)}},
yticklabel style = {xshift=0.5ex},
xticklabel style = {yshift=1.25ex},
ymode = log,
ytick={10, 100, 1000, 10000, 100000, 850000, 1000000, 10000000},
yticklabels={\tiny $\text{10}$, \tiny $\text{100}$, \tiny $\text{1,000}$, \tiny $\text{10\rm{K}}$, \tiny $\text{100\rm{K}}$, \tiny $\text{\textcolor{gray119}{850\rm{K}}}$, , \tiny  $\text{10\rm{M}}$},
ticklabel style={font=\tiny},
xminorticks=true,
minor x tick num=1,
grid=both,
]
\addplot [line width=1pt, gray119, mark=none, dashed, mark options={solid,rotate=180}, mark size=0.5, opacity=0.5, fill opacity=0.8]
table {%
0 849942
0.05 849942
0.1 849942
0.15 849942
0.2 849942
0.3 849942
0.4 849942
0.5 849942
0.6 849942
0.7 849942
0.8 849942
0.85 849942
0.9 849942
0.95 849942
1 849942
};
\addplot [line width=1.25pt,  gdino, mark=square*, mark options={solid,rotate=180}, mark size=1, opacity=0.75, fill opacity=0.8]
table {%
0.4 713480
};
\addplot [line width=1.25pt,  yoloe, mark=diamond*, mark options={solid,rotate=180}, mark size=1, opacity=0.75, fill opacity=0.8]
table {%
0.2 946312
};
\addplot [line width=1.25pt,  yolow, mark=triangle*, mark options={solid,rotate=0}, mark size=1, opacity=0.75, fill opacity=0.8]
table {%
0.3 807534
};
\addplot [line width=1.25pt, gdino, mark=none, mark options={solid,rotate=180}, mark size=0.5, opacity=0.75, fill opacity=0.8]
table {%
0.025 13326455
0.05 10099155
0.1 5388641
0.15 3303436
0.2 2232127
0.3 1182967
0.4 713480
0.5 479099
0.6 329510
0.7 211857
0.8 108240
0.85 60829
0.9 15319
0.95 16
};
\addplot [line width=1.25pt,  yoloe, mark=none, mark options={solid,rotate=180}, mark size=0.5, opacity=0.75, fill opacity=0.8]
table {%
0.025 3031620
0.05 2072345
0.1 1409309
0.15 1119737
0.2 946312
0.3 733307
0.4 594872
0.5 487148
0.6 393869
0.7 306560
0.8 218291
0.85 170493
0.9 114590
0.95 36511
0.975 1937
};
\addplot [line width=1.25pt, yolow, mark=none, mark options={solid,rotate=180}, mark size=0.5, opacity=0.75, fill opacity=0.8]
table {%
0.025 2948657
0.05 2082042
0.1 1461535
0.15 1182712
0.2 1013600
0.3 807534
0.4 666947
0.5 557894
0.6 462360
0.7 369744
0.8 267692
0.85 206832
0.9 134640
0.95 37456
0.975 2085
};
\addplot [line width=1.25pt, gdino, dashed, mark=none, mark options={solid,rotate=180}, mark size=0.5, opacity=0.5, fill opacity=0.8]
table {%
	0.95 16
	0.95 0.1
};
\end{axis}

\end{tikzpicture}
	\end{minipage}%
	\begin{minipage}{0.245\textwidth}
		\begin{tikzpicture}

\definecolor{chocolate2196855}{RGB}{219,68,55}
\definecolor{darkgrey176}{RGB}{176,176,176}
\definecolor{darkorange2551094}{RGB}{255,109,4}
\definecolor{gainsboro229}{RGB}{229,229,229}
\definecolor{lightgrey204}{RGB}{204,204,204}
\definecolor{royalblue66133244}{RGB}{66,133,244}
\definecolor{seagreen1515788}{RGB}{15,157,88}

\definecolor{gainsboro247}{RGB}{253,253,253}
\definecolor{gainsboro229}{RGB}{239,239,239}
\definecolor{mediumpurple152142213}{RGB}{152,142,213}
\definecolor{gray119}{RGB}{119,119,119}
\definecolor{sandybrown25119394}{RGB}{251,193,94}

\definecolor{yolo11x}{RGB}{68,    3,   87} 
\definecolor{yolo11l}{RGB}{52,   71,  113.5} 
\definecolor{yolo11m}{RGB}{36,  139,  140} 
\definecolor{yolo11s}{RGB}{140.5, 184.5,  89.5} 
\definecolor{yolo11n}{RGB}{245,  230,   39} 	

\definecolor{yolow}{RGB}{255,109,4}
\definecolor{yoloe}{RGB}{0,48,73}
\definecolor{gdino}{RGB}{68,137,89} 

\tikzstyle{every node}=[font=\scriptsize]

\begin{axis}[
legend cell align={left},
legend style={
  fill opacity=0,
  draw opacity=0,
  text opacity=1,
  at={(0.075,0.075)},
  anchor=south west,
  draw=lightgrey204,
  fill=gainsboro247
},
tick align=outside,
tick pos=left,
x grid style={darkgrey176},
xmajorgrids,
xmin=0, xmax=1,
y grid style={darkgrey176},
ymajorgrids,
ymin=0, ymax=1,
axis background/.style={fill=gainsboro247},
axis line style={gainsboro229},
y grid style={gainsboro229},
x grid style={gainsboro229},
axis background/.style={fill=gainsboro247},
ticklabel style={font=\scriptsize},
xtick style={draw=none},
ytick style={draw=none},
ytick align=inside,
width=5.25cm,
height=3.5cm,
x label style={at={(axis description cs:0.5, 0.0725)}},
yticklabel style = {xshift=0.5ex},
xticklabel style = {yshift=1.25ex},
ticklabel style={font=\tiny},
title style={at={(0.5,0.925)},anchor=south},
xminorticks=true,
minor x tick num=1,
yminorticks=true,
minor y tick num=1,
grid=both,
]
\addplot [line width=1.25pt, gdino, mark=none, mark options={solid,rotate=180}, mark size=0.5, opacity=0.75, fill opacity=0.8]
table {%

0.025 0.0331677854313094
0.05 0.0521333715543528
0.1 0.107884900849769
0.15 0.188554583772775
0.2 0.281444111378967
0.3 0.495486349154287
0.4 0.701046980994562
0.5 0.825883585647225
0.6 0.906907225880853
0.7 0.956480078543546
0.8 0.983268662232077
0.85 0.991089776258035
0.9 0.994320778118676
0.95 1
};
\addplot [line width=1.25pt, gdino, dashed, mark=none, mark options={solid,rotate=180}, mark size=0.5, opacity=0.5, fill opacity=0.8]
table {%
	0.95 1
	0.95 0
};
\addplot [line width=1.25pt,  yoloe, mark=none, mark options={solid,rotate=180}, mark size=0.5, opacity=0.75, fill opacity=0.8]
table {%
	0.025 0.217136052671509
	0.05 0.303365028506354
	0.1 0.417388237781778
	0.15 0.496900611482875
	0.2 0.55910841244748
	0.3 0.655245347446567
	0.4 0.729878024179992
	0.5 0.792046770180725
	0.6 0.847220776451054
	0.7 0.896933716075157
	0.8 0.94057473739183
	0.85 0.960320951593321
	0.9 0.978217994589406
	0.95 0.990989017008573
	0.975 0.992772328342798
};
\addplot [line width=1.25pt, yolow, mark=none, mark options={solid,rotate=180}, mark size=0.5, opacity=0.75, fill opacity=0.8]
table {%
	0.025 0.231514889659937
	0.05 0.31558777392579
	0.1 0.425309691522954
	0.15 0.502331928652115
	0.2 0.562579913180742
	0.3 0.653250513291081
	0.4 0.727339653675629
	0.5 0.787289341702904
	0.6 0.84058309542348
	0.7 0.891562811026007
	0.8 0.93975165488696
	0.85 0.961282586833759
	0.9 0.979270647653001
	0.95 0.990602306706536
	0.975 0.991846522781775
};
\end{axis}

\end{tikzpicture}
	\end{minipage}%
	\begin{minipage}{0.245\textwidth}
		\begin{tikzpicture}

\definecolor{chocolate2196855}{RGB}{219,68,55}
\definecolor{darkgrey176}{RGB}{176,176,176}
\definecolor{darkorange2551094}{RGB}{255,109,4}
\definecolor{gainsboro229}{RGB}{229,229,229}
\definecolor{lightgrey204}{RGB}{204,204,204}
\definecolor{royalblue66133244}{RGB}{66,133,244}
\definecolor{seagreen1515788}{RGB}{15,157,88}

\definecolor{gainsboro247}{RGB}{253,253,253}
\definecolor{gainsboro229}{RGB}{239,239,239}
\definecolor{mediumpurple152142213}{RGB}{152,142,213}
\definecolor{gray119}{RGB}{119,119,119}
\definecolor{sandybrown25119394}{RGB}{251,193,94}

\definecolor{yolo11x}{RGB}{68,    3,   87} 
\definecolor{yolo11l}{RGB}{52,   71,  113.5} 
\definecolor{yolo11m}{RGB}{36,  139,  140} 
\definecolor{yolo11s}{RGB}{140.5, 184.5,  89.5} 
\definecolor{yolo11n}{RGB}{245,  230,   39} 	

\definecolor{yolow}{RGB}{255,109,4}
\definecolor{yoloe}{RGB}{0,48,73}
\definecolor{gdino}{RGB}{68,137,89} 

\tikzstyle{every node}=[font=\scriptsize]

\begin{axis}[
legend cell align={left},
legend style={
  fill opacity=0,
  draw opacity=0,
  text opacity=1,
  at={(0.075,0.075)},
  anchor=south west,
  draw=lightgrey204,
  fill=gainsboro247
},
tick align=outside,
tick pos=left,
x grid style={darkgrey176},
xmajorgrids,
xmin=0, xmax=1,
y grid style={darkgrey176},
ymajorgrids,
ymin=0, ymax=1,
axis background/.style={fill=gainsboro247},
axis line style={gainsboro229},
y grid style={gainsboro229},
x grid style={gainsboro229},
axis background/.style={fill=gainsboro247},
ticklabel style={font=\scriptsize},
xtick style={draw=none},
ytick style={draw=none},
ytick align=inside,
width=5.25cm,
height=3.5cm,
y label style={at={(axis description cs:0.19, 0.5)}},
yticklabel style = {xshift=0.5ex},
xticklabel style = {yshift=1.25ex},
ticklabel style={font=\tiny},
title style={at={(0.5,0.925)},anchor=south},
xminorticks=true,
minor x tick num=1,
grid=both,
yminorticks=true,
minor y tick num=1,
]
\addplot [line width=1.25pt, gdino, mark=none, mark options={solid,rotate=180}, mark size=0.5, opacity=0.75, fill opacity=0.8]
table {%
	0.025 0.52004607373209
	0.05 0.619457562986651
	0.1 0.683991378235221
	0.15 0.732847653133979
	0.2 0.739131611333479
	0.3 0.68962823345593
	0.4 0.588490744074302
	0.5 0.465537648451306
	0.6 0.351594579394829
	0.7 0.238412738751585
	0.8 0.125219132599636
	0.85 0.0709307223316415
	0.9 0.0179212228599128
	0.95 1.88248139284798e-05
	0.975 0
};
\addplot [line width=1.25pt,  yoloe, mark=none, mark options={solid,rotate=180}, mark size=0.5, opacity=0.75, fill opacity=0.8]
table {%
	0.025 0.774492847747258
	0.05 0.739670471632182
	0.1 0.692081342020985
	0.15 0.654630551261145
	0.2 0.622502476639582
	0.3 0.565327987086178
	0.4 0.510839563170193
	0.5 0.453965094088773
	0.6 0.392607966190634
	0.7 0.323509133564408
	0.8 0.241568248186347
	0.85 0.192634320930134
	0.9 0.131884293281189
	0.95 0.0425699635975161
	0.975 0.00226250732402917
};
\addplot [line width=1.25pt, yolow, mark=none, mark options={solid,rotate=180}, mark size=0.5, opacity=0.75, fill opacity=0.8]
table {%
	0.025 0.803181864174261
	0.05 0.773072750846528
	0.1 0.731349903875794
	0.15 0.699005343894054
	0.2 0.670905779453186
	0.3 0.620656468323721
	0.4 0.570741297641486
	0.5 0.516769379557664
	0.6 0.457268848933221
	0.7 0.387849994470211
	0.8 0.295977843194006
	0.85 0.233926550282255
	0.9 0.155127055728509
	0.95 0.0436547435001447
	0.975 0.00243310720025602
};
\end{axis}

\end{tikzpicture}
	\end{minipage}%
	\begin{minipage}{0.245\textwidth}
		\begin{tikzpicture}

\definecolor{chocolate2196855}{RGB}{219,68,55}
\definecolor{darkgrey176}{RGB}{176,176,176}
\definecolor{darkorange2551094}{RGB}{255,109,4}
\definecolor{gainsboro229}{RGB}{229,229,229}
\definecolor{lightgrey204}{RGB}{204,204,204}
\definecolor{royalblue66133244}{RGB}{66,133,244}
\definecolor{seagreen1515788}{RGB}{15,157,88}

\definecolor{gainsboro247}{RGB}{253,253,253}
\definecolor{gainsboro229}{RGB}{239,239,239}
\definecolor{mediumpurple152142213}{RGB}{152,142,213}
\definecolor{gray119}{RGB}{119,119,119}
\definecolor{sandybrown25119394}{RGB}{251,193,94}

\definecolor{yolo11x}{RGB}{68,    3,   87} 
\definecolor{yolo11l}{RGB}{52,   71,  113.5} 
\definecolor{yolo11m}{RGB}{36,  139,  140} 
\definecolor{yolo11s}{RGB}{140.5, 184.5,  89.5} 
\definecolor{yolo11n}{RGB}{245,  230,   39} 	

\definecolor{yolow}{RGB}{255,109,4}
\definecolor{yoloe}{RGB}{0,48,73}
\definecolor{gdino}{RGB}{68,137,89} 

\tikzstyle{every node}=[font=\scriptsize]

\begin{axis}[
legend cell align={left},
legend style={
  fill opacity=0,
  draw opacity=0,
  text opacity=1,
  at={(0.075,0.075)},
  anchor=south west,
  draw=lightgrey204,
  fill=gainsboro247
},
tick align=outside,
tick pos=left,
x grid style={darkgrey176},
xmajorgrids,
xmin=0, xmax=1,
y grid style={darkgrey176},
ymajorgrids,
ymin=0, ymax=1,
axis background/.style={fill=gainsboro247},
axis line style={gainsboro229},
y grid style={gainsboro229},
x grid style={gainsboro229},
axis background/.style={fill=gainsboro247},
ticklabel style={font=\tiny},
xtick style={draw=none},
ytick style={draw=none},
ytick align=inside,
width=5.25cm,
height=3.5cm,
y label style={at={(axis description cs:0.19, 0.5)}},
x label style={at={(axis description cs:0.5, 0.0725)}},
yticklabel style = {xshift=0.5ex},
xticklabel style = {yshift=1.25ex},
title style={at={(0.5,0.925)},anchor=south},
xminorticks=true,
minor x tick num=1,
grid=both,
yminorticks=true,
minor y tick num=1,
]
\addplot [line width=1.25pt,  gdino, mark=square*, mark options={solid,rotate=180}, mark size=1, opacity=0.75, fill opacity=0.8]
table {%
	0.4 0.639856673374175
};
\addplot [line width=1.25pt,  yoloe, mark=diamond*, mark options={solid,rotate=180}, mark size=1, opacity=0.75, fill opacity=0.8]
table {%
	0.3 0.606974645175838
};
\addplot [line width=1.25pt,  yolow, mark=triangle*, mark options={solid,rotate=0}, mark size=1, opacity=0.75, fill opacity=0.8]
table {%
	0.4 0.639594591298374
};
\addplot [line width=1.25pt, gdino, mark=none, mark options={solid,rotate=180}, mark size=0.5, opacity=0.75, fill opacity=0.8]
table {%
	0.025 0.0623584398772128
	0.05 0.0961728624744123
	0.1 0.18637341203924
	0.15 0.299938026348673
	0.2 0.407660568274104
	0.3 0.576655423336706
	0.4 0.639856673374175
	0.5 0.595436860111915
	0.6 0.50673533132336
	0.7 0.38168617600883
	0.8 0.222147775683534
	0.85 0.132386736073063
	0.9 0.0352078736936023
	0.95 3.76489191230626e-05
	0.975 0
};
\addplot [line width=1.25pt,  yoloe, mark=none, mark options={solid,rotate=180}, mark size=0.5, opacity=0.75, fill opacity=0.8]
table {%
	0.025 0.339179948690759
	0.05 0.430263694154612
	0.1 0.520729215124836
	0.15 0.564963123432803
	0.2 0.589104881603604
	0.3 0.606974645175838
	0.4 0.601024076455516
	0.5 0.577139908308341
	0.6 0.536567050781831
	0.7 0.475509769978781
	0.8 0.38440864493046
	0.85 0.320898440371018
	0.9 0.232431894431704
	0.95 0.0816332055957845
	0.975 0.0045147256828728
};
\addplot [line width=1.25pt, yolow, mark=none, mark options={solid,rotate=180}, mark size=0.5, opacity=0.75, fill opacity=0.8]
table {%
	0.025 0.359426198974938
	0.05 0.448206402217747
	0.1 0.53784225410852
	0.15 0.584569730017996
	0.2 0.611986206911355
	0.3 0.636536516969175
	0.4 0.639594591298374
	0.5 0.623970405643839
	0.6 0.592320974897546
	0.7 0.540548960962084
	0.8 0.45017241780404
	0.85 0.376284806401369
	0.9 0.267827362271502
	0.95 0.0836242587880523
	0.975 0.00485430626024762
};
\end{axis}

\end{tikzpicture}
	\end{minipage}%
	\vspace{-1.65em}
	\\
	\begin{minipage}{0.265\textwidth}
		\begin{tikzpicture}

\definecolor{chocolate2196855}{RGB}{219,68,55}
\definecolor{darkgrey176}{RGB}{176,176,176}
\definecolor{darkorange2551094}{RGB}{255,109,4}
\definecolor{gainsboro229}{RGB}{229,229,229}
\definecolor{lightgrey204}{RGB}{204,204,204}
\definecolor{royalblue66133244}{RGB}{66,133,244}
\definecolor{seagreen1515788}{RGB}{15,157,88}

\definecolor{gainsboro247}{RGB}{253,253,253}
\definecolor{gainsboro229}{RGB}{239,239,239}
\definecolor{mediumpurple152142213}{RGB}{152,142,213}
\definecolor{gray119}{RGB}{119,119,119}
\definecolor{sandybrown25119394}{RGB}{251,193,94}

\definecolor{yolo11x}{RGB}{68,    3,   87} 
\definecolor{yolo11l}{RGB}{52,   71,  113.5} 
\definecolor{yolo11m}{RGB}{36,  139,  140} 
\definecolor{yolo11s}{RGB}{140.5, 184.5,  89.5} 
\definecolor{yolo11n}{RGB}{245,  230,   39} 	

\definecolor{yolow}{RGB}{255,109,4}
\definecolor{yoloe}{RGB}{0,48,73}
\definecolor{gdino}{RGB}{68,137,89} 

\tikzstyle{every node}=[font=\scriptsize]

\begin{axis}[
legend cell align={left},
legend style={
  fill opacity=0,
  draw opacity=0,
  text opacity=1,
  at={(0.0, -0.07)},
  anchor=south west,
  draw=lightgrey204,
  fill=gainsboro247,
},
tick align=outside,
tick pos=left,
x grid style={darkgrey176},
xlabel={\textcolor{white}{A}},
xmajorgrids,
xmin=0, xmax=1,
y grid style={darkgrey176},
ylabel={\bf LVIS},
ymajorgrids,
ymin=1000, ymax=15000000,
axis background/.style={fill=gainsboro247},
axis line style={gainsboro229},
y grid style={gainsboro229},
x grid style={gainsboro229},
axis background/.style={fill=gainsboro247},
ticklabel style={font=\scriptsize},
xtick style={draw=none},
ytick style={draw=none},
ytick align=inside,
width=5.25cm,
height=3.5cm,
y label style={at={(axis description cs:0.165, 0.5)}},
x label style={at={(axis description cs:0.5, 1.365)}},
yticklabel style = {xshift=0.5ex},
xticklabel style = {yshift=1.25ex},
ymode = log,
ticklabel style={font=\tiny},
xminorticks=true,
minor x tick num=1,
grid=both,
ytick={1000, 10000, 100000, 1000000, 1300000, 10000000},
yticklabels={\tiny $\text{1,000}$, \tiny $\text{10\rm{K}}$, \tiny $\text{100\rm{K}}$, , \tiny  $\text{\textcolor{gray119}{1.3\rm{M}}}$, \tiny  $\text{10\rm{M}}$},
]
\addplot [line width=1pt, gray119, mark=none, dashed, mark options={solid,rotate=180}, mark size=0.5, opacity=0.5, fill opacity=0.8]
table {%
0.0 1270139
0.05 1270139
0.1 1270139
0.15 1270139
0.2 1270139
0.3 1270139
0.4 1270139
0.5 1270139
0.6 1270139
0.7 1270139
0.8 1270139
0.85 1270139
0.9 1270139
0.95 1270139
1.0 1270139
};
\addplot [line width=1.25pt,  yoloe, mark=diamond*, mark options={solid,rotate=180}, mark size=1, opacity=0.75, fill opacity=0.8]
table {%
0.2 1311999
};
\addplot [line width=1.25pt,  yolow, mark=triangle*, mark options={solid,rotate=0}, mark size=1, opacity=0.75, fill opacity=0.8]
table {%
0.3 1166848
};
\addplot [line width=1.25pt,  yoloe, mark=none, mark options={solid,rotate=180}, mark size=0.5, opacity=0.75, fill opacity=0.8]
table {%
0.025 5437559
0.05 3522809
0.1 2215458
0.15 1648620
0.2 1311999
0.3 909603
0.4 661769
0.5 486847
0.6 352255
0.7 242101
0.8 145360
0.85 99288
0.9 54231
0.95 13471
0.975 1118
};
\addplot [line width=1.25pt, yolow, mark=none, mark options={solid,rotate=180}, mark size=0.5, opacity=0.75, fill opacity=0.8]
table {%
0.025 7322599
0.05 4724850
0.1 2923119
0.15 2153357
0.2 1701295
0.3 1166848
0.4 842098
0.5 610040
0.6 424728
0.7 267003
0.8 135523
0.85 83707
0.9 42737
0.95 15357
0.975 6257
};
\end{axis}

\end{tikzpicture}
	\end{minipage}%
	\begin{minipage}{0.245\textwidth}
		\begin{tikzpicture}

\definecolor{chocolate2196855}{RGB}{219,68,55}
\definecolor{darkgrey176}{RGB}{176,176,176}
\definecolor{darkorange2551094}{RGB}{255,109,4}
\definecolor{gainsboro229}{RGB}{229,229,229}
\definecolor{lightgrey204}{RGB}{204,204,204}
\definecolor{royalblue66133244}{RGB}{66,133,244}
\definecolor{seagreen1515788}{RGB}{15,157,88}

\definecolor{gainsboro247}{RGB}{253,253,253}
\definecolor{gainsboro229}{RGB}{239,239,239}
\definecolor{mediumpurple152142213}{RGB}{152,142,213}
\definecolor{gray119}{RGB}{119,119,119}
\definecolor{sandybrown25119394}{RGB}{251,193,94}

\definecolor{yolo11x}{RGB}{68,    3,   87} 
\definecolor{yolo11l}{RGB}{52,   71,  113.5} 
\definecolor{yolo11m}{RGB}{36,  139,  140} 
\definecolor{yolo11s}{RGB}{140.5, 184.5,  89.5} 
\definecolor{yolo11n}{RGB}{245,  230,   39} 	

\definecolor{yolow}{RGB}{255,109,4}
\definecolor{yoloe}{RGB}{0,48,73}
\definecolor{gdino}{RGB}{68,137,89} 

\tikzstyle{every node}=[font=\scriptsize]

\begin{axis}[
legend cell align={left},
legend style={
  fill opacity=0,
  draw opacity=0,
  text opacity=1,
  at={(0.075,0.075)},
  anchor=south west,
  draw=lightgrey204,
  fill=gainsboro247
},
tick align=outside,
tick pos=left,
x grid style={darkgrey176},
xmajorgrids,
xmin=0, xmax=1,
y grid style={darkgrey176},
ymajorgrids,
ymin=0, ymax=1,
axis background/.style={fill=gainsboro247},
axis line style={gainsboro229},
y grid style={gainsboro229},
x grid style={gainsboro229},
axis background/.style={fill=gainsboro247},
ticklabel style={font=\scriptsize},
xtick style={draw=none},
ytick style={draw=none},
ytick align=inside,
width=5.25cm,
height=3.5cm,
x label style={at={(axis description cs:0.5, 0.0725)}},
yticklabel style = {xshift=0.5ex},
xticklabel style = {yshift=1.25ex},
ticklabel style={font=\tiny},
xminorticks=true,
minor x tick num=1,
yminorticks=true,
minor y tick num=1,
grid=both,
]
\addplot [line width=1.25pt,  yoloe, mark=none, mark options={solid,rotate=180}, mark size=0.5, opacity=0.75, fill opacity=0.8]
table {%
0.025 0.0767577878235436
0.05 0.106840876130383
0.1 0.148507893176039
0.15 0.180427266441023
0.2 0.208028359777713
0.3 0.257053901537264
0.4 0.302109950753208
0.5 0.346190897756379
0.6 0.390799278931456
0.7 0.438903598085097
0.8 0.496319482663731
0.85 0.532884135041495
0.9 0.586177647471004
0.95 0.66587484225373
0.975 0.796064400715564
};
\addplot [line width=1.25pt, yolow, mark=none, mark options={solid,rotate=180}, mark size=0.5, opacity=0.75, fill opacity=0.8]
table {%
0.025 0.0593132301796125
0.05 0.0832415843889224
0.1 0.117335968874343
0.15 0.143316226710202
0.2 0.165035458283249
0.3 0.201814632240018
0.4 0.233523889143544
0.5 0.261835289489214
0.6 0.287087265261532
0.7 0.311168788365674
0.8 0.33125742493894
0.85 0.334500101544674
0.9 0.333972904040995
0.95 0.319268086214755
0.975 0.29774652389324
};
\end{axis}

\end{tikzpicture}
	\end{minipage}%
	\begin{minipage}{0.245\textwidth}
		\begin{tikzpicture}

\definecolor{chocolate2196855}{RGB}{219,68,55}
\definecolor{darkgrey176}{RGB}{176,176,176}
\definecolor{darkorange2551094}{RGB}{255,109,4}
\definecolor{gainsboro229}{RGB}{229,229,229}
\definecolor{lightgrey204}{RGB}{204,204,204}
\definecolor{royalblue66133244}{RGB}{66,133,244}
\definecolor{seagreen1515788}{RGB}{15,157,88}

\definecolor{gainsboro247}{RGB}{253,253,253}
\definecolor{gainsboro229}{RGB}{239,239,239}
\definecolor{mediumpurple152142213}{RGB}{152,142,213}
\definecolor{gray119}{RGB}{119,119,119}
\definecolor{sandybrown25119394}{RGB}{251,193,94}

\definecolor{yolo11x}{RGB}{68,    3,   87} 
\definecolor{yolo11l}{RGB}{52,   71,  113.5} 
\definecolor{yolo11m}{RGB}{36,  139,  140} 
\definecolor{yolo11s}{RGB}{140.5, 184.5,  89.5} 
\definecolor{yolo11n}{RGB}{245,  230,   39} 	

\definecolor{yolow}{RGB}{255,109,4}
\definecolor{yoloe}{RGB}{0,48,73}
\definecolor{gdino}{RGB}{68,137,89} 

\tikzstyle{every node}=[font=\scriptsize]

\begin{axis}[
legend cell align={left},
legend style={
  fill opacity=0,
  draw opacity=0,
  text opacity=1,
  at={(0.025,0.025)},
  anchor=south west,
  draw=lightgrey204,
  fill=gainsboro247
},
tick align=outside,
tick pos=left,
x grid style={darkgrey176},
xmajorgrids,
xmin=0, xmax=1,
y grid style={darkgrey176},
ymajorgrids,
ymin=0, ymax=1,
axis background/.style={fill=gainsboro247},
axis line style={gainsboro229},
y grid style={gainsboro229},
x grid style={gainsboro229},
axis background/.style={fill=gainsboro247},
ticklabel style={font=\scriptsize},
xtick style={draw=none},
ytick style={draw=none},
ytick align=inside,
width=5.25cm,
height=3.5cm,
y label style={at={(axis description cs:0.19, 0.5)}},
yticklabel style = {xshift=0.5ex},
xticklabel style = {yshift=1.25ex},
ticklabel style={font=\tiny},
xminorticks=true,
minor x tick num=1,
grid=both,
yminorticks=true,
minor y tick num=1,
]
\addplot [line width=1.25pt,  yoloe, mark=none, mark options={solid,rotate=180}, mark size=0.5, opacity=0.75, fill opacity=0.8]
table {%
	0.025 0.328605766770409
	0.05 0.296329771780884
	0.1 0.259037003036676
	0.15 0.234191690830689
	0.2 0.214884355176874
	0.3 0.184087725831582
	0.4 0.157405606787918
	0.5 0.13269571283143
	0.6 0.108382625838589
	0.7 0.083659347520232
	0.8 0.0568008698260584
	0.85 0.0416560707135203
	0.9 0.0250279693797293
	0.95 0.00706221917443681
	0.975 0.000700710709615247
};
\addplot [line width=1.25pt, yolow, mark=none, mark options={solid,rotate=180}, mark size=0.5, opacity=0.75, fill opacity=0.8]
table {%
	0.025 0.341952337500069
	0.05 0.30965429767923
	0.1 0.270038948493039
	0.15 0.242974194163001
	0.2 0.221057695260125
	0.3 0.185402542556366
	0.4 0.154825574208807
	0.5 0.125757889490835
	0.6 0.0960005164789051
	0.7 0.0654125257156894
	0.8 0.0353449504345587
	0.85 0.0220448313137381
	0.9 0.011237352762178
	0.95 0.00386020742611635
	0.975 0.00146676859776765
};
\end{axis}

\end{tikzpicture}
	\end{minipage}%
	\begin{minipage}{0.245\textwidth}
		\begin{tikzpicture}

\definecolor{chocolate2196855}{RGB}{219,68,55}
\definecolor{darkgrey176}{RGB}{176,176,176}
\definecolor{darkorange2551094}{RGB}{255,109,4}
\definecolor{gainsboro229}{RGB}{229,229,229}
\definecolor{lightgrey204}{RGB}{204,204,204}
\definecolor{royalblue66133244}{RGB}{66,133,244}
\definecolor{seagreen1515788}{RGB}{15,157,88}

\definecolor{gainsboro247}{RGB}{253,253,253}
\definecolor{gainsboro229}{RGB}{239,239,239}
\definecolor{mediumpurple152142213}{RGB}{152,142,213}
\definecolor{gray119}{RGB}{119,119,119}
\definecolor{sandybrown25119394}{RGB}{251,193,94}

\definecolor{yolo11x}{RGB}{68,    3,   87} 
\definecolor{yolo11l}{RGB}{52,   71,  113.5} 
\definecolor{yolo11m}{RGB}{36,  139,  140} 
\definecolor{yolo11s}{RGB}{140.5, 184.5,  89.5} 
\definecolor{yolo11n}{RGB}{245,  230,   39} 	

\definecolor{yolow}{RGB}{255,109,4}
\definecolor{yoloe}{RGB}{0,48,73}
\definecolor{gdino}{RGB}{68,137,89} 

\tikzstyle{every node}=[font=\scriptsize]

\begin{axis}[
legend cell align={left},
legend style={
  fill opacity=0,
  draw opacity=0,
  text opacity=1,
  at={(0.075,0.075)},
  anchor=south west,
  draw=lightgrey204,
  fill=gainsboro247
},
tick align=outside,
tick pos=left,
x grid style={darkgrey176},
xmajorgrids,
xmin=0, xmax=1,
y grid style={darkgrey176},
ymajorgrids,
ymin=0, ymax=1,
axis background/.style={fill=gainsboro247},
axis line style={gainsboro229},
y grid style={gainsboro229},
x grid style={gainsboro229},
axis background/.style={fill=gainsboro247},
ticklabel style={font=\tiny},
xtick style={draw=none},
ytick style={draw=none},
ytick align=inside,
width=5.25cm,
height=3.5cm,
y label style={at={(axis description cs:0.19, 0.5)}},
x label style={at={(axis description cs:0.5, 0.0725)}},
yticklabel style = {xshift=0.5ex},
xticklabel style = {yshift=1.25ex},
xminorticks=true,
minor x tick num=1,
grid=both,
yminorticks=true,
minor y tick num=1,
]
\addplot [line width=1.25pt,  yoloe, mark=diamond*, mark options={solid,rotate=180}, mark size=1, opacity=0.75, fill opacity=0.8]
table {%
	0.3 0.214536399261931
};
\addplot [line width=1.25pt,  yolow, mark=triangle*, mark options={solid,rotate=0}, mark size=1, opacity=0.75, fill opacity=0.8]
table {%
	0.3 0.193260776524454
};
\addplot [line width=1.25pt,  yoloe, mark=none, mark options={solid,rotate=180}, mark size=0.5, opacity=0.75, fill opacity=0.8]
table {%
	0.025 0.1244465687036
	0.05 0.157055741059573
	0.1 0.188784302947243
	0.15 0.203823611336188
	0.2 0.211400784930937
	0.3 0.214536399261931
	0.4 0.206973624002799
	0.5 0.19185354920301
	0.6 0.169701071379702
	0.7 0.140531926149288
	0.8 0.101935783776605
	0.85 0.077271734820476
	0.9 0.0480062218262268
	0.95 0.0139762077266459
	0.975 0.00140018894684552
};
\addplot [line width=1.25pt, yolow, mark=none, mark options={solid,rotate=180}, mark size=0.5, opacity=0.75, fill opacity=0.8]
table {%
	0.025 0.101091642733666
	0.05 0.131210916316944
	0.1 0.163589743345151
	0.15 0.180289972589423
	0.2 0.18898215474414
	0.3 0.193260776524454
	0.4 0.18620069622869
	0.5 0.169909354375301
	0.6 0.143886216440582
	0.7 0.108100617899973
	0.8 0.0638745302924885
	0.85 0.0413636410640501
	0.9 0.0217431044516009
	0.95 0.00762818398501434
	0.975 0.00291915675072626
};
\end{axis}

\end{tikzpicture}
	\end{minipage}%
	\vspace{-1.65em}
	\\
	\begin{minipage}{0.265\textwidth}
		\begin{tikzpicture}

\definecolor{chocolate2196855}{RGB}{219,68,55}
\definecolor{darkgrey176}{RGB}{176,176,176}
\definecolor{darkorange2551094}{RGB}{255,109,4}
\definecolor{gainsboro229}{RGB}{229,229,229}
\definecolor{lightgrey204}{RGB}{204,204,204}
\definecolor{royalblue66133244}{RGB}{66,133,244}
\definecolor{seagreen1515788}{RGB}{15,157,88}

\definecolor{gainsboro247}{RGB}{253,253,253}
\definecolor{gainsboro229}{RGB}{239,239,239}
\definecolor{gainsboro245}{RGB}{247,247,247}
\definecolor{mediumpurple152142213}{RGB}{152,142,213}
\definecolor{gray119}{RGB}{119,119,119}
\definecolor{sandybrown25119394}{RGB}{251,193,94}

\definecolor{yolo11x}{RGB}{68,    3,   87} 
\definecolor{yolo11l}{RGB}{52,   71,  113.5} 
\definecolor{yolo11m}{RGB}{36,  139,  140} 
\definecolor{yolo11s}{RGB}{140.5, 184.5,  89.5} 
\definecolor{yolo11n}{RGB}{245,  230,   39} 	

\definecolor{yolow}{RGB}{255,109,4}
\definecolor{yoloe}{RGB}{0,48,73}
\definecolor{gdino}{RGB}{68,137,89} 

\tikzstyle{every node}=[font=\scriptsize]

\begin{axis}[
legend cell align={left},
legend style={
  fill opacity=0,
  draw opacity=0,
  text opacity=1,
  at={(0.0, -0.045)},
  anchor=south west,
  draw=lightgrey204,
  fill=gainsboro247
},
tick align=outside,
tick pos=left,
xlabel={\textcolor{white}{A}},
xmajorgrids,
xmin=0, xmax=1,
ylabel={\bf BDD},
ymajorgrids,
ymin=1, ymax=20000000,
axis line style={gainsboro229},
y grid style={gainsboro229},
x grid style={gainsboro229},
axis background/.style={fill=gainsboro247},
ticklabel style={font=\scriptsize},
xtick style={draw=none},
ytick style={draw=none},
ytick align=inside,
width=5.25cm,
height=3.5cm,
title style={at={(0.5,1.125)},anchor=north},
y label style={at={(axis description cs:0.165, 0.5)}},
x label style={at={(axis description cs:0.5, 0.32)}},
yticklabel style = {xshift=0.5ex},
xticklabel style = {yshift=1.25ex},
ymode = log,
ytick={1, 10, 100, 1000, 10000, 100000, 1000000, 1300000, 10000000},
yticklabels={\tiny $\text{1}$, \tiny $\text{10}$, \tiny $\text{100}$, \tiny $\text{1,000}$, \tiny $\text{10\rm{K}}$, \tiny $\text{100\rm{K}}$, , \tiny $\text{\textcolor{gray119}{1.3\rm{M}}}$, \tiny  $\text{10\rm{M}}$},
ticklabel style={font=\tiny},
xminorticks=true,
minor x tick num=1,
grid=both,
x label style={at={(axis description cs:0.5, 0.18)}},
]
\addplot [line width=1pt, gray119, mark=none, dashed, mark options={solid,rotate=180}, mark size=0.5, opacity=0.5, fill opacity=0.8]
table {%
	0 1286871
	0.05 1286871
	0.1 1286871
	0.15 1286871
	0.2 1286871
	0.3 1286871
	0.4 1286871
	0.5 1286871
	0.6 1286871
	0.7 1286871
	0.8 1286871
	0.85 1286871
	0.9 1286871
	0.95 1286871
	1 1286871
};
\addlegendentry{\tiny Human Labels}
\addplot [line width=1.25pt,  gdino, mark=square*, mark options={solid,rotate=180}, mark size=1, opacity=0.75, fill opacity=0.8]
table {%
	0.2 1381568
};
\addlegendentry{\tiny GDINO}
\addplot [line width=1.25pt,  yoloe, mark=diamond*, mark options={solid,rotate=180}, mark size=1, opacity=0.75, fill opacity=0.8]
table {%
	0.025 1527717
};
\addlegendentry{\tiny YOLOE}
\addplot [line width=1.25pt,  yolow, mark=triangle*, mark options={solid,rotate=0}, mark size=1, opacity=0.75, fill opacity=0.8]
table {%
	0.025 1288490
};
\addlegendentry{\tiny YOLOW}
\addplot [line width=1.25pt, gdino, mark=none, mark options={solid,rotate=180}, mark size=0.5, opacity=0.75, fill opacity=0.8]
table {%
	0.025 11119404
	0.05 7834320
	0.1 3618438
	0.15 2086884
	0.2 1381568
	0.3 734523
	0.4 395748
	0.5 190577
	0.6 66357
	0.7 9774
	0.8 655
	0.85 78
};
\addplot [line width=1.25pt,  yoloe, mark=none, mark options={solid,rotate=180}, mark size=0.5, opacity=0.75, fill opacity=0.8]
table {%
	0.025 1527717
	0.05 1039046
	0.1 701344
	0.15 547150
	0.2 449403
	0.3 323098
	0.4 236798
	0.5 170489
	0.6 115305
	0.7 67621
	0.8 27507
	0.85 12799
	0.9 4156
	0.95 450
	0.975 27
};
\addplot [line width=1.25pt, yolow, mark=none, mark options={solid,rotate=180}, mark size=0.5, opacity=0.75, fill opacity=0.8]
table {%
	0.025 1288490
	0.05 900512
	0.1 622133
	0.15 488451
	0.2 401485
	0.3 286144
	0.4 208290
	0.5 148782
	0.6 98876
	0.7 56083
	0.8 21053
	0.85 9423
	0.9 2652
	0.95 104
	0.975 2
};
\addplot [line width=1.25pt, gdino, dashed, mark=none, mark options={solid,rotate=180}, mark size=0.5, opacity=0.5, fill opacity=0.8]
table {%
	0.85 78
	0.85 0.1
};
\end{axis}

\end{tikzpicture}
	\end{minipage}%
	\begin{minipage}{0.245\textwidth}
		\begin{tikzpicture}

\definecolor{chocolate2196855}{RGB}{219,68,55}
\definecolor{darkgrey176}{RGB}{176,176,176}
\definecolor{darkorange2551094}{RGB}{255,109,4}
\definecolor{gainsboro229}{RGB}{229,229,229}
\definecolor{lightgrey204}{RGB}{204,204,204}
\definecolor{royalblue66133244}{RGB}{66,133,244}
\definecolor{seagreen1515788}{RGB}{15,157,88}

\definecolor{gainsboro247}{RGB}{253,253,253}
\definecolor{gainsboro229}{RGB}{239,239,239}
\definecolor{mediumpurple152142213}{RGB}{152,142,213}
\definecolor{gray119}{RGB}{119,119,119}
\definecolor{sandybrown25119394}{RGB}{251,193,94}

\definecolor{yolo11x}{RGB}{68,    3,   87} 
\definecolor{yolo11l}{RGB}{52,   71,  113.5} 
\definecolor{yolo11m}{RGB}{36,  139,  140} 
\definecolor{yolo11s}{RGB}{140.5, 184.5,  89.5} 
\definecolor{yolo11n}{RGB}{245,  230,   39} 	

\definecolor{yolow}{RGB}{255,109,4}
\definecolor{yoloe}{RGB}{0,48,73}
\definecolor{gdino}{RGB}{68,137,89} 

\tikzstyle{every node}=[font=\scriptsize]

\begin{axis}[
legend cell align={left},
legend style={
  fill opacity=0,
  draw opacity=0,
  text opacity=1,
  at={(0.075,0.075)},
  anchor=south west,
  draw=lightgrey204,
  fill=gainsboro247
},
tick align=outside,
tick pos=left,
x grid style={darkgrey176},
xlabel={\textcolor{white}{A}},
xmajorgrids,
xmin=0, xmax=1,
y grid style={darkgrey176},
ymajorgrids,
ymin=0, ymax=1,
axis background/.style={fill=gainsboro247},
axis line style={gainsboro229},
y grid style={gainsboro229},
x grid style={gainsboro229},
axis background/.style={fill=gainsboro247},
ticklabel style={font=\scriptsize},
xtick style={draw=none},
ytick style={draw=none},
ytick align=inside,
width=5.25cm,
height=3.5cm,
x label style={at={(axis description cs:0.5, 0.0725)}},
yticklabel style = {xshift=0.5ex},
xticklabel style = {yshift=1.25ex},
ticklabel style={font=\tiny},
xminorticks=true,
minor x tick num=1,
yminorticks=true,
minor y tick num=1,
grid=both,
xlabel={\bf Confidence Threshold},
x label style={at={(axis description cs:1.05, 0.15)}},
]
\addplot [line width=1.25pt, gdino, mark=none, mark options={solid,rotate=180}, mark size=0.5, opacity=0.75, fill opacity=0.8]
table {
	0.05 0.0751349191761378
	0.1 0.199159416300625
	0.15 0.343773300288852
	0.2 0.477710109093436
	0.3 0.684422407467159
	0.4 0.805312976944925
	0.5 0.866673313149016
	0.6 0.890591798905918
	0.7 0.925618989154901
	0.8 0.963358778625954
	0.85 1
};
\addplot [line width=1.25pt, gdino, dashed, mark=none, mark options={solid,rotate=180}, mark size=0.5, opacity=0.5, fill opacity=0.8]
table {%
	0.85 1
	0.85 0
};
\addplot [line width=1.25pt,  yoloe, mark=none, mark options={solid,rotate=180}, mark size=0.5, opacity=0.75, fill opacity=0.8]
table {%
	0.025 0.408194711455067
	0.05 0.531437491699116
	0.1 0.654691563626409
	0.15 0.721584574613908
	0.2 0.765564537842427
	0.3 0.821546403877461
	0.4 0.859432934399784
	0.5 0.88855585990885
	0.6 0.912588352629981
	0.7 0.931441416128126
	0.8 0.93561638855564
	0.85 0.930385186342683
	0.9 0.919874879692012
	0.95 0.931111111111111
	0.975 1
};
\addplot [line width=1.25pt, yolow, mark=none, mark options={solid,rotate=180}, mark size=0.5, opacity=0.75, fill opacity=0.8]
table {%
	0.025 0.48965998960023
	0.05 0.605618803525106
	0.1 0.712389472990502
	0.15 0.767591836233317
	0.2 0.803957806642838
	0.3 0.851843826884366
	0.4 0.884555187478996
	0.5 0.909847965479695
	0.6 0.927464703264695
	0.7 0.939589536936327
	0.8 0.944900964233126
	0.85 0.943011779687997
	0.9 0.950980392156863
	0.95 0.980769230769231
	0.975 0.5
};
\end{axis}

\end{tikzpicture}
	\end{minipage}%
	\begin{minipage}{0.245\textwidth}
		\begin{tikzpicture}

\definecolor{chocolate2196855}{RGB}{219,68,55}
\definecolor{darkgrey176}{RGB}{176,176,176}
\definecolor{darkorange2551094}{RGB}{255,109,4}
\definecolor{gainsboro229}{RGB}{229,229,229}
\definecolor{lightgrey204}{RGB}{204,204,204}
\definecolor{royalblue66133244}{RGB}{66,133,244}
\definecolor{seagreen1515788}{RGB}{15,157,88}

\definecolor{gainsboro247}{RGB}{253,253,253}
\definecolor{gainsboro229}{RGB}{239,239,239}
\definecolor{mediumpurple152142213}{RGB}{152,142,213}
\definecolor{gray119}{RGB}{119,119,119}
\definecolor{sandybrown25119394}{RGB}{251,193,94}

\definecolor{yolo11x}{RGB}{68,    3,   87} 
\definecolor{yolo11l}{RGB}{52,   71,  113.5} 
\definecolor{yolo11m}{RGB}{36,  139,  140} 
\definecolor{yolo11s}{RGB}{140.5, 184.5,  89.5} 
\definecolor{yolo11n}{RGB}{245,  230,   39} 	

\definecolor{yolow}{RGB}{255,109,4}
\definecolor{yoloe}{RGB}{0,48,73}
\definecolor{gdino}{RGB}{68,137,89} 

\tikzstyle{every node}=[font=\scriptsize]

\begin{axis}[
legend cell align={left},
legend style={
  fill opacity=0,
  draw opacity=0,
  text opacity=1,
  at={(0.075,0.075)},
  anchor=south west,
  draw=lightgrey204,
  fill=gainsboro247
},
tick align=outside,
tick pos=left,
x grid style={darkgrey176},
xlabel={\textcolor{white}{A}},
xmajorgrids,
xmin=0, xmax=1,
y grid style={darkgrey176},
ymajorgrids,
ymin=0, ymax=1,
axis background/.style={fill=gainsboro247},
axis line style={gainsboro229},
y grid style={gainsboro229},
x grid style={gainsboro229},
axis background/.style={fill=gainsboro247},
ticklabel style={font=\scriptsize},
xtick style={draw=none},
ytick style={draw=none},
ytick align=inside,
width=5.25cm,
height=3.5cm,
y label style={at={(axis description cs:0.19, 0.5)}},
yticklabel style = {xshift=0.5ex},
xticklabel style = {yshift=1.25ex},
ticklabel style={font=\tiny},
xminorticks=true,
minor x tick num=1,
grid=both,
yminorticks=true,
minor y tick num=1,
x label style={at={(axis description cs:0.5, 0.15)}},
]
\addplot [line width=1.25pt, gdino, mark=none, mark options={solid,rotate=180}, mark size=0.5, opacity=0.75, fill opacity=0.8]
table {%
	0.025 0.291091337049323
	0.05 0.457412592248951
	0.1 0.559998632341548
	0.15 0.557487891171687
	0.2 0.512863371697707
	0.3 0.390656095288494
	0.4 0.247655747934331
	0.5 0.128348529106647
	0.6 0.0459230179248736
	0.7 0.00703023069134358
	0.8 0.0004903366382489
	0.85 6.06121359483585e-05
	0.9 0
	0.95 0
	0.975 0
};
\addplot [line width=1.25pt,  yoloe, mark=none, mark options={solid,rotate=180}, mark size=0.5, opacity=0.75, fill opacity=0.8]
table {%
	0.025 0.484590918592462
	0.05 0.429093514423746
	0.1 0.356806548597334
	0.15 0.306802313518604
	0.2 0.267351583802883
	0.3 0.20626776110426
	0.4 0.158144833475927
	0.5 0.11771887003437
	0.6 0.0817688797089996
	0.7 0.0489442997782995
	0.8 0.0199988965482943
	0.85 0.00925345275478272
	0.9 0.00297077174013557
	0.95 0.000325595961055926
	0.975 2.09811239821241e-05
};
\addplot [line width=1.25pt, yolow, mark=none, mark options={solid,rotate=180}, mark size=0.5, opacity=0.75, fill opacity=0.8]
table {%
	0.025 0.490276026112952
	0.05 0.423793060842928
	0.1 0.344402041851903
	0.15 0.291350881323769
	0.2 0.250823120576965
	0.3 0.18941292483862
	0.4 0.143172081739351
	0.5 0.105192361938376
	0.6 0.0712612219872854
	0.7 0.0409481603051122
	0.8 0.0154584259028294
	0.85 0.00690512102611684
	0.9 0.00195979239566359
	0.95 7.92620239324688e-05
	0.975 7.77078666004596e-07
};
\end{axis}

\end{tikzpicture}
	\end{minipage}%
	\begin{minipage}{0.245\textwidth}
		\begin{tikzpicture}

\definecolor{chocolate2196855}{RGB}{219,68,55}
\definecolor{darkgrey176}{RGB}{176,176,176}
\definecolor{darkorange2551094}{RGB}{255,109,4}
\definecolor{gainsboro229}{RGB}{229,229,229}
\definecolor{lightgrey204}{RGB}{204,204,204}
\definecolor{royalblue66133244}{RGB}{66,133,244}
\definecolor{seagreen1515788}{RGB}{15,157,88}

\definecolor{gainsboro247}{RGB}{253,253,253}
\definecolor{gainsboro229}{RGB}{239,239,239}
\definecolor{mediumpurple152142213}{RGB}{152,142,213}
\definecolor{gray119}{RGB}{119,119,119}
\definecolor{sandybrown25119394}{RGB}{251,193,94}

\definecolor{yolo11x}{RGB}{68,    3,   87} 
\definecolor{yolo11l}{RGB}{52,   71,  113.5} 
\definecolor{yolo11m}{RGB}{36,  139,  140} 
\definecolor{yolo11s}{RGB}{140.5, 184.5,  89.5} 
\definecolor{yolo11n}{RGB}{245,  230,   39} 	

\definecolor{yolow}{RGB}{255,109,4}
\definecolor{yoloe}{RGB}{0,48,73}
\definecolor{gdino}{RGB}{68,137,89} 

\tikzstyle{every node}=[font=\scriptsize]

\begin{axis}[
legend cell align={left},
legend style={
  fill opacity=0,
  draw opacity=0,
  text opacity=1,
  at={(0.075,0.075)},
  anchor=south west,
  draw=lightgrey204,
  fill=gainsboro247
},
tick align=outside,
tick pos=left,
x grid style={darkgrey176},
xlabel={\textcolor{white}{A}},
xmajorgrids,
xmin=0, xmax=1,
y grid style={darkgrey176},
ymajorgrids,
ymin=0, ymax=1,
axis background/.style={fill=gainsboro247},
axis line style={gainsboro229},
y grid style={gainsboro229},
x grid style={gainsboro229},
axis background/.style={fill=gainsboro247},
ticklabel style={font=\tiny},
xtick style={draw=none},
ytick style={draw=none},
ytick align=inside,
width=5.25cm,
height=3.5cm,
y label style={at={(axis description cs:0.19, 0.5)}},
x label style={at={(axis description cs:0.5, 0.0725)}},
yticklabel style = {xshift=0.5ex},
xticklabel style = {yshift=1.25ex},
xminorticks=true,
minor x tick num=1,
grid=both,
yminorticks=true,
minor y tick num=1,
x label style={at={(axis description cs:0.5, 0.15)}},
]
\addplot [line width=1.25pt,  gdino, mark=square*, mark options={solid,rotate=180}, mark size=1, opacity=0.75, fill opacity=0.8]
table {%
	0.3 0.497403277144386
};
\addplot [line width=1.25pt,  yoloe, mark=diamond*, mark options={solid,rotate=180}, mark size=1, opacity=0.75, fill opacity=0.8]
table {%
	0.05 0.474813159712922
};
\addplot [line width=1.25pt,  yolow, mark=triangle*, mark options={solid,rotate=0}, mark size=1, opacity=0.75, fill opacity=0.8]
table {%
	0.05 0.498647927683446
};
\addplot [line width=1.25pt, gdino, mark=none, mark options={solid,rotate=180}, mark size=0.5, opacity=0.75, fill opacity=0.8]
table {%
	0.025 0.0603883115600775
	0.05 0.129068890235935
	0.1 0.293822876397797
	0.15 0.425291700197554
	0.2 0.494662984613851
	0.3 0.497403277144386
	0.4 0.378815406220897
	0.5 0.2235855339748
	0.6 0.0873422660482934
	0.7 0.0139544748177026
	0.8 0.00098017438094454
	0.85 0.000121216924679999
	0.9 0
	0.95 0
	0.975 0
};
\addplot [line width=1.25pt,  yoloe, mark=none, mark options={solid,rotate=180}, mark size=0.5, opacity=0.75, fill opacity=0.8]
table {%
	0.025 0.443124180164202
	0.05 0.474813159712922
	0.1 0.461885661258968
	0.15 0.430545778919653
	0.2 0.396304961083331
	0.3 0.329745479571346
	0.4 0.267134134775991
	0.5 0.207895097985398
	0.6 0.150089575060477
	0.7 0.0930016567096742
	0.8 0.039160728496673
	0.85 0.01832465164234
	0.9 0.00592241680460595
	0.95 0.000650964289404119
	0.975 4.19613675675928e-05
};
\addplot [line width=1.25pt, yolow, mark=none, mark options={solid,rotate=180}, mark size=0.5, opacity=0.75, fill opacity=0.8]
table {%
	0.025 0.489967814220997
	0.05 0.498647927683446
	0.1 0.464326947455322
	0.15 0.422380841334699
	0.2 0.382356564610781
	0.3 0.309914400053401
	0.4 0.246453726387994
	0.5 0.188581781252155
	0.6 0.13235316403355
	0.7 0.0784762545850416
	0.8 0.0304191986690358
	0.85 0.0137098528574536
	0.9 0.00391152387355635
	0.95 0.000158511237592028
	0.975 1.55415491660793e-06
};
\end{axis}

\end{tikzpicture}
	\end{minipage}%
	\vspace{-2.0em}
	\caption{{\bf Human Label-based Evaluation of Auto-Labels across all Confidence Thresholds and Datasets}.
		For number of object labels and $F_1$ score, the AL results closest to human labels are individually marked.
		For number of object labels and precision, dashed vertical lines indicate where a final confidence threshold ($\alpha$) precedes zero label generation thereafter.
		The number of object labels are in log scale.
	}
	\label{fig:f1}
\end{figure*}
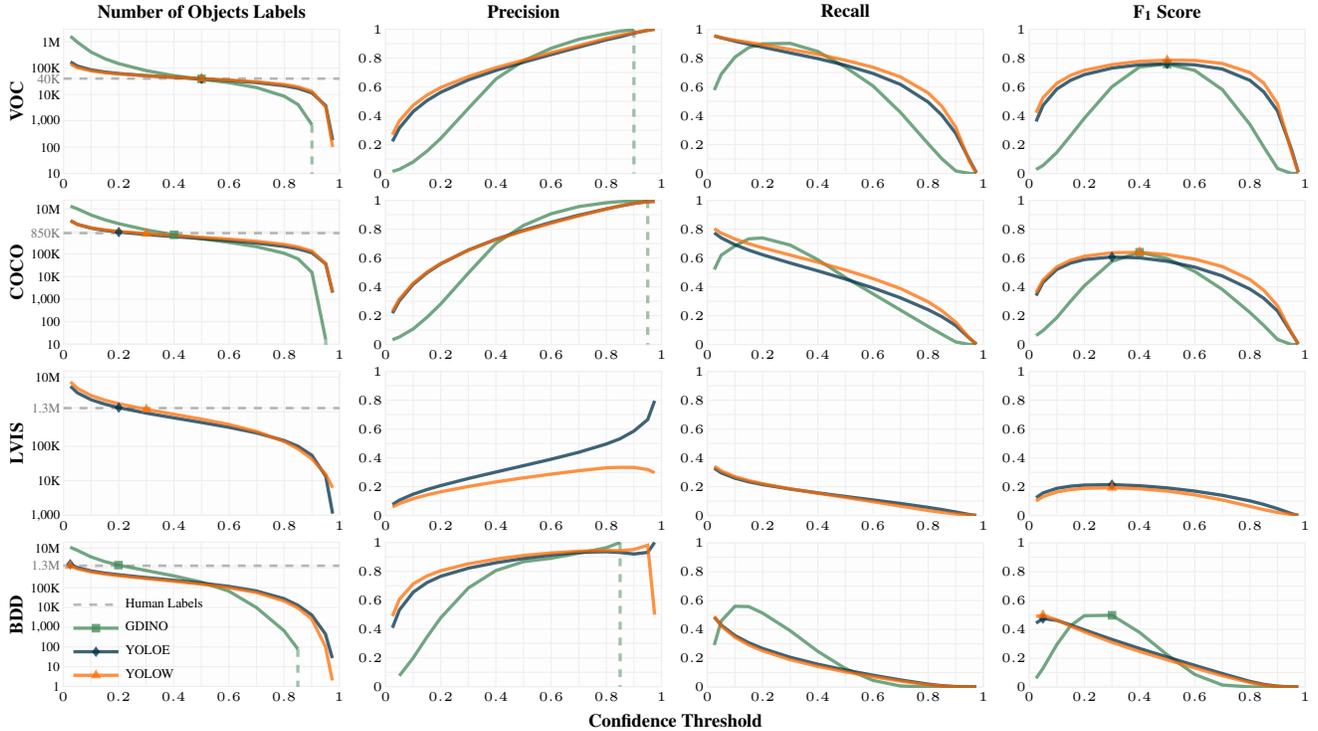

We compare estimated costs for auto-labeling, human labeling, and using annotation services for each dataset train set in \cref{tab:cost}.
Human labeling time is based on the number of object instances and an estimated 7 seconds per bounding box for annotation \cite{JaGr13}.
Annotation service cost is based on AWS SageMaker's price of \$0.036 per bounding box, which is currently among the least expensive annotation services.
Finally, auto-labeling cost is based on the time YOLOW ($\alpha=0.2$) takes to generate labels for an entire train set on a single NVIDIA L40S GPU and the cost of renting that GPU, which is currently \$0.93 per hour at the high range ({\footnotesize \url{https://vast.ai/pricing/gpu/L40S}}).

Using compute-based AL in place of human labeling or an annotation service results in a substantial time and cost reduction.
Overall, AL approximately takes ${\tiny \frac{\text{\scriptsize 1}}{\text{\scriptsize 5\rm{K}}}}$ the time of human labeling at ${\tiny \frac{\text{\scriptsize 1}}{\text{\scriptsize 100\rm{K}}}}$ the cost of an annotation service. 

\subsection{Auto-Label Evaluation with Human Labels}
\label{sec:label}
We compare auto-labels to human labels across all AL models, confidence thresholds ($\alpha$), and datasets.
Our evaluation metrics include the relative number of object labels, precision, recall, and $F_1$ scores, which we compare in \cref{fig:f1}.

In general, all AL models can generate a higher, equal, or lower number of labels than humans by varying confidence threshold $\alpha$ (\cref{fig:f1}, left).
In terms of accuracy, at lower $\alpha$ values, AL models generate more labels and recall is higher (column 3).
On the other hand, at higher $\alpha$ values, AL models generate less labels but those labels have a higher precision (column 2).

Our primary metric to evaluate AL relative to human labels is the $F_1$ score.
The $F_1$ score is the harmonic mean of precision and recall (\cref{eq:f1}) and correspondingly peaks at $\alpha$ settings between highest precision and recall.
In \cref{fig:f1} (right), all $F_1$ peaks occur at mid-to-low $\alpha$ values with an average of 0.33, which particularly emphasizes $\alpha$ selection for high recall near the expected number of overall labels.
On the other hand, high $\alpha$ values achieve the objective of precise labels but with increasingly fewer labels being generated.
Thus, although practitioners will intuitively want high precision, we {\bf caution against using high confidence thresholds that leave too many objects unlabeled}.

In terms of relative {\bf dataset auto-labeling difficulty}, VOC has the highest $F_1$ scores (\cref{fig:f1}, top) followed by COCO (row 2), BDD (bottom), and LVIS (row 3).
These increases in AL difficulty make intuitive sense.
COCO has 4$\times$ the number of classes as VOC, including more nuanced class labels like ``hair drier."
BDD consists entirely of autonomous driving viewpoint images that are not representative of the original AL model training data distribution (\cref{tab:fa}).
Finally, LVIS has 60$\times$ the number of classes as VOC, including rare classes like ``eye dropper" and ``car battery" that make high recall particularly challenging.

In terms of relative {\bf performance between AL models}, each model achieves a top $F_1$ score on at least one dataset (\cref{fig:f1}, right). 
For VOC, YOLOW-0.5 ($\alpha=0.5$) has the highest $F_1$ score of 0.785.
For COCO, YOLOW-0.4 \& GDINO-0.4 share the highest $F_1$ score of 0.640.
Finally, for the more challenging datasets, YOLOE-0.3 has the highest $F_1$ score of 0.215 on LVIS and YOLOW-0.05 has the highest $F_1$ score of 0.499 on BDD.
Notably, the peak $F_1$ scores of all AL models are relatively close.

The relative {\bf consequence of changing confidence threshold} $\alpha$ varies dramatically between AL models.
GDINO has the greatest sensitivity to changes in $\alpha$, with dramatic increases and decreases to the number of labels at low and high $\alpha$ values respectively.
Consequently, YOLOW \& YOLOE both have a wider $\alpha$ range for peak $F_1$ performance.
Additionally, YOLOW \& YOLOE both have much higher precision and $F_1$ scores at lower $\alpha$ values, likely due to better internal non-maximum suppression of predicted labels.
In all cases, using high $\alpha$ causes a collapse of $F_1$ scores, which validates the significance of correctly configuring confidence threshold $\alpha$ for auto-labeling.

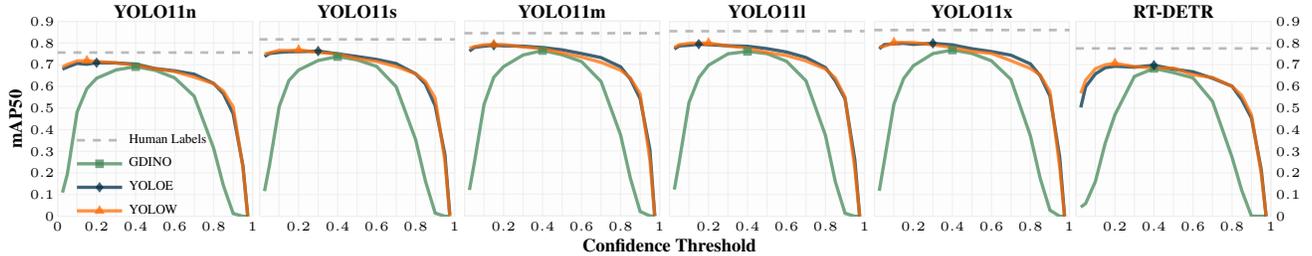
\begin{figure*}
	\centering
	\begin{minipage}{0.188\textwidth}
		\begin{tikzpicture}

\definecolor{chocolate2196855}{RGB}{219,68,55}
\definecolor{darkgrey176}{RGB}{176,176,176}
\definecolor{darkorange2551094}{RGB}{255,109,4}
\definecolor{gainsboro229}{RGB}{229,229,229}
\definecolor{lightgrey204}{RGB}{204,204,204}
\definecolor{royalblue66133244}{RGB}{66,133,244}
\definecolor{seagreen1515788}{RGB}{15,157,88}

\definecolor{gainsboro247}{RGB}{253,253,253}
\definecolor{gainsboro229}{RGB}{239,239,239}
\definecolor{mediumpurple152142213}{RGB}{152,142,213}
\definecolor{gray119}{RGB}{119,119,119}
\definecolor{sandybrown25119394}{RGB}{251,193,94}

\definecolor{yolo11x}{RGB}{68,    3,   87} 
\definecolor{yolo11l}{RGB}{52,   71,  113.5} 
\definecolor{yolo11m}{RGB}{36,  139,  140} 
\definecolor{yolo11s}{RGB}{140.5, 184.5,  89.5} 
\definecolor{yolo11n}{RGB}{245,  230,   39} 	

\definecolor{yolow}{RGB}{255,109,4}
\definecolor{yoloe}{RGB}{0,48,73}
\definecolor{gdino}{RGB}{68,137,89} 

\tikzstyle{every node}=[font=\scriptsize]

\begin{axis}[
legend cell align={left},
legend style={
  fill opacity=0,
  draw opacity=0,
  text opacity=1,
  at={(0.0475,-0.0625)},
  anchor=south west,
  draw=lightgrey204,
  fill=gainsboro247
},
tick align=outside,
tick pos=left,
title={voc},
x grid style={darkgrey176},
xlabel={\textcolor{white}{A}},
xmajorgrids,
xmin=0, xmax=1,
y grid style={darkgrey176},
ylabel={\bf mAP50},
ymajorgrids,
ymin=0, ymax=0.9,
axis background/.style={fill=gainsboro247},
axis line style={gainsboro229},
y grid style={gainsboro229},
x grid style={gainsboro229},
axis background/.style={fill=gainsboro247},
ticklabel style={font=\tiny},
xtick style={draw=none},
ytick style={draw=none},
ytick align=inside,
width=4.175cm,
height=4.175cm,
title={\bf  YOLO11n},
title style={at={(0.5,1.05)},anchor=north},
y label style={at={(axis description cs:0.26, 0.5)}},
x label style={at={(axis description cs:0.5, 0.135)}},
yticklabel style = {xshift=0.5ex},
xticklabel style = {yshift=1.25ex},
xminorticks=true,
minor x tick num=1,
grid=both,
yminorticks=true,
minor y tick num=0,
ytick={0, 0.1, 0.2, 0.3, 0.4, 0.5, 0.6, 0.7, 0.8, 0.9},
xtick={0, 0.2, 0.4, 0.6, 0.8, 1},
xticklabels={$0$, $0.2$, $0.4$, $0.6$, $0.8$, $1$},
]
\addplot [line width=1pt, gray119, mark=none, dashed, mark options={solid,rotate=180}, mark size=0.5, opacity=0.5, fill opacity=0.8]
table {%
	0 0.755879815922295
	0.1 0.755879815922295
	0.2 0.755879815922295
	0.3 0.755879815922295
	0.4 0.755879815922295
	0.5 0.755879815922295
	0.6 0.755879815922295
	0.7 0.755879815922295
	0.8 0.755879815922295
	0.9 0.755879815922295
	1 0.755879815922295
};
\addlegendentry{\tiny Human Labels}
\addplot [line width=1.25pt,  gdino, mark=square*, mark options={solid,rotate=180}, mark size=1, opacity=0.75, fill opacity=0.8]
table {%
	0.4 0.690645309859582
};
\addlegendentry{\tiny GDINO}
\addplot [line width=1.25pt,  yoloe, mark=diamond*, mark options={solid,rotate=180}, mark size=1, opacity=0.75, fill opacity=0.8]
table {%
0.2 0.708910488535897
};
\addlegendentry{\tiny YOLOE}
\addplot [line width=1.25pt,  yolow, mark=triangle*, mark options={solid,rotate=0}, mark size=1, opacity=0.75, fill opacity=0.8]
table {%
	0.15 0.718341715363312
};
\addlegendentry{\tiny YOLOW}
\addplot [line width=1.25pt, gdino, mark=none, mark options={solid,rotate=180}, mark size=0.5, opacity=0.75, fill opacity=0.8]
table {%
	0.025 0.109184169382849
	0.05 0.195098835251349
	0.1 0.481442762647939
	0.15 0.589432604929476
	0.2 0.637121408992923
	0.3 0.675880822160058
	0.4 0.690645309859582
	0.5 0.673985946089972
	0.6 0.639110741096113
	0.7 0.554554610882226
	0.8 0.313876267838924
	0.85 0.141680433190635
	0.9 0.0127264878264296
	0.95 0
	0.975 0
};
\addplot [line width=1.25pt,  yoloe, mark=none, mark options={solid,rotate=180}, mark size=0.5, opacity=0.75, fill opacity=0.8]
table {%
0.025 0.679237321261571
0.05 0.688615279968037
0.1 0.705987967917236
0.15 0.701697444279976
0.2 0.708910488535897
0.3 0.70783178237348
0.4 0.702562289319267
0.5 0.682111151219109
0.6 0.671572563929806
0.7 0.656039170774121
0.8 0.613231101328802
0.85 0.565785733540322
0.9 0.473559162823561
0.95 0.233763355005946
0.975 0
};
\addplot [line width=1.25pt, yolow, mark=none, mark options={solid,rotate=180}, mark size=0.5, opacity=0.75, fill opacity=0.8]
table {%
	0.025 0.689162592372176
	0.05 0.700732479952618
	0.1 0.716887370116343
	0.15 0.718341715363312
	0.2 0.714931859211686
	0.3 0.708052002048271
	0.4 0.697266071322653
	0.5 0.680897374182605
	0.6 0.66814430725038
	0.7 0.64251981451702
	0.8 0.612140845232438
	0.85 0.577774706113686
	0.9 0.504632293294645
	0.95 0.225592062511523
	0.975 0
};
\end{axis}

\end{tikzpicture}
	\end{minipage}%
	\begin{minipage}{0.156\textwidth}
		\begin{tikzpicture}

\definecolor{chocolate2196855}{RGB}{219,68,55}
\definecolor{darkgrey176}{RGB}{176,176,176}
\definecolor{darkorange2551094}{RGB}{255,109,4}
\definecolor{gainsboro229}{RGB}{229,229,229}
\definecolor{lightgrey204}{RGB}{204,204,204}
\definecolor{royalblue66133244}{RGB}{66,133,244}
\definecolor{seagreen1515788}{RGB}{15,157,88}

\definecolor{gainsboro247}{RGB}{253,253,253}
\definecolor{gainsboro229}{RGB}{239,239,239}
\definecolor{mediumpurple152142213}{RGB}{152,142,213}
\definecolor{gray119}{RGB}{119,119,119}
\definecolor{sandybrown25119394}{RGB}{251,193,94}

\definecolor{yolo11x}{RGB}{68,    3,   87} 
\definecolor{yolo11l}{RGB}{52,   71,  113.5} 
\definecolor{yolo11m}{RGB}{36,  139,  140} 
\definecolor{yolo11s}{RGB}{140.5, 184.5,  89.5} 
\definecolor{yolo11n}{RGB}{245,  230,   39} 	

\definecolor{yolow}{RGB}{255,109,4}
\definecolor{yoloe}{RGB}{0,48,73}
\definecolor{gdino}{RGB}{68,137,89} 

\tikzstyle{every node}=[font=\scriptsize]

\begin{axis}[
legend cell align={left},
legend style={
  fill opacity=0,
  draw opacity=0,
  text opacity=1,
  at={(0.0475,-0.0375)},
  anchor=south west,
  draw=lightgrey204,
  fill=gainsboro247
},
tick align=outside,
tick pos=left,
title={voc},
x grid style={darkgrey176},
xlabel={\textcolor{white}{A}},
xmajorgrids,
xmin=0, xmax=1,
y grid style={darkgrey176},
ymajorgrids,
ymin=0, ymax=0.9,
axis background/.style={fill=gainsboro247},
axis line style={gainsboro229},
y grid style={gainsboro229},
x grid style={gainsboro229},
axis background/.style={fill=gainsboro247},
ticklabel style={font=\tiny},
xtick style={draw=none},
ytick style={draw=none},
ytick align=inside,
width=4.175cm,
height=4.175cm,
title={\bf  YOLO11s},
title style={at={(0.5,1.05)},anchor=north},
y label style={at={(axis description cs:0.26, 0.5)}},
x label style={at={(axis description cs:0.5, 0.135)}},
yticklabel style = {xshift=0.5ex},
xticklabel style = {yshift=1.25ex},
xminorticks=true,
minor x tick num=1,
grid=both,
yminorticks=true,
minor y tick num=0,
ytick={0, 0.1, 0.2, 0.3, 0.4, 0.5, 0.6, 0.7, 0.8, 0.9},
yticklabels={},
xtick={0, 0.2, 0.4, 0.6, 0.8, 1},
xticklabels={, $0.2$, $0.4$, $0.6$, $0.8$, $1$},
]
\addplot [line width=1pt, gray119, mark=none, dashed, mark options={solid,rotate=180}, mark size=0.5, opacity=0.5, fill opacity=0.8]
table {%
	0 0.816753624445951
	0.1 0.816753624445951
	0.2 0.816753624445951
	0.3 0.816753624445951
	0.4 0.816753624445951
	0.5 0.816753624445951
	0.6 0.816753624445951
	0.7 0.816753624445951
	0.8 0.816753624445951
	0.9 0.816753624445951
	1 0.816753624445951
};
\addplot [line width=1.25pt,  gdino, mark=square*, mark options={solid,rotate=180}, mark size=1, opacity=0.75, fill opacity=0.8]
table {%
	0.4 0.737171831430671
};
\addplot [line width=1.25pt,  yoloe, mark=diamond*, mark options={solid,rotate=180}, mark size=1, opacity=0.75, fill opacity=0.8]
table {%
	0.3 0.76270775665132
};
\addplot [line width=1.25pt,  yolow, mark=triangle*, mark options={solid,rotate=0}, mark size=1, opacity=0.75, fill opacity=0.8]
table {%
	0.2 0.768405647903957
};
\addplot [line width=1.25pt, gdino, mark=none, mark options={solid,rotate=180}, mark size=0.5, opacity=0.75, fill opacity=0.8]
table {%
	0.025 0.116167074410765
	0.05 0.232738027606369
	0.1 0.505316506369325
	0.15 0.626700246726936
	0.2 0.67517896089114
	0.3 0.71915442321461
	0.4 0.737171831430671
	0.5 0.721932442698093
	0.6 0.691842421224324
	0.7 0.597927158034434
	0.8 0.353013788287097
	0.85 0.162652653173332
	0.9 0.0143522571403489
	0.95 0
	0.975 0
};
\addplot [line width=1.25pt,  yoloe, mark=none, mark options={solid,rotate=180}, mark size=0.5, opacity=0.75, fill opacity=0.8]
table {%
	0.025 0.737372817739495
	0.05 0.751387521222863
	0.1 0.756725796875628
	0.15 0.759523982957058
	0.2 0.759999861119108
	0.3 0.76270775665132
	0.4 0.749999998157486
	0.5 0.737488963683421
	0.6 0.724019299979698
	0.7 0.705000166045537
	0.8 0.657545725074013
	0.85 0.610069385022936
	0.9 0.51020354327406
	0.95 0.283578413819228
	0.975 0.00531148916408669
};
\addplot [line width=1.25pt, yolow, mark=none, mark options={solid,rotate=180}, mark size=0.5, opacity=0.75, fill opacity=0.8]
table {%
	0.025 0.74919924835155
	0.05 0.754127507788968
	0.1 0.766191389631273
	0.15 0.76477211516791
	0.2 0.768405647903957
	0.3 0.756497375443248
	0.4 0.749217953173001
	0.5 0.727490701574232
	0.6 0.715050875523249
	0.7 0.691154945943099
	0.8 0.65792395158891
	0.85 0.622567684844788
	0.9 0.545084567986889
	0.95 0.254925936943606
	0.975 0
};
\end{axis}

\end{tikzpicture}
	\end{minipage}%
	\begin{minipage}{0.156\textwidth}
		\begin{tikzpicture}

\definecolor{chocolate2196855}{RGB}{219,68,55}
\definecolor{darkgrey176}{RGB}{176,176,176}
\definecolor{darkorange2551094}{RGB}{255,109,4}
\definecolor{gainsboro229}{RGB}{229,229,229}
\definecolor{lightgrey204}{RGB}{204,204,204}
\definecolor{royalblue66133244}{RGB}{66,133,244}
\definecolor{seagreen1515788}{RGB}{15,157,88}

\definecolor{gainsboro247}{RGB}{253,253,253}
\definecolor{gainsboro229}{RGB}{239,239,239}
\definecolor{mediumpurple152142213}{RGB}{152,142,213}
\definecolor{gray119}{RGB}{119,119,119}
\definecolor{sandybrown25119394}{RGB}{251,193,94}

\definecolor{yolo11x}{RGB}{68,    3,   87} 
\definecolor{yolo11l}{RGB}{52,   71,  113.5} 
\definecolor{yolo11m}{RGB}{36,  139,  140} 
\definecolor{yolo11s}{RGB}{140.5, 184.5,  89.5} 
\definecolor{yolo11n}{RGB}{245,  230,   39} 	

\definecolor{yolow}{RGB}{255,109,4}
\definecolor{yoloe}{RGB}{0,48,73}
\definecolor{gdino}{RGB}{68,137,89} 

\tikzstyle{every node}=[font=\scriptsize]

\begin{axis}[
legend cell align={left},
legend style={
  fill opacity=0,
  draw opacity=0,
  text opacity=1,
  at={(0.0475,-0.0375)},
  anchor=south west,
  draw=lightgrey204,
  fill=gainsboro247
},
tick align=outside,
tick pos=left,
title={voc},
x grid style={darkgrey176},
xlabel={\bf Confidence Threshold},
xmajorgrids,
xmin=0, xmax=1,
y grid style={darkgrey176},
ymajorgrids,
ymin=0, ymax=0.9,
axis background/.style={fill=gainsboro247},
axis line style={gainsboro229},
y grid style={gainsboro229},
x grid style={gainsboro229},
axis background/.style={fill=gainsboro247},
ticklabel style={font=\tiny},
xtick style={draw=none},
ytick style={draw=none},
ytick align=inside,
width=4.175cm,
height=4.175cm,
title={\bf  YOLO11m},
title style={at={(0.5,1.05)},anchor=north},
y label style={at={(axis description cs:0.26, 0.5)}},
x label style={at={(axis description cs:1.05, 0.135)}},
yticklabel style = {xshift=0.5ex},
xticklabel style = {yshift=1.25ex},
xminorticks=true,
minor x tick num=1,
grid=both,
yminorticks=true,
minor y tick num=0,
ytick={0, 0.1, 0.2, 0.3, 0.4, 0.5, 0.6, 0.7, 0.8, 0.9},
yticklabels={},
xtick={0, 0.2, 0.4, 0.6, 0.8, 1},
xticklabels={, $0.2$, $0.4$, $0.6$, $0.8$, $1$},
]
\addplot [line width=1pt, gray119, mark=none, dashed, mark options={solid,rotate=180}, mark size=0.5, opacity=0.5, fill opacity=0.8]
table {%
	0 0.843530101507674
	0.1 0.843530101507674
	0.2 0.843530101507674
	0.3 0.843530101507674
	0.4 0.843530101507674
	0.5 0.843530101507674
	0.6 0.843530101507674
	0.7 0.843530101507674
	0.8 0.843530101507674
	0.9 0.843530101507674
	1 0.843530101507674
};
\addplot [line width=1.25pt,  gdino, mark=square*, mark options={solid,rotate=180}, mark size=1, opacity=0.75, fill opacity=0.8]
table {%
	0.4 0.76347314686436
};
\addplot [line width=1.25pt,  yoloe, mark=diamond*, mark options={solid,rotate=180}, mark size=1, opacity=0.75, fill opacity=0.8]
table {%
	0.15 0.787628782425956
};
\addplot [line width=1.25pt,  yolow, mark=triangle*, mark options={solid,rotate=0}, mark size=1, opacity=0.75, fill opacity=0.8]
table {%
	0.15 0.791368695146034
};
\addplot [line width=1.25pt, gdino, mark=none, mark options={solid,rotate=180}, mark size=0.5, opacity=0.75, fill opacity=0.8]
table {%
	0.025 0.119563170030125
	0.05 0.24355859674165
	0.1 0.51464822931467
	0.15 0.640358367838754
	0.2 0.689117647702293
	0.3 0.742700005334383
	0.4 0.76347314686436
	0.5 0.74380043004277
	0.6 0.712222195668119
	0.7 0.62006890135648
	0.8 0.373432940570879
	0.85 0.174203902395871
	0.9 0.0204532356955086
	0.95 0
	0.975 0
};
\addplot [line width=1.25pt,  yoloe, mark=none, mark options={solid,rotate=180}, mark size=0.5, opacity=0.75, fill opacity=0.8]
table {%
	0.025 0.760726209253299
	0.05 0.777581915736443
	0.1 0.783768707880157
	0.15 0.787628782425956
	0.2 0.78527033863923
	0.3 0.783786507106581
	0.4 0.777141330602001
	0.5 0.766717338976675
	0.6 0.749497684682729
	0.7 0.730777143647308
	0.8 0.688349216504304
	0.85 0.631677667102087
	0.9 0.541528390702389
	0.95 0.306080496392881
	0.975 0
};
\addplot [line width=1.25pt, yolow, mark=none, mark options={solid,rotate=180}, mark size=0.5, opacity=0.75, fill opacity=0.8]
table {%
	0.025 0.774001093531448
	0.05 0.781620012514527
	0.1 0.790116521991642
	0.15 0.791368695146034
	0.2 0.79096724189025
	0.3 0.779851001365299
	0.4 0.772083455954809
	0.5 0.752362436945265
	0.6 0.733190757771667
	0.7 0.708854182625071
	0.8 0.672679457712533
	0.85 0.637428747333904
	0.9 0.563382797718881
	0.95 0.259107081589961
	0.975 0
};
\end{axis}

\end{tikzpicture}
	\end{minipage}%
	\begin{minipage}{0.156\textwidth}
		\begin{tikzpicture}

\definecolor{chocolate2196855}{RGB}{219,68,55}
\definecolor{darkgrey176}{RGB}{176,176,176}
\definecolor{darkorange2551094}{RGB}{255,109,4}
\definecolor{gainsboro229}{RGB}{229,229,229}
\definecolor{lightgrey204}{RGB}{204,204,204}
\definecolor{royalblue66133244}{RGB}{66,133,244}
\definecolor{seagreen1515788}{RGB}{15,157,88}

\definecolor{gainsboro247}{RGB}{253,253,253}
\definecolor{gainsboro229}{RGB}{239,239,239}
\definecolor{mediumpurple152142213}{RGB}{152,142,213}
\definecolor{gray119}{RGB}{119,119,119}
\definecolor{sandybrown25119394}{RGB}{251,193,94}

\definecolor{yolo11x}{RGB}{68,    3,   87} 
\definecolor{yolo11l}{RGB}{52,   71,  113.5} 
\definecolor{yolo11m}{RGB}{36,  139,  140} 
\definecolor{yolo11s}{RGB}{140.5, 184.5,  89.5} 
\definecolor{yolo11n}{RGB}{245,  230,   39} 	

\definecolor{yolow}{RGB}{255,109,4}
\definecolor{yoloe}{RGB}{0,48,73}
\definecolor{gdino}{RGB}{68,137,89} 

\tikzstyle{every node}=[font=\scriptsize]

\begin{axis}[
legend cell align={left},
legend style={
  fill opacity=0,
  draw opacity=0,
  text opacity=1,
  at={(0.0475,-0.0375)},
  anchor=south west,
  draw=lightgrey204,
  fill=gainsboro247
},
tick align=outside,
tick pos=left,
title={voc},
x grid style={darkgrey176},
xlabel={\textcolor{white}{A}},
xmajorgrids,
xmin=0, xmax=1,
y grid style={darkgrey176},
ymajorgrids,
ymin=0, ymax=0.9,
axis background/.style={fill=gainsboro247},
axis line style={gainsboro229},
y grid style={gainsboro229},
x grid style={gainsboro229},
axis background/.style={fill=gainsboro247},
ticklabel style={font=\tiny},
xtick style={draw=none},
ytick style={draw=none},
ytick align=inside,
width=4.175cm,
height=4.175cm,
title={\bf  YOLO11l},
title style={at={(0.5,1.05)},anchor=north},
y label style={at={(axis description cs:0.26, 0.5)}},
x label style={at={(axis description cs:0.5, 0.135)}},
yticklabel style = {xshift=0.5ex},
xticklabel style = {yshift=1.25ex},
xminorticks=true,
minor x tick num=1,
grid=both,
yminorticks=true,
minor y tick num=0,
ytick={0, 0.1, 0.2, 0.3, 0.4, 0.5, 0.6, 0.7, 0.8, 0.9},
yticklabels={},
xtick={0, 0.2, 0.4, 0.6, 0.8, 1},
xticklabels={, $0.2$, $0.4$, $0.6$, $0.8$, $1$},
]
\addplot [line width=1pt, gray119, mark=none, dashed, mark options={solid,rotate=180}, mark size=0.5, opacity=0.5, fill opacity=0.8]
table {%
	0 0.85461383461298
	0.1 0.85461383461298
	0.2 0.85461383461298
	0.3 0.85461383461298
	0.4 0.85461383461298
	0.5 0.85461383461298
	0.6 0.85461383461298
	0.7 0.85461383461298
	0.8 0.85461383461298
	0.9 0.85461383461298
	1 0.85461383461298
};
\addplot [line width=1.25pt,  gdino, mark=square*, mark options={solid,rotate=180}, mark size=1, opacity=0.75, fill opacity=0.8]
table {%
	0.4 0.761109907557463
};
\addplot [line width=1.25pt,  yoloe, mark=diamond*, mark options={solid,rotate=180}, mark size=1, opacity=0.75, fill opacity=0.8]
table {%
	0.15 0.7946017611555
};
\addplot [line width=1.25pt,  yolow, mark=triangle*, mark options={solid,rotate=0}, mark size=1, opacity=0.75, fill opacity=0.8]
table {%
	0.2 0.800259833329772
};
\addplot [line width=1.25pt, gdino, mark=none, mark options={solid,rotate=180}, mark size=0.5, opacity=0.75, fill opacity=0.8]
table {%
	0.025 0.123326003803161
	0.05 0.250418829946672
	0.1 0.52490263015555
	0.15 0.638980376809116
	0.2 0.697129308074809
	0.3 0.750005411047867
	0.4 0.761109907557463
	0.5 0.750931802991131
	0.6 0.715682589487706
	0.7 0.625232678673725
	0.8 0.372462729307705
	0.85 0.160592272575857
	0.9 0.0137532535732227
	0.95 0
	0.975 0
};
\addplot [line width=1.25pt,  yoloe, mark=none, mark options={solid,rotate=180}, mark size=0.5, opacity=0.75, fill opacity=0.8]
table {%
	0.025 0.772256098535383
	0.05 0.786278509097519
	0.1 0.789526764980062
	0.15 0.7946017611555
	0.2 0.792283146277727
	0.3 0.787191417140436
	0.4 0.784972228126768
	0.5 0.773931334922076
	0.6 0.758265489703985
	0.7 0.73295243887423
	0.8 0.686174452708875
	0.85 0.625674394683755
	0.9 0.542458537881988
	0.95 0.261271824295237
	0.975 0
};
\addplot [line width=1.25pt, yolow, mark=none, mark options={solid,rotate=180}, mark size=0.5, opacity=0.75, fill opacity=0.8]
table {%
	0.025 0.780471050537246
	0.05 0.791661316876097
	0.1 0.798919835944796
	0.15 0.800205395581052
	0.2 0.800259833329772
	0.3 0.785758981712321
	0.4 0.77642049530441
	0.5 0.756210313309194
	0.6 0.742116250359924
	0.7 0.71717972646211
	0.8 0.679645186719352
	0.85 0.638140078442644
	0.9 0.550378072375529
	0.95 0.221905130274032
	0.975 0
};
\end{axis}

\end{tikzpicture}
	\end{minipage}%
	\begin{minipage}{0.156\textwidth}
		\begin{tikzpicture}

\definecolor{chocolate2196855}{RGB}{219,68,55}
\definecolor{darkgrey176}{RGB}{176,176,176}
\definecolor{darkorange2551094}{RGB}{255,109,4}
\definecolor{gainsboro229}{RGB}{229,229,229}
\definecolor{lightgrey204}{RGB}{204,204,204}
\definecolor{royalblue66133244}{RGB}{66,133,244}
\definecolor{seagreen1515788}{RGB}{15,157,88}

\definecolor{gainsboro247}{RGB}{253,253,253}
\definecolor{gainsboro229}{RGB}{239,239,239}
\definecolor{mediumpurple152142213}{RGB}{152,142,213}
\definecolor{gray119}{RGB}{119,119,119}
\definecolor{sandybrown25119394}{RGB}{251,193,94}

\definecolor{yolo11x}{RGB}{68,    3,   87} 
\definecolor{yolo11l}{RGB}{52,   71,  113.5} 
\definecolor{yolo11m}{RGB}{36,  139,  140} 
\definecolor{yolo11s}{RGB}{140.5, 184.5,  89.5} 
\definecolor{yolo11n}{RGB}{245,  230,   39} 	

\definecolor{yolow}{RGB}{255,109,4}
\definecolor{yoloe}{RGB}{0,48,73}
\definecolor{gdino}{RGB}{68,137,89} 

\tikzstyle{every node}=[font=\scriptsize]

\begin{axis}[
legend cell align={left},
legend style={
  fill opacity=0,
  draw opacity=0,
  text opacity=1,
  at={(0.0475,-0.0375)},
  anchor=south west,
  draw=lightgrey204,
  fill=gainsboro247
},
tick align=outside,
tick pos=left,
title={voc},
x grid style={darkgrey176},
xlabel={\textcolor{white}{A}},
xmajorgrids,
xmin=0, xmax=1,
y grid style={darkgrey176},
ymajorgrids,
ymin=0, ymax=0.9,
axis background/.style={fill=gainsboro247},
axis line style={gainsboro229},
y grid style={gainsboro229},
x grid style={gainsboro229},
axis background/.style={fill=gainsboro247},
ticklabel style={font=\tiny},
xtick style={draw=none},
ytick style={draw=none},
ytick align=inside,
width=4.175cm,
height=4.175cm,
title={\bf  YOLO11x},
title style={at={(0.5,1.05)},anchor=north},
y label style={at={(axis description cs:0.26, 0.5)}},
x label style={at={(axis description cs:0.5, 0.135)}},
yticklabel style = {xshift=0.5ex},
xticklabel style = {yshift=1.25ex},
xminorticks=true,
minor x tick num=1,
grid=both,
yminorticks=true,
minor y tick num=0,
ytick={0, 0.1, 0.2, 0.3, 0.4, 0.5, 0.6, 0.7, 0.8, 0.9},
yticklabels={},
xtick={0, 0.2, 0.4, 0.6, 0.8, 1},
xticklabels={, $0.2$, $0.4$, $0.6$, $0.8$, $1$},
]
\addplot [line width=1pt, gray119, mark=none, dashed, mark options={solid,rotate=180}, mark size=0.5, opacity=0.5, fill opacity=0.8]
table {%
	0 0.859724629653635
	0.1 0.859724629653635
	0.2 0.859724629653635
	0.3 0.859724629653635
	0.4 0.859724629653635
	0.5 0.859724629653635
	0.6 0.859724629653635
	0.7 0.859724629653635
	0.8 0.859724629653635
	0.9 0.859724629653635
	1 0.859724629653635
};
\addplot [line width=1.25pt,  gdino, mark=square*, mark options={solid,rotate=180}, mark size=1, opacity=0.75, fill opacity=0.8]
table {%
	0.4 0.767553806494453
};
\addplot [line width=1.25pt,  yoloe, mark=diamond*, mark options={solid,rotate=180}, mark size=1, opacity=0.75, fill opacity=0.8]
table {%
	0.3 0.798432162270078
};
\addplot [line width=1.25pt,  yolow, mark=triangle*, mark options={solid,rotate=0}, mark size=1, opacity=0.75, fill opacity=0.8]
table {%
	0.1 0.802362707802486
};
\addplot [line width=1.25pt, gdino, mark=none, mark options={solid,rotate=180}, mark size=0.5, opacity=0.75, fill opacity=0.8]
table {%
	0.025 0.117230649559783
	0.05 0.249578211883772
	0.1 0.517258453413602
	0.15 0.635897954519298
	0.2 0.694933196503453
	0.3 0.751050824549202
	0.4 0.767553806494453
	0.5 0.751708447488604
	0.6 0.717662041908957
	0.7 0.631985052068592
	0.8 0.370401170823611
	0.85 0.175126033970994
	0.9 0.0278881927248792
	0.95 0
	0.975 0
};
\addplot [line width=1.25pt,  yoloe, mark=none, mark options={solid,rotate=180}, mark size=0.5, opacity=0.75, fill opacity=0.8]
table {%
	0.025 0.772846449488418
	0.05 0.792107134780051
	0.1 0.796251251329578
	0.15 0.798292333819798
	0.2 0.79372835859279
	0.3 0.798432162270078
	0.4 0.791206098461281
	0.5 0.773515986374594
	0.6 0.760430475128347
	0.7 0.743853455501012
	0.8 0.703384603662816
	0.85 0.648948038867011
	0.9 0.554765223647264
	0.95 0.271554928309799
	0.975 0
};
\addplot [line width=1.25pt, yolow, mark=none, mark options={solid,rotate=180}, mark size=0.5, opacity=0.75, fill opacity=0.8]
table {%
	0.025 0.781357487148295
	0.05 0.784607930271419
	0.1 0.802362707802486
	0.15 0.802080419248103
	0.2 0.802196455571067
	0.3 0.791143016054733
	0.4 0.779400379098868
	0.5 0.760817842486166
	0.6 0.752147571420765
	0.7 0.718319551553388
	0.8 0.682430276754403
	0.85 0.651984905237898
	0.9 0.575812746281881
	0.95 0.226280070428915
	0.975 0
};
\end{axis}

\end{tikzpicture}
	\end{minipage}%
	\begin{minipage}{0.18\textwidth}
		\begin{tikzpicture}

\definecolor{chocolate2196855}{RGB}{219,68,55}
\definecolor{darkgrey176}{RGB}{176,176,176}
\definecolor{darkorange2551094}{RGB}{255,109,4}
\definecolor{gainsboro229}{RGB}{229,229,229}
\definecolor{lightgrey204}{RGB}{204,204,204}
\definecolor{royalblue66133244}{RGB}{66,133,244}
\definecolor{seagreen1515788}{RGB}{15,157,88}

\definecolor{gainsboro247}{RGB}{253,253,253}
\definecolor{gainsboro229}{RGB}{239,239,239}
\definecolor{mediumpurple152142213}{RGB}{152,142,213}
\definecolor{gray119}{RGB}{119,119,119}
\definecolor{sandybrown25119394}{RGB}{251,193,94}

\definecolor{yolo11x}{RGB}{68,    3,   87} 
\definecolor{yolo11l}{RGB}{52,   71,  113.5} 
\definecolor{yolo11m}{RGB}{36,  139,  140} 
\definecolor{yolo11s}{RGB}{140.5, 184.5,  89.5} 
\definecolor{yolo11n}{RGB}{245,  230,   39} 	

\definecolor{yolow}{RGB}{255,109,4}
\definecolor{yoloe}{RGB}{0,48,73}
\definecolor{gdino}{RGB}{68,137,89} 

\tikzstyle{every node}=[font=\scriptsize]

\begin{axis}[
legend cell align={left},
legend style={
  fill opacity=0,
  draw opacity=0,
  text opacity=1,
  at={(0.0475,-0.0375)},
  anchor=south west,
  draw=lightgrey204,
  fill=gainsboro247
},
tick align=outside,
tick pos=left,
title={voc},
x grid style={darkgrey176},
xlabel={\textcolor{white}{A}},
xmajorgrids,
xmin=0, xmax=1,
y grid style={darkgrey176},
ymajorgrids,
ymin=0, ymax=0.9,
axis background/.style={fill=gainsboro247},
axis line style={gainsboro229},
y grid style={gainsboro229},
x grid style={gainsboro229},
axis background/.style={fill=gainsboro247},
ticklabel style={font=\tiny},
xtick style={draw=none},
ytick style={draw=none},
ytick align=inside,
width=4.175cm,
height=4.175cm,
title={\bf  RT-DETR},
title style={at={(0.5,1.05)},anchor=north},
y label style={at={(axis description cs:0.26, 0.5)}},
x label style={at={(axis description cs:0.5, 0.135)}},
yticklabel style = {xshift=16.05ex, anchor=near yticklabel opposite},
xticklabel style = {yshift=1.25ex},
xminorticks=true,
minor x tick num=1,
grid=both,
yminorticks=true,
minor y tick num=0,
ytick={0, 0.1, 0.2, 0.3, 0.4, 0.5, 0.6, 0.7, 0.8, 0.9},
xtick={0, 0.2, 0.4, 0.6, 0.8, 1},
xticklabels={, $0.2$, $0.4$, $0.6$, $0.8$, $1$},
]
\addplot [line width=1pt, gray119, mark=none, dashed, mark options={solid,rotate=180}, mark size=0.5, opacity=0.5, fill opacity=0.8]
table {%
	0 0.774877138102545
	0.1 0.774877138102545
	0.2 0.774877138102545
	0.3 0.774877138102545
	0.4 0.774877138102545
	0.5 0.774877138102545
	0.6 0.774877138102545
	0.7 0.774877138102545
	0.8 0.774877138102545
	0.9 0.774877138102545
	1 0.774877138102545
};
\addplot [line width=1.25pt,  gdino, mark=square*, mark options={solid,rotate=180}, mark size=1, opacity=0.75, fill opacity=0.8]
table {%
	0.4 0.682710418242096
};
\addplot [line width=1.25pt,  yoloe, mark=diamond*, mark options={solid,rotate=180}, mark size=1, opacity=0.75, fill opacity=0.8]
table {%
	0.4 0.697398991789479
};
\addplot [line width=1.25pt,  yolow, mark=triangle*, mark options={solid,rotate=0}, mark size=1, opacity=0.75, fill opacity=0.8]
table {%
	0.2 0.706064807316846
};
\addplot [line width=1.25pt, gdino, mark=none, mark options={solid,rotate=180}, mark size=0.5, opacity=0.75, fill opacity=0.8]
table {%
	0.025 0.0422325729387357
	0.05 0.0589810639995423
	0.1 0.161006867091201
	0.15 0.33659623003049
	0.2 0.468666485394937
	0.3 0.645612540482238
	0.4 0.682710418242096
	0.5 0.663135706362432
	0.6 0.63886078484404
	0.7 0.530891812843926
	0.8 0.273720822823701
	0.85 0.119902853263845
	0.9 0.000481845661343238
	0.95 6.42942252911794e-05
	0.975 6.31143020812092e-05
};
\addplot [line width=1.25pt,  yoloe, mark=none, mark options={solid,rotate=180}, mark size=0.5, opacity=0.75, fill opacity=0.8]
table {%
	0.025 0.502422366740008
	0.05 0.598029837272246
	0.1 0.657279612179828
	0.15 0.68575766150724
	0.2 0.69263147700268
	0.3 0.688280481807952
	0.4 0.697398991789479
	0.5 0.679164564561267
	0.6 0.666868102885664
	0.7 0.637293037416417
	0.8 0.60049238589252
	0.85 0.538832029696457
	0.9 0.453384490398711
	0.95 0.215174525566042
	0.975 6.31776746782073e-06
};
\addplot [line width=1.25pt, yolow, mark=none, mark options={solid,rotate=180}, mark size=0.5, opacity=0.75, fill opacity=0.8]
table {%
	0.025 0.567263272696039
	0.05 0.628351971620439
	0.1 0.68128656738865
	0.15 0.699148235793177
	0.2 0.706064807316846
	0.3 0.691185149621062
	0.4 0.684473803363139
	0.5 0.684323368515466
	0.6 0.653254391294514
	0.7 0.641067360874911
	0.8 0.60035954225618
	0.85 0.558368144351478
	0.9 0.467663418726669
	0.95 0.196325030517363
	0.975 0
};
\end{axis}

\end{tikzpicture}
	\end{minipage}%
	\vspace{-2.25em}
	\caption{{\bf Comparison of All AL-Trained Inference Models on VOC Validation}. Marks indicate best performance from each AL model.
	}
	\label{fig:voc}
\end{figure*}

\subsection{Auto-Labeling Evaluation via Downstream Model Training and Validation}
\label{sec:model}

To supplement the direct label evaluation in \cref{sec:label}, we now evaluate the efficacy of auto-labeling in terms of downstream inference model training and validation performance.
These experiments uniquely test an important function of AL in practice, i.e., training a downstream model for a specific application. 

Our overall study includes 445 model training and validation experiments to evaluate the numerous configurations of AL model, confidence threshold $\alpha$, and inference model (\cref{eq:fi}).
In each experiment, we first auto-label one of the train sets detailed in \cref{sec:data} using a specific AL model and $\alpha$.
Next, we use the AL train set to train an inference model (\cref{tab:inf}) for 100 epochs without any pre-trained weights.
Finally, we find the inference model's mean average precision (mAP50) on the corresponding validation set to quantify the effectiveness of AL-based training for that specific dataset and application.

We compare AL-based training across all AL models, $\alpha$ settings, and inference models on {\bf VOC} in \cref{fig:voc}.
For all inference models, the top mAP50 results corresponding to each AL model are relatively close, but training on YOLOW labels achieves the best mAP50 results.
Notably, YOLOW had the best AL $F_1$ score at $\alpha=0.5$ (\cref{fig:f1}) but the best mAP50 results at $\alpha \in [0.1,0.2]$.
Thus, {\bf the best downstream model performance is closer to peak auto-label recall than best auto-label $F_1$ score}.
Specifically, the best mAP50 results for each inference model are 0.718 for n-YW-0.15 (YOLO11n trained on YOLOW, $\alpha=0.15$ labels), 0.768 for s-YW-0.2, 0.791 for m-YW-0.15, 0.800 for l-YW-0.2, 0.802 for x-YW-0.1, and 0.706 for RT-DETR-YW-0.2.

In regards to {\bf inference model selection}, mAP50 increases with architecture size on VOC (\cref{fig:voc}), although this performance comes at the cost of a higher runtime (\cref{tab:inf}).
The only exception to this trend is the transformer-based RT-DETR model architecture, which is the second largest model with the second slowest runtime but the worst mAP50.
RT-DETR particularly performs worse than the YOLO11 architectures at lower $\alpha$ settings with high AL recall but low AL precision (\cref{fig:voc}, right).
Finally, each inference model architecture's best performance uses the same AL model at a relatively consistent $\alpha$ (YOLOW, $\alpha \in [0.1,0.2]$), which indicates that {\bf inference model selection can be decoupled from auto-label model configuration}.

\begin{figure}
	\centering
	\begin{minipage}{0.255\textwidth}
		\begin{tikzpicture}

\definecolor{chocolate2196855}{RGB}{219,68,55}
\definecolor{darkgrey176}{RGB}{176,176,176}
\definecolor{darkorange2551094}{RGB}{255,109,4}
\definecolor{gainsboro229}{RGB}{229,229,229}
\definecolor{lightgrey204}{RGB}{204,204,204}
\definecolor{royalblue66133244}{RGB}{66,133,244}
\definecolor{seagreen1515788}{RGB}{15,157,88}

\definecolor{gainsboro247}{RGB}{253,253,253}
\definecolor{gainsboro229}{RGB}{239,239,239}
\definecolor{mediumpurple152142213}{RGB}{152,142,213}
\definecolor{gray119}{RGB}{119,119,119}
\definecolor{sandybrown25119394}{RGB}{251,193,94}

\definecolor{yolo11x}{RGB}{68,    3,   87} 
\definecolor{yolo11l}{RGB}{52,   71,  113.5} 
\definecolor{yolo11m}{RGB}{36,  139,  140} 
\definecolor{yolo11s}{RGB}{140.5, 184.5,  89.5} 
\definecolor{yolo11n}{RGB}{245,  230,   39} 	

\definecolor{yolow}{RGB}{255,109,4}
\definecolor{yoloe}{RGB}{0,48,73}
\definecolor{gdino}{RGB}{68,137,89} 

\tikzstyle{every node}=[font=\scriptsize]

\begin{axis}[
legend cell align={left},
legend style={
  fill opacity=0,
  draw opacity=0,
  text opacity=1,
  at={(0.062,-0.03)},
  anchor=south west,
  draw=lightgrey204,
  fill=gainsboro247
},
tick align=outside,
tick pos=left,
title={voc},
x grid style={darkgrey176},
xlabel={\bf Confidence Threshold},
xmajorgrids,
xmin=0, xmax=1,
y grid style={darkgrey176},
ylabel={\bf mAP50},
ymajorgrids,
ymin=0, ymax=0.6,
axis background/.style={fill=gainsboro247},
axis line style={gainsboro229},
y grid style={gainsboro229},
x grid style={gainsboro229},
axis background/.style={fill=gainsboro247},
ticklabel style={font=\tiny},
xtick style={draw=none},
ytick style={draw=none},
ytick align=inside,
width=5.25cm,
height=5.25cm,
title={\bf  YOLO11n},
title style={at={(0.5,1.05)},anchor=north},
y label style={at={(axis description cs:0.19, 0.5)}},
x label style={at={(axis description cs:1.05, 0.1)}},
yticklabel style = {xshift=0.5ex},
xticklabel style = {yshift=1.25ex},
xminorticks=true,
minor x tick num=1,
grid=both,
yminorticks=true,
minor y tick num=1,
ytick={0, 0.1, 0.2, 0.3, 0.4, 0.5, 0.6, 0.7, 0.8, 0.9},
]
\addplot [line width=1pt, gray119, mark=none, dashed, mark options={solid,rotate=180}, mark size=0.5, opacity=0.5, fill opacity=0.8]
table {%
0 0.495608967839086
0.1 0.495608967839086
0.2 0.495608967839086
0.3 0.495608967839086
0.4 0.495608967839086
0.5 0.495608967839086
0.6 0.495608967839086
0.7 0.495608967839086
0.8 0.495608967839086
0.9 0.495608967839086
1 0.495608967839086
};
\addlegendentry{\tiny Human Labels}
\addplot [line width=1.25pt,  gdino, mark=square*, mark options={solid,rotate=180}, mark size=1, opacity=0.75, fill opacity=0.8]
table {%
	0.3 0.449842107690597
};
\addlegendentry{\tiny GDINO}
\addplot [line width=1.25pt,  yoloe, mark=diamond*, mark options={solid,rotate=180}, mark size=1, opacity=0.75, fill opacity=0.8]
table {%
	0.2 0.457627719346397
};
\addlegendentry{\tiny YOLOE}
\addplot [line width=1.25pt,  yolow, mark=triangle*, mark options={solid,rotate=0}, mark size=1, opacity=0.75, fill opacity=0.8]
table {%
	0.2 0.460395574305489
};
\addlegendentry{\tiny YOLOW}
\addplot [line width=1.25pt, gdino, mark=none, mark options={solid,rotate=180}, mark size=0.5, opacity=0.75, fill opacity=0.8]
table {%
	0.025 0.152475767010945
	0.05 0.240203725804155
	0.1 0.338829289668034
	0.15 0.396722603798861
	0.2 0.423977117580829
	0.3 0.449842107690597
	0.4 0.443805194150444
	0.5 0.432316543580565
	0.6 0.400123652088863
	0.7 0.342630120527173
	0.8 0.2437772896192
	0.85 0.15498858706631
	0.9 0.0447623569195548
	0.95 0
	0.975 0
};
\addplot [line width=1.25pt,  yoloe, mark=none, mark options={solid,rotate=180}, mark size=0.5, opacity=0.75, fill opacity=0.8]
table {%
	0.025 0.434819499314494
	0.05 0.447586090100535
	0.1 0.452665853548423
	0.15 0.456162778017664
	0.2 0.457627719346397
	0.3 0.453006667319414
	0.4 0.452549737520286
	0.5 0.440958979184388
	0.6 0.432337624293674
	0.7 0.41964265697988
	0.8 0.386732415654115
	0.85 0.36608781272846
	0.9 0.330425653747405
	0.95 0.226437327140672
	0.975 0.0633388676834472
};
\addplot [line width=1.25pt, yolow, mark=none, mark options={solid,rotate=180}, mark size=0.5, opacity=0.75, fill opacity=0.8]
table {%
	0.025 0.440183547657777
	0.05 0.450482906869701
	0.1 0.45912120611834
	0.15 0.456156448059402
	0.2 0.460395574305489
	0.3 0.45801589641782
	0.4 0.452622326115656
	0.5 0.445398163730251
	0.6 0.438086273267727
	0.7 0.427950796717199
	0.8 0.403615302885023
	0.85 0.374640081556403
	0.9 0.333788964254252
	0.95 0.205621119215808
	0.975 0.0442928607741507
};
\end{axis}

\end{tikzpicture}
	\end{minipage}%
	\begin{minipage}{0.24\textwidth}
		\begin{tikzpicture}

\definecolor{chocolate2196855}{RGB}{219,68,55}
\definecolor{darkgrey176}{RGB}{176,176,176}
\definecolor{darkorange2551094}{RGB}{255,109,4}
\definecolor{gainsboro229}{RGB}{229,229,229}
\definecolor{lightgrey204}{RGB}{204,204,204}
\definecolor{royalblue66133244}{RGB}{66,133,244}
\definecolor{seagreen1515788}{RGB}{15,157,88}

\definecolor{gainsboro247}{RGB}{253,253,253}
\definecolor{gainsboro229}{RGB}{239,239,239}
\definecolor{mediumpurple152142213}{RGB}{152,142,213}
\definecolor{gray119}{RGB}{119,119,119}
\definecolor{sandybrown25119394}{RGB}{251,193,94}

\definecolor{yolo11x}{RGB}{68,    3,   87} 
\definecolor{yolo11l}{RGB}{52,   71,  113.5} 
\definecolor{yolo11m}{RGB}{36,  139,  140} 
\definecolor{yolo11s}{RGB}{140.5, 184.5,  89.5} 
\definecolor{yolo11n}{RGB}{245,  230,   39} 	

\definecolor{yolow}{RGB}{255,109,4}
\definecolor{yoloe}{RGB}{0,48,73}
\definecolor{gdino}{RGB}{68,137,89} 

\tikzstyle{every node}=[font=\scriptsize]

\begin{axis}[
legend cell align={left},
legend style={
  fill opacity=0,
  draw opacity=0,
  text opacity=1,
  at={(0.0475,-0.0375)},
  anchor=south west,
  draw=lightgrey204,
  fill=gainsboro247
},
tick align=outside,
tick pos=left,
title={voc},
x grid style={darkgrey176},
xlabel={\textcolor{white}{A}},
xmajorgrids,
xmin=0, xmax=1,
y grid style={darkgrey176},
ymajorgrids,
ymin=0, ymax=0.6,
axis background/.style={fill=gainsboro247},
axis line style={gainsboro229},
y grid style={gainsboro229},
x grid style={gainsboro229},
axis background/.style={fill=gainsboro247},
ticklabel style={font=\tiny},
xtick style={draw=none},
ytick style={draw=none},
ytick align=inside,
width=5.25cm,
height=5.25cm,
title={\bf  YOLO11s},
title style={at={(0.5,1.05)},anchor=north},
y label style={at={(axis description cs:0.1, 0.5)}},
x label style={at={(axis description cs:0.5, -0.02)}},
yticklabel style = {xshift=0.5ex},
xticklabel style = {yshift=1.25ex},
xminorticks=true,
minor x tick num=1,
grid=both,
yminorticks=true,
minor y tick num=1,
ytick={0, 0.1, 0.2, 0.3, 0.4, 0.5, 0.6, 0.7, 0.8, 0.9},
yticklabels={},
]
\addplot [line width=1pt, gray119, mark=none, dashed, mark options={solid,rotate=180}, mark size=0.5, opacity=0.5, fill opacity=0.8]
table {%
0 0.588281630578545
0.1 0.588281630578545
0.2 0.588281630578545
0.3 0.588281630578545
0.4 0.588281630578545
0.5 0.588281630578545
0.6 0.588281630578545
0.7 0.588281630578545
0.8 0.588281630578545
0.9 0.588281630578545
1 0.588281630578545
};
\addplot [line width=1.25pt,  gdino, mark=square*, mark options={solid,rotate=180}, mark size=1, opacity=0.75, fill opacity=0.8]
table {%
0.3 0.528988083664346
};
\addplot [line width=1.25pt,  yoloe, mark=diamond*, mark options={solid,rotate=180}, mark size=1, opacity=0.75, fill opacity=0.8]
table {%
0.15 0.530402170940181
};
\addplot [line width=1.25pt,  yolow, mark=triangle*, mark options={solid,rotate=0}, mark size=1, opacity=0.75, fill opacity=0.8]
table {%
0.2 0.538484825600994
};
\addplot [line width=1.25pt, gdino, mark=none, mark options={solid,rotate=180}, mark size=0.5, opacity=0.75, fill opacity=0.8]
table {%
	0.025 0.165048464071789
	0.05 0.269596535143515
	0.1 0.398939770487758
	0.15 0.466388108336023
	0.2 0.499618771843004
	0.3 0.528988083664346
	0.4 0.52382590868228
	0.5 0.501230820029049
	0.6 0.464189737865796
	0.7 0.399657706115702
	0.8 0.285036289891645
	0.85 0.180562174477971
	0.9 0.0497338943384669
	0.95 0
	0.975 0
};
\addplot [line width=1.25pt,  yoloe, mark=none, mark options={solid,rotate=180}, mark size=0.5, opacity=0.75, fill opacity=0.8]
table {%
	0.025 0.505076759426284
	0.05 0.517592754921032
	0.1 0.526059103493975
	0.15 0.530402170940181
	0.2 0.525538289211314
	0.3 0.525941801982406
	0.4 0.514967475001716
	0.5 0.509763433223388
	0.6 0.496329878244676
	0.7 0.477549314227442
	0.8 0.443014757905162
	0.85 0.424723567625452
	0.9 0.382645266476133
	0.95 0.252328797410901
	0.975 0.0850470553089257
};
\addplot [line width=1.25pt, yolow, mark=none, mark options={solid,rotate=180}, mark size=0.5, opacity=0.75, fill opacity=0.8]
table {%
	0.025 0.515271693057001
	0.05 0.528848564204141
	0.1 0.534239949398778
	0.15 0.537787731284775
	0.2 0.538484825600994
	0.3 0.530108275661879
	0.4 0.525124552223134
	0.5 0.519169707062895
	0.6 0.507120429173351
	0.7 0.492571188552125
	0.8 0.469237753567349
	0.85 0.440557980639725
	0.9 0.389047002010782
	0.95 0.237856255037725
	0.975 0.0563783113212662
};
\end{axis}

\end{tikzpicture}
	\end{minipage}%
	\vspace{-2.25em}
	\caption{{\bf Comparison of AL Training on COCO Validation}.
	}
	\label{fig:coco}
\end{figure}
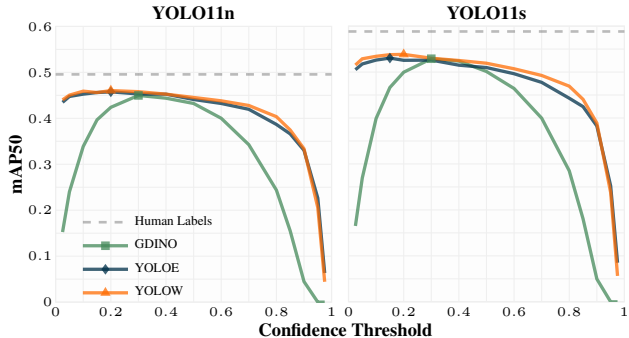

We compare AL-based training on {\bf COCO} in \cref{fig:coco}.
Similar to VOC, the top COCO mAP50 corresponding to each AL model is relatively close, but the best mAP50 results from training on YOLOW labels, which had tied GDINO for the highest AL $F_1$ score (\cref{fig:f1}).
As with VOC, the top mAP50 results for each AL model use relatively low $\alpha$ settings between peak AL recall and AL $F_1$ score, and mAP50 increases with inference model size. 
Specifically,  training on YOLOW-0.2 labels achieves the best mAP50 results of 0.460 for YOLO11n and 0.538 for YOLO11s.

\setlength{\tabcolsep}{4pt}
\begin{table}
	\centering
	\caption{ {\bf AL, Human Labels, \& Inference Model Performance}.
		All YOLOW labels use $\alpha=0.2$ confidence threshold.
		Bold font is best dataset AL \& human label validation performance.
		Notably, inference model determines rank order more than label source.
	}
	\vspace{-0.75em}
	\scriptsize
	\begin{tabular}{ c | l | r | c| c | c | c |}
		\hline
		\rowcolor{tableheader} \multicolumn{1}{| c |}{} & \multicolumn{1}{ c |}{\bf Inference} & \multicolumn{1}{ c |}{\bf \# of} & \multicolumn{1}{ c |}{\bf Label} & \multicolumn{3}{ c |}{\bf Validation Performance}  \\ 
		\rowcolor{tableheader} \multicolumn{1}{| c |}{\bf Dataset} & \multicolumn{1}{ c |}{\bf Model} & \multicolumn{1}{ c |}{\bf Params} & \multicolumn{1}{ c |}{\bf Source} & \multicolumn{1}{ c }{\bf mAP50} & \multicolumn{1}{ c }{\bf mAP75} & \multicolumn{1}{ c |}{\bf mAP50-95} \\ \hline
		& YOLO11s	&	9.5	\textrm{M}&	Human &	\bf 0.817	&	\bf0.672	&	\bf0.613	\\  \cline{2-7}
		\bf VOC & YOLO11s	&	9.5	\textrm{M}&	YOLOW &	\bf0.768 & \bf0.636 & \bf0.577	\\  \cline{2-7}
		&YOLO11n	&	2.6	\textrm{M}&	Human	&	0.756	&	0.605	&	0.549	\\ \cline{2-7}
		& YOLO11n	&	2.6	\textrm{M}&	YOLOW &	0.715 & 0.573 & 0.519	\\  \cline{2-7}
		\noalign{\vskip2.0pt} \cline{2-7}
		& YOLO11s	&	9.5	\textrm{M}&	Human &	\bf 0.588	& \bf 0.463 & \bf 0.428	\\  \cline{2-7}
		\bf COCO & YOLO11s	&	9.5	\textrm{M}&	YOLOW &	\bf 0.538 &\bf 0.419 & \bf 0.386	\\  \cline{2-7}
		&YOLO11n	&	2.6	\textrm{M}&	Human	&	0.496	&	0.378 &	0.349	\\ \cline{2-7}
		& YOLO11n	&	2.6	\textrm{M}&	YOLOW &	0.460 & 0.345 & 0.320	\\  \cline{2-7}
	\end{tabular}
	\label{tab:alhl}
\end{table}

Although the AL-trained inference model performance on COCO is lower than VOC, {\bf AL performance is competitive with human labels on both datasets}.
We compare training on a single AL model and $\alpha$ setting to human label-based training in \cref{tab:alhl}, which shows that the inference model architecture itself determines the ranked order of performance more than the decision to use AL or human labels.
This result presents an interesting tradeoff in terms of cost and performance.
That is, if we save costs by using AL instead of an annotation service (\cref{tab:cost}) and redirect our budget to accommodate a larger inference model, the net result is higher overall performance.
Specifically, switching from YOLO11n with human labels to YOLO11s with AL increases VOC mAP 2--5\% and COCO mAP 9--11\%.

\begin{figure}
	\centering
	\begin{minipage}{0.245\textwidth}
		\begin{tikzpicture}

\definecolor{chocolate2196855}{RGB}{219,68,55}
\definecolor{darkgrey176}{RGB}{176,176,176}
\definecolor{darkorange2551094}{RGB}{255,109,4}
\definecolor{gainsboro229}{RGB}{229,229,229}
\definecolor{lightgrey204}{RGB}{204,204,204}
\definecolor{royalblue66133244}{RGB}{66,133,244}
\definecolor{seagreen1515788}{RGB}{15,157,88}

\definecolor{gainsboro247}{RGB}{253,253,253}
\definecolor{gainsboro229}{RGB}{239,239,239}
\definecolor{mediumpurple152142213}{RGB}{152,142,213}
\definecolor{gray119}{RGB}{119,119,119}
\definecolor{sandybrown25119394}{RGB}{251,193,94}

\definecolor{yolo11x}{RGB}{68,    3,   87} 
\definecolor{yolo11l}{RGB}{52,   71,  113.5} 
\definecolor{yolo11m}{RGB}{36,  139,  140} 
\definecolor{yolo11s}{RGB}{140.5, 184.5,  89.5} 
\definecolor{yolo11n}{RGB}{245,  230,   39} 	

\definecolor{yolow}{RGB}{255,109,4}
\definecolor{yoloe}{RGB}{0,48,73}
\definecolor{gdino}{RGB}{68,137,89} 

\tikzstyle{every node}=[font=\scriptsize]

\begin{axis}[
legend cell align={left},
legend style={
  fill opacity=0,
  draw opacity=0,
  text opacity=1,
  at={(0.062,-0.015)},
  anchor=south west,
  draw=lightgrey204,
  fill=gainsboro247
},
tick align=outside,
tick pos=left,
title={voc},
x grid style={darkgrey176},
xlabel={\bf Confidence Threshold},
xmajorgrids,
xmin=0, xmax=1,
y grid style={darkgrey176},
ylabel={\bf mAP50},
ymajorgrids,
ymin=0, ymax=0.1,
axis background/.style={fill=gainsboro247},
axis line style={gainsboro229},
y grid style={gainsboro229},
x grid style={gainsboro229},
axis background/.style={fill=gainsboro247},
ticklabel style={font=\tiny},
xtick style={draw=none},
ytick style={draw=none},
ytick align=inside,
width=5cm,
height=5cm,
title={\bf  LVIS},
title style={at={(0.5,1.05)},anchor=north},
y label style={at={(axis description cs:0.175, 0.5)}},
x label style={at={(axis description cs:1.05, 0.1)}},
yticklabel style = {xshift=0.5ex},
xticklabel style = {yshift=1.25ex},
xminorticks=true,
minor x tick num=1,
grid=both,
yminorticks=true,
minor y tick num=1,
ytick={0, 0.02, 0.04, 0.06, 0.08, 0.10},
yticklabels={$0$, $0.02$, $0.04$, $0.06$, $0.08$, $0.10$},
]
\addplot [line width=1pt, gray119, mark=none, dashed, mark options={solid,rotate=180}, mark size=0.5, opacity=0.5, fill opacity=0.8]
table {%
0 0.0872404754586397
0.1 0.0872404754586397
0.2 0.0872404754586397
0.3 0.0872404754586397
0.4 0.0872404754586397
0.5 0.0872404754586397
0.6 0.0872404754586397
0.7 0.0872404754586397
0.8 0.0872404754586397
0.9 0.0872404754586397
1 0.0872404754586397
};
\addlegendentry{\tiny Human Labels}
\addplot [line width=1.25pt,  gdino, mark=square*, mark options={solid,rotate=180}, mark size=1, opacity=0.75, fill opacity=0.8]
table {%
	0.3 -0.02
};
\addlegendentry{\tiny GDINO}
\addplot [line width=1.25pt,  yoloe, mark=diamond*, mark options={solid,rotate=180}, mark size=1, opacity=0.75, fill opacity=0.8]
table {%
0.1 0.0659136829048976
};
\addlegendentry{\tiny YOLOE}
\addplot [line width=1.25pt,  yolow, mark=triangle*, mark options={solid,rotate=0}, mark size=1, opacity=0.75, fill opacity=0.8]
table {%
0.2 0.0587843599952048
};
\addlegendentry{\tiny YOLOW}
\addplot [line width=1.25pt,  yoloe, mark=none, mark options={solid,rotate=180}, mark size=0.5, opacity=0.75, fill opacity=0.8]
table {%
0.05 0.062178343355225
0.1 0.0659136829048976
0.15 0.0641995523435659
0.2 0.0641565187105717
0.3 0.0623289804597649
0.4 0.0623944918638393
0.5 0.057169014821918
0.6 0.0544632934914501
0.7 0.0490493424695998
0.8 0.04049838605254
0.85 0.0348864090123039
0.9 0.0245874300170142
0.95 0.0119264551742067
0.975 0
};
\addplot [line width=1.25pt, yolow, mark=none, mark options={solid,rotate=180}, mark size=0.5, opacity=0.75, fill opacity=0.8]
table {%
0.025 0.0505781434538038
0.05 0.0542176956665581
0.1 0.0570868381036168
0.15 0.0565682077209933
0.2 0.0587843599952048
0.3 0.0566447455328617
0.4 0.0530905664555614
0.5 0.0500363733085684
0.6 0.0452144804681169
0.7 0.0395219001538753
0.8 0.0308639977269051
0.85 0.0241340196538109
0.9 0.0157698222722216
0.95 0.0063426212545403
0.975 0.00227967868187047
};
\end{axis}

\end{tikzpicture}
	\end{minipage}%
	\begin{minipage}{0.24\textwidth}
		\begin{tikzpicture}
	
	\definecolor{chocolate2196855}{RGB}{219,68,55}
	\definecolor{darkgrey176}{RGB}{176,176,176}
	\definecolor{darkorange2551094}{RGB}{255,109,4}
	\definecolor{gainsboro229}{RGB}{229,229,229}
	\definecolor{lightgrey204}{RGB}{204,204,204}
	\definecolor{royalblue66133244}{RGB}{66,133,244}
	\definecolor{seagreen1515788}{RGB}{15,157,88}
	
	\definecolor{gainsboro247}{RGB}{253,253,253}
	\definecolor{gainsboro229}{RGB}{239,239,239}
	\definecolor{mediumpurple152142213}{RGB}{152,142,213}
	\definecolor{gray119}{RGB}{119,119,119}
	\definecolor{sandybrown25119394}{RGB}{251,193,94}
	
	\definecolor{yolo11x}{RGB}{68,    3,   87} 
	\definecolor{yolo11l}{RGB}{52,   71,  113.5} 
	\definecolor{yolo11m}{RGB}{36,  139,  140} 
	\definecolor{yolo11s}{RGB}{140.5, 184.5,  89.5} 
	\definecolor{yolo11n}{RGB}{245,  230,   39} 	
	
	\definecolor{yolow}{RGB}{255,109,4}
	\definecolor{yoloe}{RGB}{0,48,73}
	\definecolor{gdino}{RGB}{68,137,89} 
	
	\tikzstyle{every node}=[font=\scriptsize]
	
	\begin{axis}[
		legend cell align={left},
		legend style={
			fill opacity=0,
			draw opacity=0,
			text opacity=1,
			at={(0.0475,-0.0375)},
			anchor=south west,
			draw=lightgrey204,
			fill=gainsboro247
		},
		tick align=outside,
		tick pos=left,
		title={voc},
		x grid style={darkgrey176},
		xlabel={\textcolor{white}{A}},
		xmajorgrids,
		xmin=0, xmax=1,
		y grid style={darkgrey176},
		ymajorgrids,
		ymin=0, ymax=0.5,
		axis background/.style={fill=gainsboro247},
		axis line style={gainsboro229},
		y grid style={gainsboro229},
		x grid style={gainsboro229},
		axis background/.style={fill=gainsboro247},
		ticklabel style={font=\tiny},
		xtick style={draw=none},
		ytick style={draw=none},
		ytick align=inside,
		width=5cm,
		height=5cm,
		title={\bf  BDD},
		title style={at={(0.5,1.05)},anchor=north},
		y label style={at={(axis description cs:0.1, 0.5)}},
		x label style={at={(axis description cs:0.5, -0.02)}},
		yticklabel style = {xshift=0.5ex},
		xticklabel style = {yshift=1.25ex},
		xminorticks=true,
		minor x tick num=1,
		grid=both,
		yminorticks=true,
		minor y tick num=1,
		ytick={0, 0.1, 0.2, 0.3, 0.4, 0.5},
		yticklabels={$0$, $0.1$, $0.2$, $0.3$, $0.4$, $0.5$},
		]
		\addplot [line width=1pt, gray119, mark=none, dashed, mark options={solid,rotate=180}, mark size=0.5, opacity=0.5, fill opacity=0.8]
		table {%
			0 0.434468817217912
			0.1 0.434468817217912
			0.2 0.434468817217912
			0.3 0.434468817217912
			0.4 0.434468817217912
			0.5 0.434468817217912
			0.6 0.434468817217912
			0.7 0.434468817217912
			0.8 0.434468817217912
			0.9 0.434468817217912
			1 0.434468817217912
		};
		\addplot [line width=1.25pt,  gdino, mark=square*, mark options={solid,rotate=180}, mark size=1, opacity=0.75, fill opacity=0.8]
		table {%
			0.3 0.298264325935749
		};
		\addplot [line width=1.25pt,  yoloe, mark=diamond*, mark options={solid,rotate=180}, mark size=1, opacity=0.75, fill opacity=0.8]
		table {%
			0.1 0.256890262269388
		};
		\addplot [line width=1.25pt,  yolow, mark=triangle*, mark options={solid,rotate=0}, mark size=1, opacity=0.75, fill opacity=0.8]
		table {%
			0.05 0.278400622241106
		};
		\addplot [line width=1.25pt, gdino, mark=none, mark options={solid,rotate=180}, mark size=0.5, opacity=0.75, fill opacity=0.8]
		table {%
			0.025 0.0367141591707362
			0.05 0.0867938068989057
			0.1 0.186767412324337
			0.15 0.24762151359077
			0.2 0.280272309120282
			0.3 0.298264325935749
			0.4 0.288709297127919
			0.5 0.278201186157192
			0.6 0.262664816513317
			0.7 0.251623912996751
			0.8 0
			0.85 0
			0.9 0
			0.95 0
			0.975 0
		};
		\addplot [line width=1.25pt,  yoloe, mark=none, mark options={solid,rotate=180}, mark size=0.5, opacity=0.75, fill opacity=0.8]
		table {%
			0.025 0.243964440163422
			0.05 0.252345838276393
			0.1 0.256890262269388
			0.15 0.252545112270733
			0.2 0.254934080813296
			0.3 0.25160351931583
			0.4 0.24863352480128
			0.5 0.243700884677853
			0.6 0.239519072076063
			0.7 0.241792819453667
			0.8 0.240454029891328
			0.85 0.239186437429022
			0.9 0.237385865940638
			0.95 0.081877068409763
			0.975 0
		};
		\addplot [line width=1.25pt, yolow, mark=none, mark options={solid,rotate=180}, mark size=0.5, opacity=0.75, fill opacity=0.8]
		table {%
			0.025 0.277204080250751
			0.05 0.278400622241106
			0.1 0.275260701723877
			0.15 0.271565647502581
			0.2 0.270714750301952
			0.3 0.265483784309325
			0.4 0.259283312344606
			0.5 0.259034202113215
			0.6 0.254174089513366
			0.7 0.254277484430792
			0.8 0.247159609472309
			0.85 0.252652869352723
			0.9 0.211031418536172
			0.95 0
			0.975 0
		};
	\end{axis}
	
\end{tikzpicture}
	\end{minipage}%
	\vspace{-2.25em}
	\caption{{\bf Comparison of AL Training on LVIS \& BDD Validation}.
		Inference training and validation uses YOLO11n model.
	}
	\label{fig:lvis}
\end{figure}
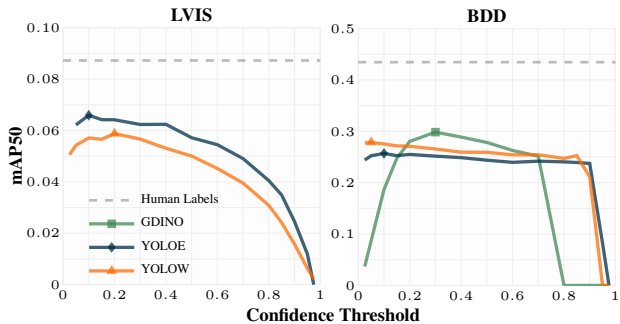

We compare AL-based training on LVIS and BDD in \cref{fig:lvis}. 
Each dataset provides unique challenges, and we find that object detection on {\bf LVIS} with 1,203 unique classes is the most challenging.
In fact, none of the inference models achieve an $\text{mAP50}>0.09$, even when training on human labels.
For AL specifically, inference models trained on YOLOE labels have a higher mAP50 than those trained on YOLOW labels except when $\alpha>0.9$, which causes YOLOE to generate dramatically fewer labels (\cref{fig:f1}). 

For AL-based training on {\bf BDD}, the best mAP50 of 0.298 results from training on GDINO labels.
Interestingly, GDINO had a lower AL $F_1$ score than YOLOW (\cref{fig:f1}), which demonstrates that {\bf labels with the best $F_1$ score relative to human labels are not necessarily the best labels for downstream model training.}
Notably, the performance gap between inference models trained on auto-labels vs. human labels is greater on BDD than any other dataset, which indicates a deficiency in label quality rather than object detection difficulty.
This likely occurs because BDD consists entirely of autonomous driving viewpoints that are atypical of the original AL model training data (\cref{tab:fa}).

\subsection{Detailed Comparison of Auto-Label and \\Downstream Inference Model Metrics}
\label{sec:bp}

\setlength{\tabcolsep}{3.625pt}
\begin{table*}
	\centering
	\caption{ {\bf Label Evaluation Metrics vs. AL-Trained Inference Model Validation}.
		Results are organized by confidence threshold $\alpha = \{\text{\scriptsize \bf 0.2, 0.5, 0.8}\}$.
		Green and red highlighting indicate the ``best" and ``worst" result for a metric, while bold font indicates the $\alpha$ setting leading to the best downstream YOLO11n model performance.
		Notably, $\alpha= \text{\scriptsize \bf 0.8}$ has highest precision but worst model performance in every row.
	}
	\vspace{-0.75em}
	\scriptsize
	\begin{tabular}{  c | c | c  | r | r | r | c  | c | c | c | c | c | c | c | c | c | c | c | c | c | c | c | }
		\cline{1-2} \cline{4-6} \cline{8-10} \cline{12-14} \cline{16-18} \cline{20-22}
		\noalign{\vskip0.4pt} 
		\rowcolor{tableheader} \multicolumn{1}{| c }{} & \multicolumn{1}{ c |}{\bf AL}  & \cellcolor{white} & \multicolumn{3}{c |}{\bf Number of Train Labels} & \cellcolor{white} & \multicolumn{3}{c |}{\bf Train Label Precision} & \cellcolor{white} & \multicolumn{3}{c |}{\bf Train Labels Recall} & \cellcolor{white} & \multicolumn{3}{c |}{\bf Train Label ${\text{F}}_\text{1}$ Score} & \cellcolor{white} & \multicolumn{3}{c |}{\bf Validation mAP50} \\
		\rowcolor{tableheader} \multicolumn{1}{| c }{\bf Dataset} & \multicolumn{1}{ c |}{\bf Model} & \cellcolor{white} & \multicolumn{1}{ c }{\bf 0.2} & \multicolumn{1}{ c }{\bf 0.5} & \multicolumn{1}{ c |}{\bf 0.8} & \cellcolor{white} & \multicolumn{1}{ c }{\bf 0.2} & \multicolumn{1}{ c }{\bf 0.5} & \multicolumn{1}{ c |}{\bf 0.8}& \cellcolor{white} & \multicolumn{1}{ c }{\bf 0.2} & \multicolumn{1}{ c }{\bf 0.5} & \multicolumn{1}{ c |}{\bf 0.8}& \cellcolor{white} & \multicolumn{1}{ c }{\bf 0.2} & \multicolumn{1}{ c }{\bf 0.5} & \multicolumn{1}{ c |}{\bf 0.8}& \cellcolor{white} & \multicolumn{1}{ c }{\bf 0.2} & \multicolumn{1}{ c }{\bf 0.5} & \multicolumn{1}{ c |}{\bf 0.8} \\ \cline{1-2} \cline{4-6} \cline{8-10} \cline{12-14} \cline{16-18} \cline{20-22}
		\noalign{\vskip2.0pt} \cline{2-2} \cline{4-6} \cline{8-10} \cline{12-14} \cline{16-18} \cline{20-22} \noalign{\vskip0.4pt}
		& YOLOW	 &&	\bf 60,026	&	\cellcolor{tablegreen} 40,028	&	23,872	&&	\bf 0.596	&	0.786	&	0.937	&&	\bf 0.893	&	0.785	&	0.558	&&	\bf 0.715	&	\cellcolor{tablegreen} 0.785	&	0.700	&&	\cellcolor{tablegreen} \bf 0.715	&	0.681	&	0.612	\\ \cline{2-2} \cline{4-6} \cline{8-10} \cline{12-14} \cline{16-18} \cline{20-22}
		\bf VOC & YOLOE	&&	\bf 62,349	&	38,964	&	21,411	&&	\bf 0.563	&	0.771	&	0.927	&&	\bf 0.876	&	0.750	&	0.495	&&	\bf 0.685	&	0.761	&	0.646	&&	\bf 0.709	&	0.682	&	0.613	\\ \cline{2-2} \cline{4-6} \cline{8-10} \cline{12-14} \cline{16-18} \cline{20-22} \noalign{\vskip0.4pt}
		& GDINO	&&	\cellcolor{tablered} 147,976	&	\bf 37,800	&	8,507	&&	\cellcolor{tablered} 0.243	&	\bf 0.781	&	\cellcolor{tablegreen} 0.970	&&	\cellcolor{tablegreen} 0.899	&	\bf 0.737	&	\cellcolor{tablered} 0.206	&&	0.383	&\bf 	0.759	&	\cellcolor{tablered} 0.340	&&	0.637	&	\bf 0.674	&	\cellcolor{tablered} 0.314	\\  \cline{2-2}   \cline{4-6} \cline{8-10} \cline{12-14} \cline{16-18} \cline{20-22} 
		\noalign{\vskip2.0pt} \cline{4-6} \cline{8-10} \cline{12-14} \cline{16-18} \cline{20-22} \cline{2-2}  \noalign{\vskip0.4pt}
		&	YOLOW	& &	\bf 1,013,600	&	557,894	&	267,692	& &	\bf 0.563	&	0.787	&	0.940	& &	\bf 0.671	&	0.517	&	0.296	& &	\bf 0.612	&	\cellcolor{tablegreen} 0.624	&	0.450	& &	\bf \cellcolor{tablegreen} \bf 0.460	&	0.445	&	0.404	\\ \cline{2-2} \cline{4-6} \cline{8-10} \cline{12-14} \cline{16-18} \cline{20-22} \noalign{\vskip0.4pt}
		\bf COCO	&	YOLOE	& &	\bf \cellcolor{tablegreen} 946,312	&	487,148	&	218,291	& &	\bf 0.559	&	0.792	&	0.941	& &	\bf 0.623	&	0.454	&	0.242	& &	\bf 0.589	&	0.577	&	0.384	& &	\bf 0.458	&	0.441	&	0.387	\\ \cline{2-2} \cline{4-6} \cline{8-10} \cline{12-14} \cline{16-18} \cline{20-22} \noalign{\vskip0.4pt}
		&	GDINO	& &	\cellcolor{tablered} 2,232,127	&	\bf 479,099	&	108,240	& &	\cellcolor{tablered} 0.281	&	\bf 0.826	&	\cellcolor{tablegreen} 0.983	& &	\cellcolor{tablegreen} 0.739	&	\bf 0.466	&	\cellcolor{tablered} 0.125	& &	0.408	&	\bf 0.595	&	\cellcolor{tablered} 0.222	& &	0.424	&	\bf 0.432	&	\cellcolor{tablered} 0.244	\\\cline{4-6} \cline{8-10} \cline{12-14} \cline{16-18} \cline{20-22} \cline{2-2} 
		\noalign{\vskip2.0pt} \cline{4-6} \cline{8-10} \cline{12-14} \cline{16-18} \cline{20-22} \cline{2-2}  \noalign{\vskip0.4pt}
		\bf LVIS &	YOLOW	& &	\bf 1,701,295	&	610,040	&	\cellcolor{tablered}135,523	& &	\cellcolor{tablered}\bf 0.165	&	0.262	&	0.331	& &	\bf \cellcolor{tablegreen} 0.221	&	0.126	&	\cellcolor{tablered}0.035	& &	\bf 0.189	&	0.170	&	\cellcolor{tablered}0.064	& &	\bf 0.059	&	0.050	&	\cellcolor{tablered}0.031	\\ \cline{2-2} \cline{4-6} \cline{8-10} \cline{12-14} \cline{16-18} \cline{20-22}\noalign{\vskip0.4pt}
		&	YOLOE	& &	\bf \cellcolor{tablegreen} 1,311,999	&	486,847	&	145,360	& &	\bf 0.208	&	0.346	&	\cellcolor{tablegreen} 0.496	& &	\bf 0.215	&	0.133	&	0.057	& &	\bf \cellcolor{tablegreen} 0.211	&	0.192	&	0.102	& &	\bf \cellcolor{tablegreen} 0.064	&	0.057	&	0.040	\\ \cline{2-2} \cline{4-6} \cline{8-10} \cline{12-14} \cline{16-18} \cline{20-22}
		\noalign{\vskip2.0pt} \cline{4-6} \cline{8-10} \cline{12-14} \cline{16-18} \cline{20-22}  \cline{2-2} 
		&	YOLOW	& &	\bf 401,485	&	148,782	&	21,053	& &	\bf 0.804	&	0.910	&	0.945	& &	\bf 0.251	&	0.105	&	0.015	& &	\bf 0.382	&	0.189	&	0.030	& &	\bf 0.271	&	0.259	&	0.247	\\ \cline{2-2} \cline{4-6} \cline{8-10} \cline{12-14} \cline{16-18} \cline{20-22}
		\bf BDD	&	YOLOE	& &	\bf 449,403	&	170,489	&	27,507	& &	\bf 0.766	&	0.889	&	0.936	& &	\bf 0.267	&	0.118	&	0.020	& &	\bf 0.396	&	0.208	&	0.039	& &	\bf 0.255	&	 0.244	&	0.240	\\ \cline{2-2} \cline{4-6} \cline{8-10} \cline{12-14} \cline{16-18} \cline{20-22} \noalign{\vskip0.4pt}
		&	GDINO	& &\bf 	\cellcolor{tablegreen} 1,381,568	&	190,577	&	\cellcolor{tablered} 655	& &\bf 	\cellcolor{tablered} 0.478	&	0.867	&	\cellcolor{tablegreen} 0.963	& &\bf 	\cellcolor{tablegreen} 0.513	&	0.128	&	\cellcolor{tablered} 0.000	& &\bf 	\cellcolor{tablegreen} 0.495	&	0.224	&	\cellcolor{tablered} 0.001	& &\bf  \cellcolor{tablegreen} \bf  0.280	&	0.278	&	\cellcolor{tablered} 0.000	\\  \cline{2-2}  \cline{4-6} \cline{8-10} \cline{12-14} \cline{16-18} \cline{20-22}
	\end{tabular}
	\label{tab:f1map}
\end{table*}

To establish {\bf best auto-labeling practices}, we perform a combined analysis of label evaluation metrics (\cref{sec:label}) and model performance metrics (\cref{sec:model}) using a representative subset of auto-labeling configurations in \cref{tab:f1map}. 
Note that each of the 11 rows in \cref{tab:f1map} corresponds to a unique AL model-dataset pair with corresponding results for three confidence thresholds, $\alpha = \{0.2, 0.5, 0.8\}$. 
From this analysis, we find the single auto-labeling configuration that best trains downstream inference models across all datasets.

We start by sharing a few {\bf general findings} from the results in \cref{tab:f1map}.
First, $\alpha=0.2$ results in the best AL recall for all rows, best mAP50 for 9 rows, and best AL $F_1$ score for 6 rows.
Second, $\alpha=0.5$ results in the best AL $F_1$ score for 5 rows and best mAP50 for 2 rows.
Finally, $\alpha=0.8$ results in the best AL precision \textit{and} the worst mAP50 for all rows.
Thus, for experiments in \cref{tab:f1map}, we can conclude the following.
First, $\alpha=0.2$ is the best confidence threshold for AL recall, AL $F_1$ score, and downstream model performance.
Second, $\alpha=0.8$ is the worst confidence threshold for all metrics except for AL precision.
Finally, {\bf high auto-label recall is the best single predictor of downstream model performance} followed by high auto-label $F_1$ score.

We now compare the {\bf general performance of each AL model} in \cref{tab:f1map}.
All AL models trains an inference model with a top mAP50 result on at least one dataset, but the relative consistency of AL models differs.
For performance across datasets and confidence thresholds, YOLOW \& YOLOE labels train inference models with relatively consistent mAP50 at all three $\alpha$ settings (except for $\alpha=0.8$ on LVIS), while GDINO mAP50 results dramatically decrease at $\alpha=0.8$ on all datasets. 
For performance across datasets when using a single confidence threshold, YOLOW \& YOLOE are consistent in that a single $\alpha=0.2$ setting always results in their best downstream inference model mAP50, while the best mAP50 setting for GDINO varies between $\alpha=0.2$ (BDD) and $\alpha=0.5$ (VOC \& COCO).

\begin{figure*}
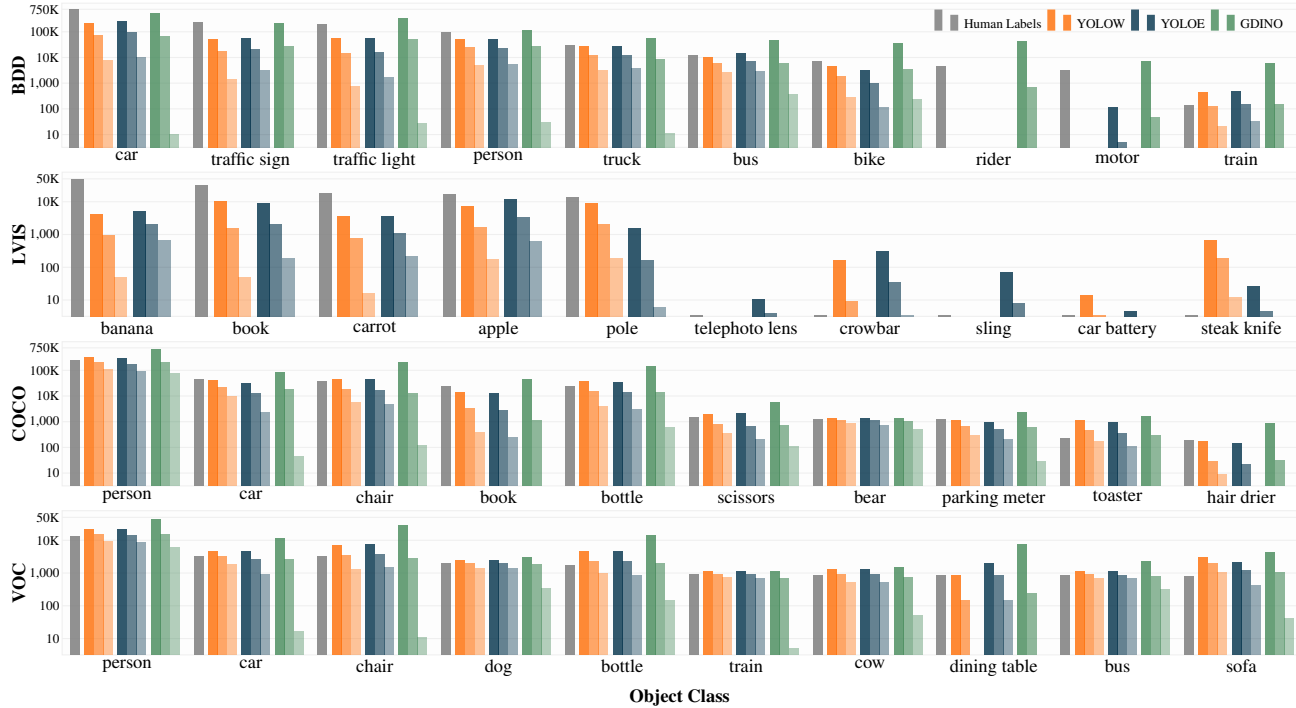

	\centering
	\begin{minipage}{0.975\textwidth}
		\input{paper-bdd-class-labels-20-50-80.tex}
	\end{minipage}%
	\\
	\vspace{-1.75em}
	\begin{minipage}{0.975\textwidth}
		\input{paper-lvis-class-labels-20-50-80.tex}
	\end{minipage}%
	\\
	\vspace{-1.75em}
	\begin{minipage}{0.975\textwidth}
		\input{paper-coco-class-labels-20-50-80.tex}
	\end{minipage}%
	\\
	\vspace{-1.75em}
	\begin{minipage}{0.975\textwidth}
		\input{paper-voc-class-labels-20-50-80.tex}
	\end{minipage}%
	\vspace{-2em}
	\caption{{\bf Number of Labels for the Five Most (left) and Least (right) Frequent Object Classes across all Datasets}.
		Each object class includes results for three confidence thresholds of 0.2, 0.5, \& 0.8 (left to right with increasing transparency) for each AL model.
	}
	\label{fig:class}
\end{figure*}

Finally, we compare the {\bf performance of individual AL configurations} in \cref{tab:f1map}.
As further evidence to avoid high confidence threshold configurations, inference models with the worst mAP50 for each dataset train on G-0.8 labels (GDINO, $\alpha=0.8$) for VOC, COCO, \& BDD and YW-0.8 labels (YOLOW) for LVIS.
In contrast, inference models with the best mAP50 for each dataset train on YW-0.2 labels for VOC \& COCO, YE-0.2 labels (YOLOE) for LVIS, and G-0.2 labels for BDD ($\alpha=0.2$ in all cases).
Notably, models trained on YW-0.2 labels are close to the best mAP50 for LVIS and BDD.
Thus, for experiments in \cref{tab:f1map}, we find that {\bf YOLOW with a confidence threshold of 0.2 is the single most reliable auto-labeling configuration} to train downstream inference models across all datasets.

\setlength{\tabcolsep}{1.65pt}
\begin{table}
	\centering
	\caption{{\bf Inference Model Class AP50 on VOC Validation}. 
		Classes are five most (top) and least (bottom) frequent in train set (see \cref{fig:class}).
		\% Diff.~is the mAP50 change from 5 Most to 5 Least.
		Bold font is best AL-trained YOLO11n result.
	}
	\vspace{-0.75em}
	\scriptsize
	\begin{tabular}{|r|rrr|rrr|rrr|c|}
		\hline
		\rowcolor{tableheader} & \multicolumn{3}{c|}{\bf YOLOW / $\alpha$} & \multicolumn{3}{c|}{\bf YOLOE / $\alpha$} & \multicolumn{3}{c|}{\bf GDINO / $\alpha$} & \bf Human \\
		\rowcolor{tableheader} \multicolumn{1}{|c|}{\bf Class} &  \multicolumn{1}{ c }{\bf 0.2} & \multicolumn{1}{ c }{\bf 0.5} & \multicolumn{1}{ c |}{\bf 0.8} & \multicolumn{1}{ c }{\bf 0.2} & \multicolumn{1}{ c }{\bf 0.5} & \multicolumn{1}{ c |}{\bf 0.8} &\multicolumn{1}{ c }{\bf 0.2} & \multicolumn{1}{ c }{\bf 0.5} & \multicolumn{1}{ c |}{\bf 0.8} & \bf Label \\ \hline
		person 	&	0.840	&	\bf 0.841	&	0.816	&	0.828	&	0.834	&	0.823	&	0.826	&	0.832	&	0.757	&	0.855	\\ \hline
		car 	& \bf 0.877	&	0.863	&	0.830	&	0.856	&	0.831	&	0.785	&	0.808	&	0.812	&	0.240	&	0.886	\\ \hline
		chair 	&	0.442	&	0.483	&	0.469	&	0.447	&	0.449	&	0.434	&	0.326	&	\bf 0.489	&	0.034	&	0.560	\\ \hline
		dog 	&	\bf 0.772	&	0.765	&	0.746	&	0.765	&	0.758	&	0.722	&	0.744	&	0.758	&	0.380	&	0.779	\\ \hline
		bottle 	&	0.586	&	0.578	&	0.512	&	\bf 0.593	&	0.578	&	0.521	&	0.429	&	0.592	&	0.142	&	0.602	\\ \hline \hline
		train 	&	0.841	&	\bf 0.859	&	0.834	&	0.845	&	0.856	&	0.814	&	0.773	&	0.826	&	0.083	&	0.856	\\ \hline
		cow 	&	0.706	&	0.706	&	0.706	&	0.703	&	0.638	&	0.633	&	0.639	&	\bf 0.710	&	0.263	&	0.727	\\ \hline
		dining table 	&	\bf 0.706	&	0.475	&	0.000	&	0.563	&	0.572	&	0.428	&	0.285	&	0.120	&	0.000	&	0.756	\\ \hline
		bus 	&	\bf 0.842	&	0.827	&	0.818	&	0.834	&	0.833	&	0.781	&	0.792	&	0.825	&	0.650	&	0.825	\\ \hline
		sofa 	&	0.585	&	0.608	&	0.654	&	0.621	&	0.658	&	0.584	&	0.615	&	\bf 0.692	&	0.235	&	0.725	\\ \hline \hline
		\rowcolor{tableheader}\multicolumn{11}{|c|}{\bf Mean Average Precision (mAP50)} \\ \hline
		5 Most &	0.703	&	\bf 0.706	&	0.675	&	0.698	&	0.690	&	0.657	&	0.627	&	0.697	&	0.311	&	0.736	\\ \hline
		5 Least &	\bf 0.736	&	0.695	&	0.603	&	0.713	&	0.711	&	0.648	&	0.621	&	0.635	&	0.246	&	0.778	\\ \hline
		\% Diff.	&	\bf 5\%	&	-2\%	&	-11\%	&	2\%	&	3\%	&	-1\%	&	-1\%	&	-9\%	&	-21\%	&	6\%	\\ \hline
		All 20	&	\bf 0.715	&	0.681	&	0.612	&	0.709	&	0.682	&	0.613	&	0.637	&	0.674	&	0.314	&	0.756	\\ \hline
	\end{tabular}
	\label{tab:voc}
\end{table}

\setlength{\tabcolsep}{1.4pt}
\begin{table}
	\centering
	\caption{{\bf Inference Model Class AP50 on COCO Validation}. 
	}
	\vspace{-0.75em}
	\scriptsize
	\begin{tabular}{|r|rrr|rrr|rrr|c|}
		\hline
		\rowcolor{tableheader} & \multicolumn{3}{c|}{\bf YOLOW / $\alpha$} & \multicolumn{3}{c|}{\bf YOLOE / $\alpha$} & \multicolumn{3}{c|}{\bf GDINO / $\alpha$} & \bf Human \\
		\rowcolor{tableheader} \multicolumn{1}{|c|}{\bf Class} &  \multicolumn{1}{ c }{\bf 0.2} & \multicolumn{1}{ c }{\bf 0.5} & \multicolumn{1}{ c |}{\bf 0.8} & \multicolumn{1}{ c }{\bf 0.2} & \multicolumn{1}{ c }{\bf 0.5} & \multicolumn{1}{ c |}{\bf 0.8} &\multicolumn{1}{ c }{\bf 0.2} & \multicolumn{1}{ c }{\bf 0.5} & \multicolumn{1}{ c |}{\bf 0.8} & \bf Label \\ \hline
		person 	&	0.693	&	0.695	&	0.643	&	0.694	&	0.680	&	0.612	&	0.679	&	\bf 0.711	&	0.603	&	0.729	\\ \hline
		car 	&	\bf 0.524	&	0.507	&	0.440	&	0.492	&	0.430	&	0.352	&	0.515	&	0.492	&	0.329	&	0.537	\\ \hline
		chair 	&	0.301	&	0.325	&	0.254	&	\bf 0.333	&	0.332	&	0.258	&	0.107	&	0.306	&	0.110	&	0.359	\\ \hline
		book 	&	0.076	&	0.052	&	0.049	&	0.060	&	0.046	&	0.035	&	\bf 0.086	&	0.049	&	0.000	&	0.179	\\ \hline
		bottle 	&	\bf 0.405	&	0.372	&	0.280	&	0.403	&	0.368	&	0.276	&	0.210	&	0.386	&	0.177	&	0.411	\\ \hline \hline
		scissors 	&	0.195	&	\bf 0.231	&	0.187	&	0.133	&	0.202	&	0.208	&	0.145	&	0.166	&	0.060	&	0.277	\\ \hline
		bear 	&	\bf 0.831	&	0.815	&	0.781	&	0.813	&	0.787	&	0.731	&	0.800	&	0.818	&	0.706	&	0.822	\\ \hline
		parking meter 	&	0.282	&	0.247	&	\bf 0.319	&	0.302	&	0.250	&	0.279	&	0.263	&	0.228	&	0.169	&	0.535	\\ \hline
		toaster 	&	\bf 0.483	&	0.190	&	0.461	&	0.201	&	0.315	&	0.300	&	0.253	&	0.287	&	0.000	&	0.297	\\ \hline
		hair drier 	&	0.003	&	0.000	&	0.000	&	0.004	&	0.000	&	0.000	&	\bf 0.058	&	0.000	&	0.000	&	0.001	\\ \hline \hline
		\rowcolor{tableheader}\multicolumn{11}{|c|}{\bf Mean Average Precision (mAP50)} \\ \hline
		 5 Most	&	\bf 0.400	&	0.390	&	0.333	&	0.397	&	0.371	&	0.307	&	0.319	&	0.389	&	0.244	&	0.443	\\ \hline
		 5	Least &	\bf 0.359	&	0.296	&	0.349	&	0.290	&	0.311	&	0.304	&	0.304	&	0.300	&	0.187	&	0.386	\\ \hline
		\% Difference	&	-10\%	&	-24\%	&	\bf 5\%	&	-27\%	&	-16\%	&	-1\%	&	-5\%	&	-23\%	&	-23\%	&	-13\%	\\ \hline
		All 80	&	\bf 0.460	&	0.445	&	0.404	&	0.458	&	0.441	&	0.387	&	0.424	&	0.432	&	0.244	&	0.496	\\ \hline
	\end{tabular}
	\label{tab:coco}
\end{table}

\subsection{Class Level Evaluation}
\label{sec:class}

\cref{sec:label}-\ref{sec:bp} evaluate auto-labeling at the dataset level.
However, we find that auto-labeling performance varies across individual object classes within datasets as well.

We compare the number of objects labels for the five most and least frequent classes across all datasets and AL models in \cref{fig:class}.
Notably, baseline label frequency is determined using human labels of each dataset train split, which we compare to AL frequency across confidence thresholds $\alpha=\{0.2,0.5,0.8\}$.
As expected, the number of auto-labels decreases with increasing confidence threshold (especially for GDINO), but there is generally less variation for common classes like ``person" (VOC, COCO, \& BDD) and more variation for rarer classes like ``hair drier" (COCO) or ``steak knife" (LVIS).
Furthermore, several classes with somewhat ambiguous class names like ``rider" (BDD) or ``sling" (LVIS) are left entirely unlabeled by some AL models regardless of $\alpha$.
Overall, the number of auto-labels and human labels best match on VOC, COCO, and the five most frequent BDD classes but have greater discrepancies on LVIS and the five least frequent BDD classes.

To understand the effect of label frequency on downstream inference model performance, we compare the average precision (AP50, \cref{sec:inf}) of the five most and least frequent object classes of each dataset in \cref{tab:voc}-\ref{tab:bdd}.

For {\bf VOC} (\cref{tab:voc}), the difference in AP50 for frequent and infrequent classes are relatively similar.
In fact, multiple YOLOW-, YOLOE-, and human label-trained inference models perform better on the infrequent classes.
Notably, some YOLOW- \& YOLOE-trained inference models outperform human label training on the infrequent ``train" and ``bus" classes.

For {\bf COCO} (\cref{tab:coco}), the AP50 is higher for all frequent classes relative to infrequent classes, except for YOLOW-0.8.
This change from VOC is likely due to COCO including rarer object classes.
Notably, the inference models with the best AP50 for the infrequent ``bear," ``toaster," and ``hair drier" classes all train on auto-labels.

\setlength{\tabcolsep}{4.25pt}
\begin{table}
	\centering
	\caption{{\bf Inference Model Class AP50 on LVIS Validation}. \\
		LVIS only uses 1,035 of the 1,203 training classes for validation, so we include results for the five most (top) and least (bottom) frequent train set classes that are \textit{also} in validation (see \cref{fig:class}).
	}
	\vspace{-0.75em}
	\scriptsize
	\begin{tabular}{|r|rrr|rrc|c|}
		\hline
		\rowcolor{tableheader} & \multicolumn{3}{c|}{\bf YOLOW / $\alpha$} & \multicolumn{3}{c|}{\bf YOLOE / $\alpha$} & \bf Human \\
		\rowcolor{tableheader} \multicolumn{1}{|c|}{\bf Class} &  \multicolumn{1}{ c }{\bf 0.2} & \multicolumn{1}{ c }{\bf 0.5} & \multicolumn{1}{ c |}{\bf 0.8} & \multicolumn{1}{ c }{\bf 0.2} & \multicolumn{1}{ c }{\bf 0.5} & \multicolumn{1}{ c |}{\bf 0.8}  & \bf Label \\ \hline
		banana 	&	\bf 0.098	&	0.068	&	0.049	&	0.079	&	0.060	&	0.059	&	0.508	\\ \hline
		book 	&	\bf 0.029	&	0.018	&	0.025	&	0.025	&	0.017	&	0.022	&	0.201	\\ \hline
		carrot 	&	\bf 0.228	&	0.159	&	0.100	&	0.211	&	0.151	&	0.097	&	0.358	\\ \hline
		apple 	&	0.299	&	0.137	&	0.080	&	\bf 0.309	&	0.210	&	0.128	&	0.417	\\ \hline
		pole/post 	&	0.009	&	\bf 0.011	&	0.009	&	0.010	&	0.008	&	0.008	&	0.016	\\ \hline \hline
		subwoofer 	&	0.017	&	0.030	&	0.016	&	0.013	&	\bf 0.055	&	0.000	&	0.000	\\ \hline
		string cheese 	&	0.000	&	0.000	&	0.000	&	0.000	&	0.000	&	0.000	&	0.000	\\ \hline
		milkshake 	&	0.000	&	0.000	&	0.000	&	0.000	&	0.000	&	0.000	&	0.000	\\ \hline
		vinegar 	&	0.000	&	0.000	&	0.000	&	0.000	&	0.000	&	0.000	&	0.000	\\ \hline
		eye dropper	&	0.000	&	0.000	&	0.000	&	0.000	&	0.000	&	0.000	&	0.000	\\ \hline \hline
		\rowcolor{tableheader}\multicolumn{8}{|c|}{\bf Mean Average Precision (mAP50)} \\ \hline
		5	Most &	\bf 0.133	&	0.079	&	0.053	&	0.127	&	0.089	&	0.063	&	0.300	\\ \hline
		5	Least &	0.003	&	0.006	&	0.003	&	0.003	&	\bf 0.011	&	0.000	&	0.000	\\ \hline
		\% Difference	&	-97\%	&	-92\%	&	-94\%	&	-98\%	&	\bf -88\%	&	-100\%	&	-100\%	\\  \hline
		All 1,035 in Val.	&	0.059	&	0.050	&	0.031	&	\bf 0.064	&	0.057	&	0.040	&	0.087	\\ \hline
	\end{tabular}
	\label{tab:lvis}
\end{table}

For {\bf LVIS} (\cref{tab:lvis}), which has the most classes of any dataset, the relative performance decrease between the five most and least frequent classes is greater than on any other dataset.
Furthermore, the AP50 for basically all of the 1,203 object classes is lower than with VOC and COCO. 
For frequent classes, human label training has an $\text{AP50}>0.2$ for 4 classes, while AL-based training has an $\text{AP50}>0.2$ for only 2 classes.
For infrequent classes, the only $\text{AP50}>0$ is from AL-trained models for the ``subwoofer" class.

For {\bf BDD} (\cref{tab:bdd}), the absolute AP50 decrease between the frequent and infrequent classes is greater than on any other dataset.
This is particularly the case for AL-trained inference models on the ``rider," ``motor," and ``train" classes, which are completely unlabeled in the train set for some AL configurations (\cref{fig:class}). 
Notably, this lower AP50 on infrequent BDD classes helps explain the mAP50 performance gap between the AL- and human label-trained inference models in \cref{fig:lvis}.
In practice, missing auto-labels could likely be addressed by using more descriptive class names for the AL model input text prompt ($\sT$ in \cref{eq:fi}).

\setlength{\tabcolsep}{1.75pt}
\begin{table}
	\centering
	\caption{{\bf Inference Model Class AP50 on BDD Validation}. 
		Classes ordered by decreasing frequency in train set (see \cref{fig:class}).
	}
	\vspace{-0.75em}
	\scriptsize
	\begin{tabular}{|r|rrr|rrr|rrr|c|}
		\hline
		\rowcolor{tableheader} & \multicolumn{3}{c|}{\bf YOLOW / $\alpha$} & \multicolumn{3}{c|}{\bf YOLOE / $\alpha$} & \multicolumn{3}{c|}{\bf GDINO / $\alpha$} & \bf Human \\
		\rowcolor{tableheader} \multicolumn{1}{|c|}{\bf Class} &  \multicolumn{1}{ c }{\bf 0.2} & \multicolumn{1}{ c }{\bf 0.5} & \multicolumn{1}{ c |}{\bf 0.8} & \multicolumn{1}{ c }{\bf 0.2} & \multicolumn{1}{ c }{\bf 0.5} & \multicolumn{1}{ c |}{\bf 0.8} &\multicolumn{1}{ c }{\bf 0.2} & \multicolumn{1}{ c }{\bf 0.5} & \multicolumn{1}{ c |}{\bf 0.8} & \bf Label \\ \hline
		car 	&	0.524	&	0.493	&	0.488	&	0.531	&	0.508	&	0.505	&	\bf 0.607	&	0.465	&	0.000	&	0.739	\\ \hline
		traffic sign 	&	0.373	&	0.328	&	0.373	&	0.353	&	0.331	&	0.375	&	\bf 0.456	&	0.360	&	0.000	&	0.544	\\ \hline
		traffic light 	&	0.278	&	0.274	&	0.277	&	0.246	&	0.243	&	0.271	&	\bf 0.365	&	0.358	&	0.000	&	0.501	\\ \hline
		person 	&	\bf 0.427	&	0.390	&	0.319	&	0.395	&	0.355	&	0.311	&	0.327	&	0.400	&	0.000	&	0.511	\\ \hline
		truck 	&	0.407	&	0.428	&	0.422	&	0.390	&	0.408	&	0.407	&	0.417	&	\bf 0.438	&	0.000	&	0.556	\\ \hline
		bus 	&	\bf 0.448	&	0.430	&	0.407	&	0.394	&	0.388	&	0.372	&	0.340	&	0.421	&	0.000	&	0.528	\\ \hline
		bike 	&	0.249	&	0.247	&	0.185	&	0.209	&	0.204	&	0.164	&	\bf 0.251	&	\bf 0.251	&	0.000	&	0.333	\\ \hline
		rider 	&	0.000	&	0.000	&	0.000	&	0.000	&	0.000	&	0.000	&	0.031	&	\bf 0.044	&	0.000	&	0.313	\\ \hline
		motor 	&	0.000	&	0.000	&	0.000	&	0.030	&	0.000	&	0.000	&	0.006	&	\bf 0.044	&	0.000	&	0.320	\\ \hline
		train 	&	\bf 0.001	&	0.000	&	0.000	&	\bf 0.001	&	0.000	&	0.000	&	\bf 0.001	&	\bf 0.001	&	0.000	&	0.000	\\ \hline \hline
		\rowcolor{tableheader}\multicolumn{11}{|c|}{\bf Mean Average Precision (mAP50)} \\ \hline
		 5	Most &	0.402	&	0.383	&	0.376	&	0.383	&	0.369	&	0.374	&	\bf 0.435	&	0.404	&	0.000	&	0.570	\\ \hline
		5 Least &	0.140	&	0.135	&	0.118	&	0.127	&	0.119	&	0.107	&	0.126	&	\bf 0.152	&	0.000	&	0.299	\\ \hline
		\% Diff.	&	-65\%	&	-65\%	&	-69\%	&	-67\%	&	-68\%	&	-71\%	&	-71\%	&	\bf -62\%	&	\multicolumn{1}{c|}{-}	&	-48\%	\\ \hline
		All	10 &	0.271	&	0.259	&	0.247	&	0.255	&	0.244	&	0.240	&	\bf 0.280	&	0.278	&	0.000	&	0.434	\\ \hline
	\end{tabular}
	\label{tab:bdd}
\end{table}

\begin{figure*}
	\centering
	\begin{minipage}{0.25\textwidth}
		\includegraphics[width=0.975\textwidth]{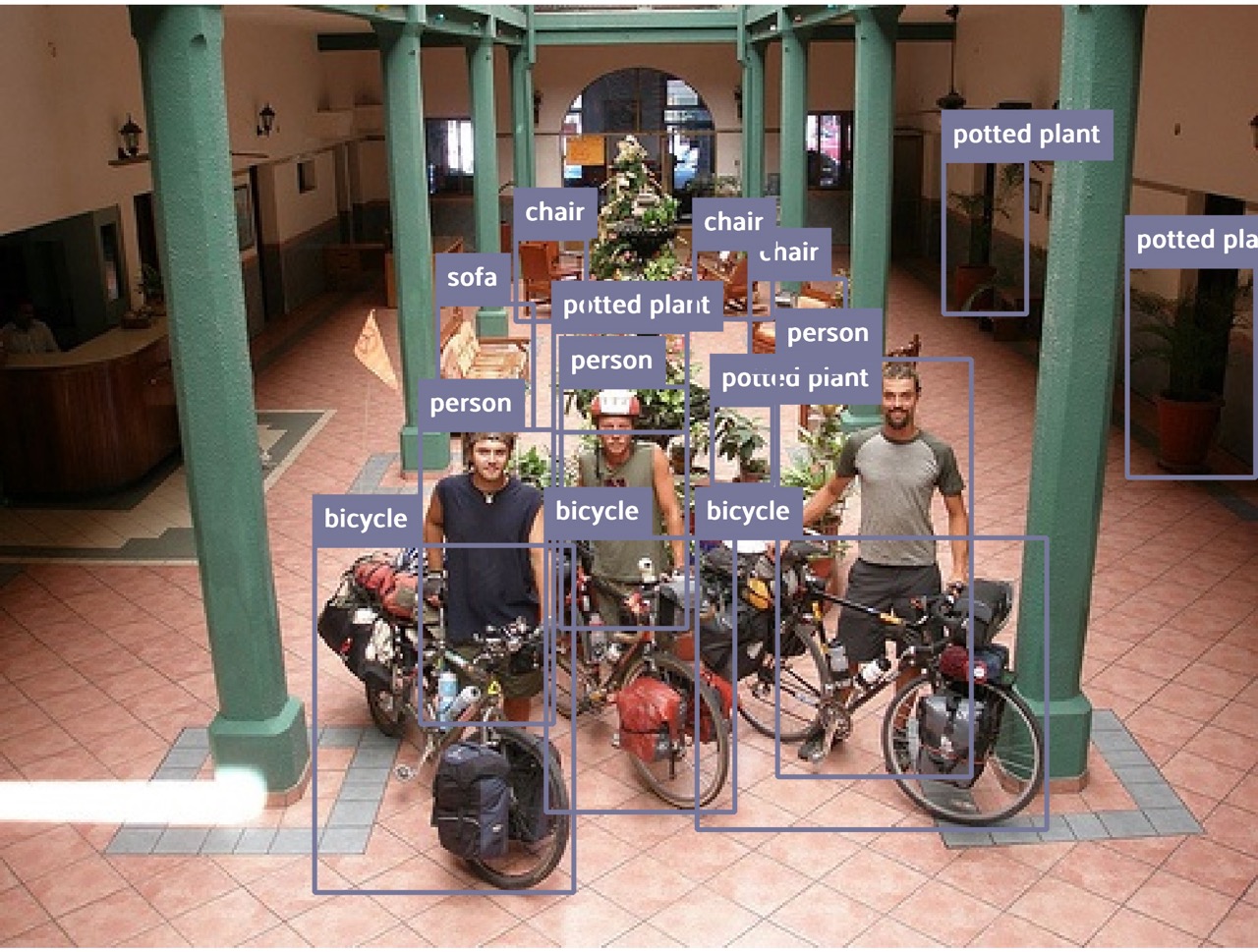}
	\end{minipage}%
	\begin{minipage}{0.25\textwidth}
		\includegraphics[width=0.975\textwidth]{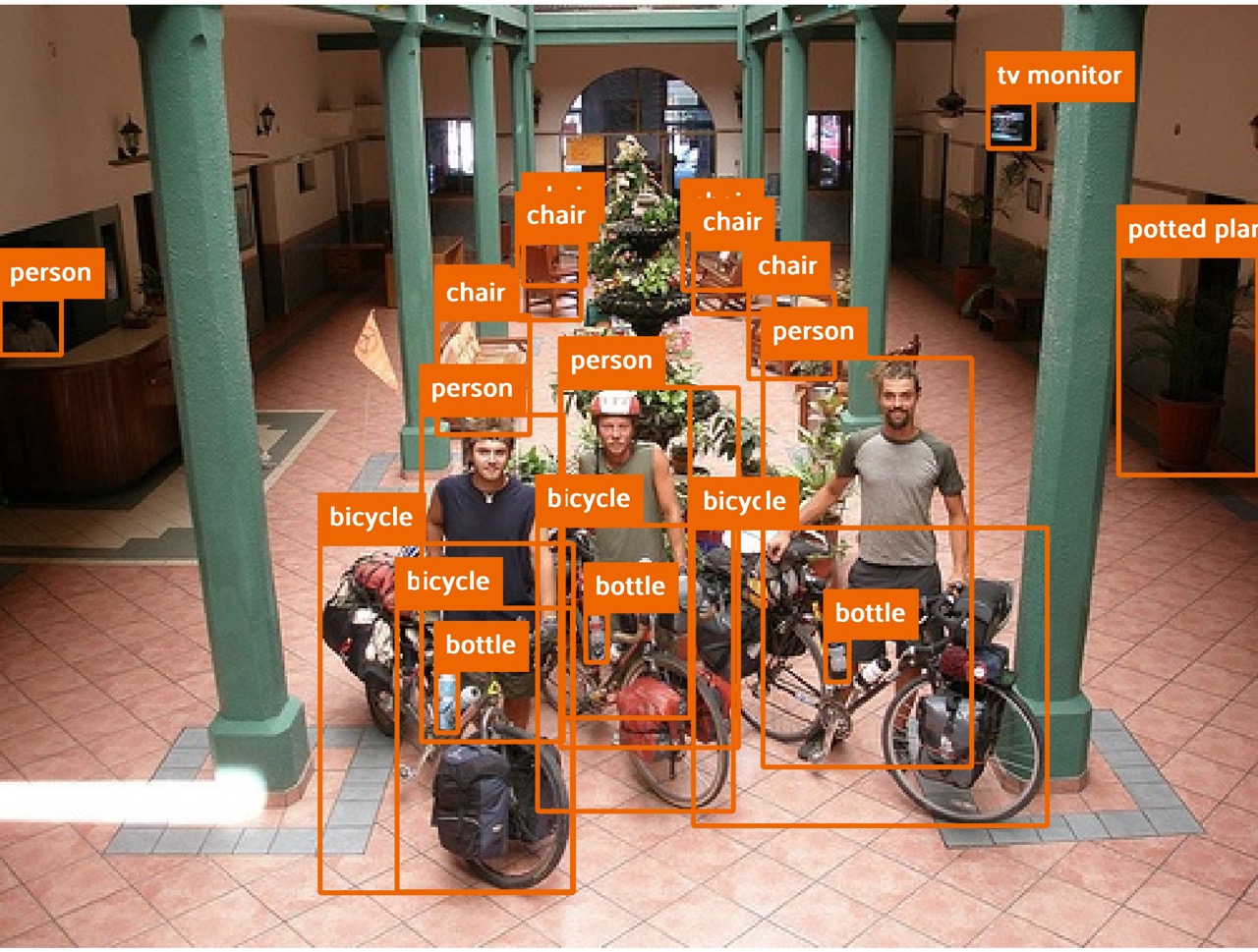}
	\end{minipage}%
	\begin{minipage}{0.25\textwidth}
		\includegraphics[width=0.975\textwidth]{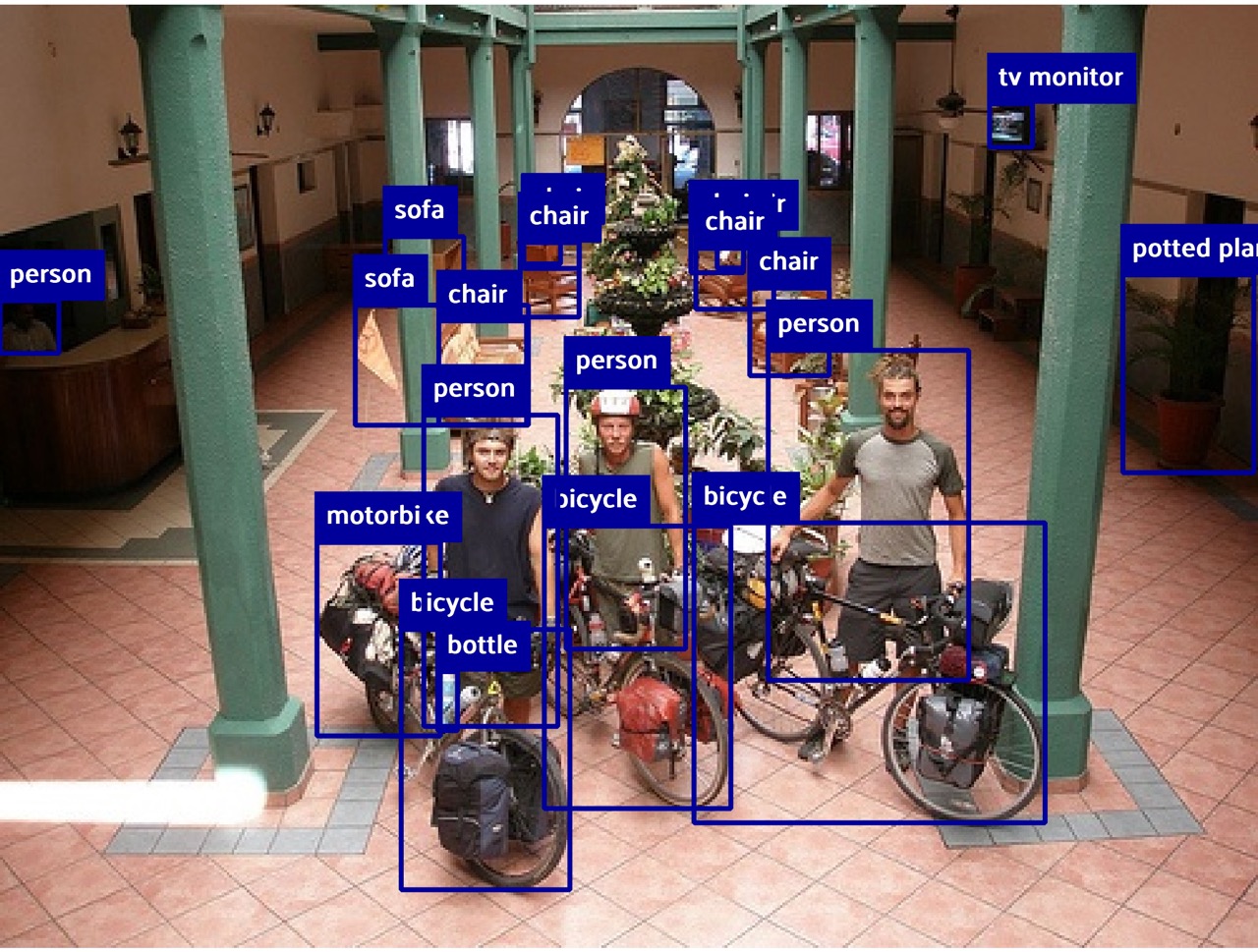}
	\end{minipage}%
	\begin{minipage}{0.25\textwidth}
		\includegraphics[width=0.975\textwidth]{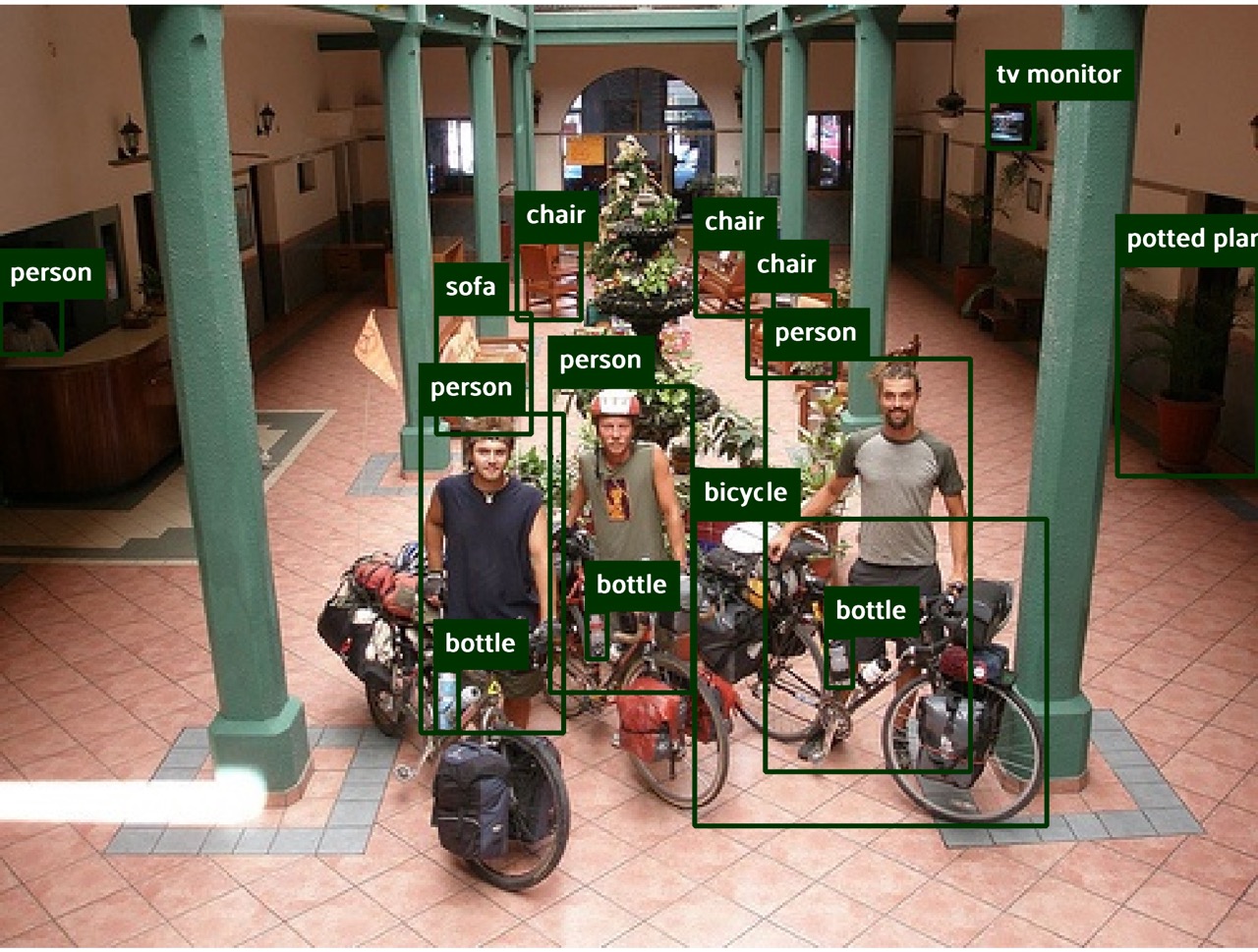}
	\end{minipage}%
	\\ 	\vspace{0.1em}
	\begin{minipage}{0.25\textwidth}
		\includegraphics[width=0.975\textwidth]{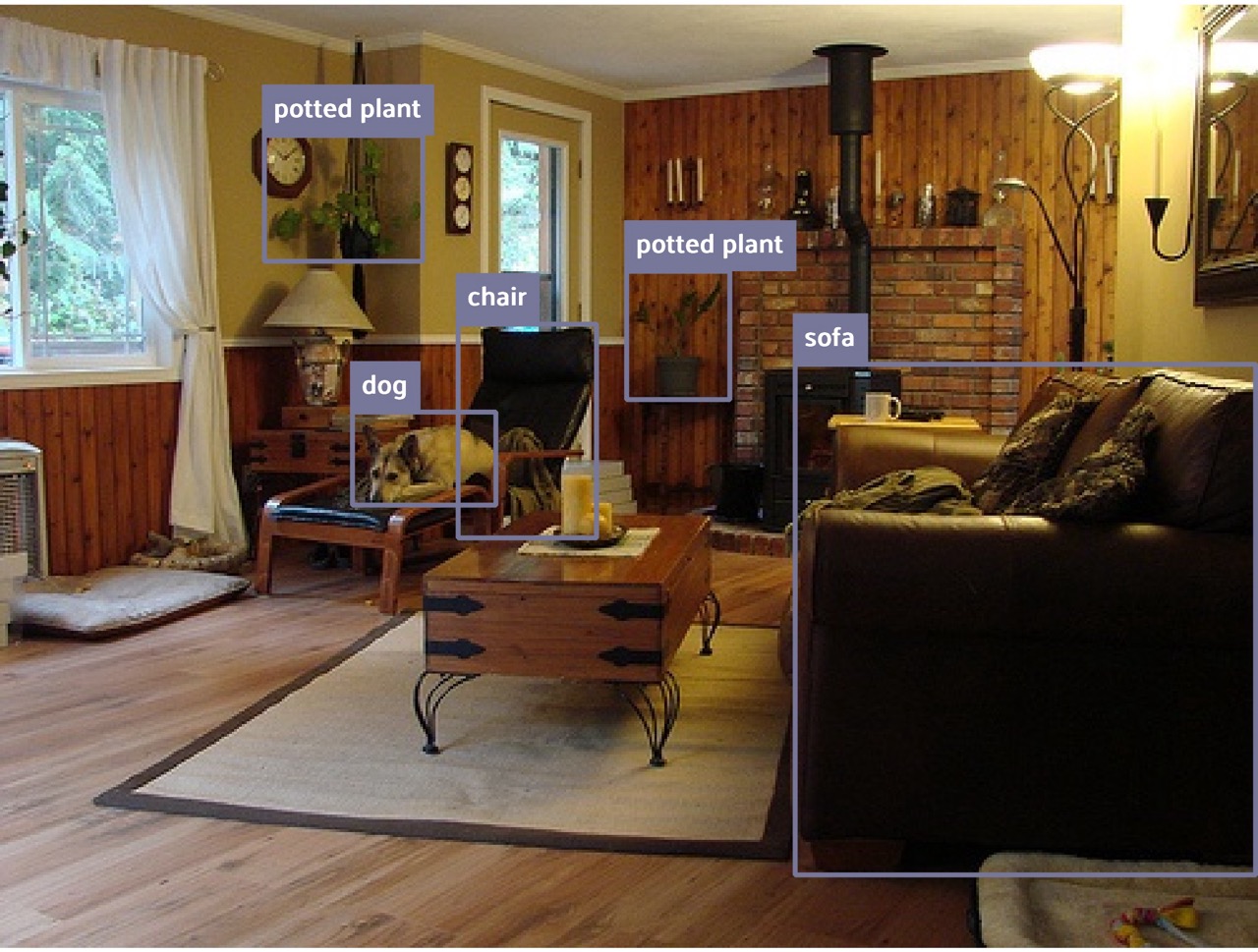}
	\end{minipage}%
	\begin{minipage}{0.25\textwidth}
		\includegraphics[width=0.975\textwidth]{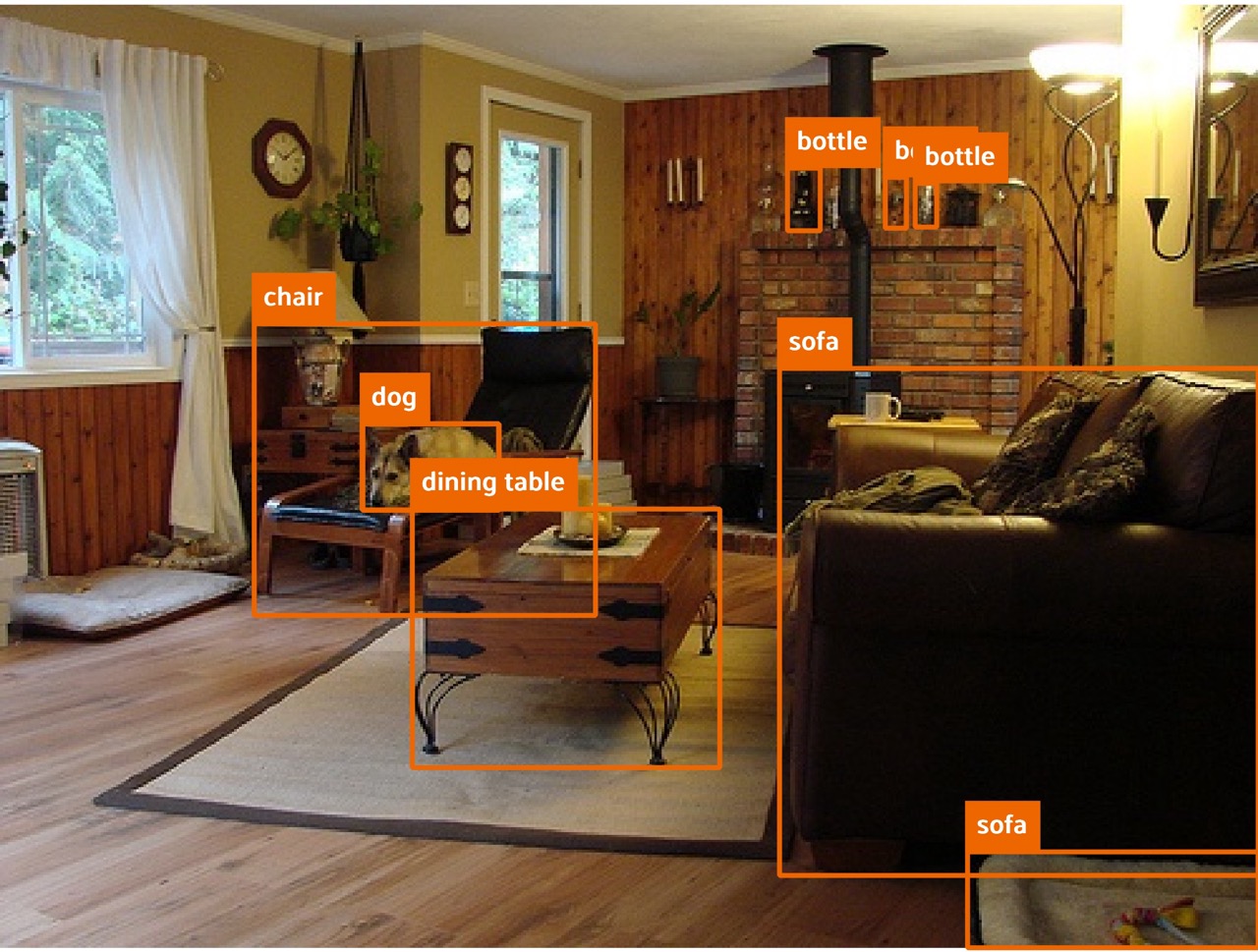}
	\end{minipage}%
	\begin{minipage}{0.25\textwidth}
		\includegraphics[width=0.975\textwidth]{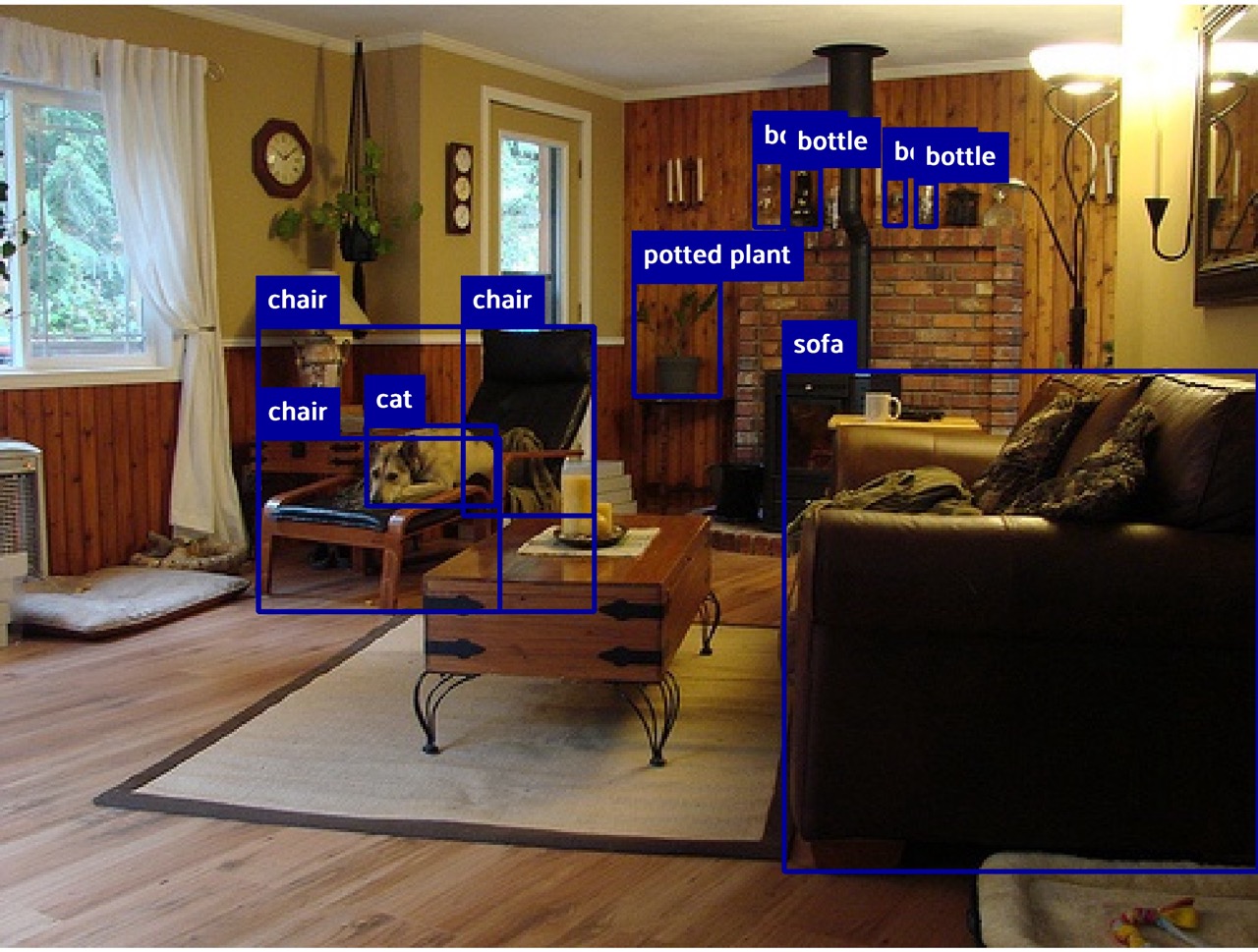}
	\end{minipage}%
	\begin{minipage}{0.25\textwidth}
		\includegraphics[width=0.975\textwidth]{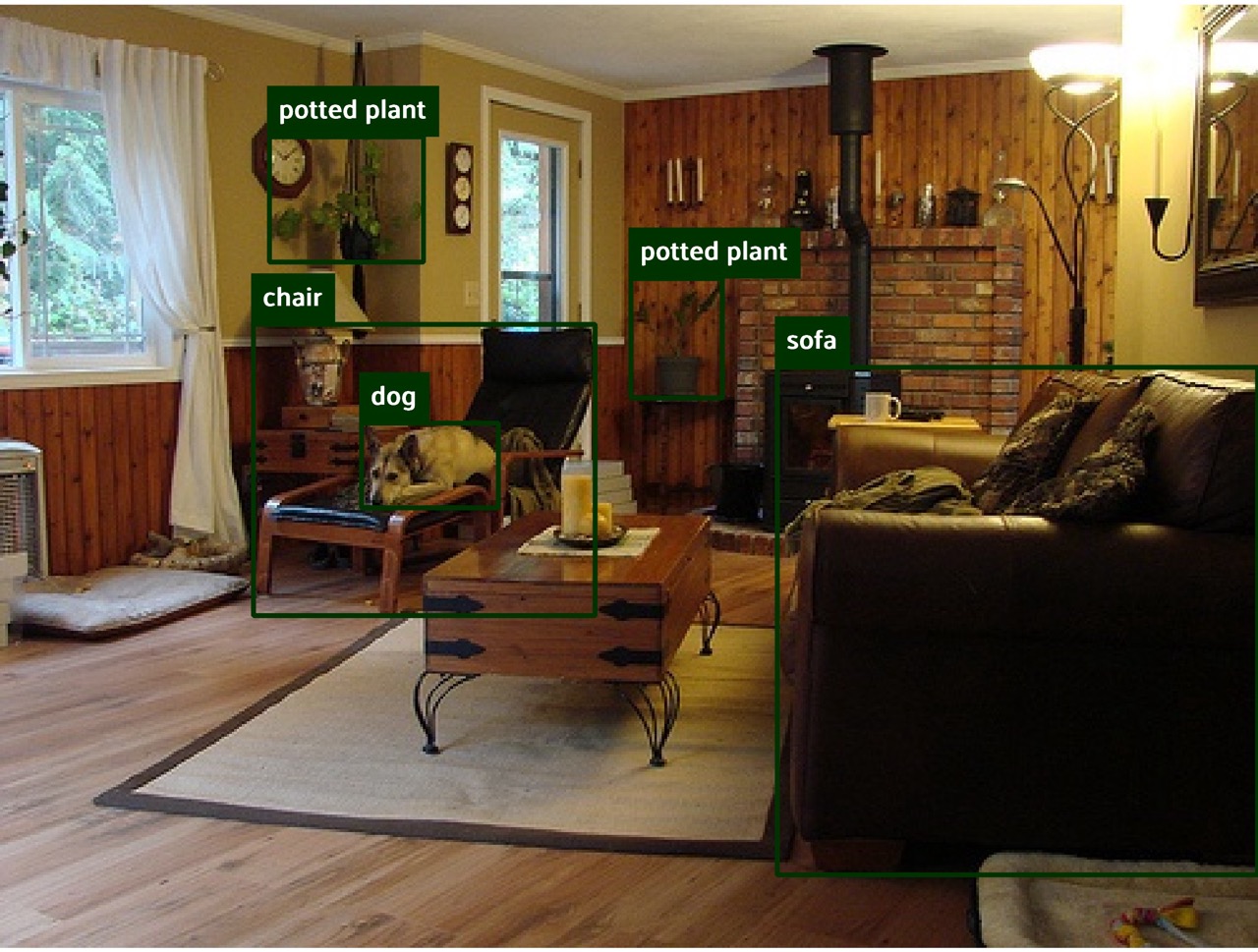}
	\end{minipage}%
	\\ \vspace{0.2em}
	\begin{minipage}{0.25\textwidth}
		\includegraphics[width=0.975\textwidth,trim={14.1cm 0 14.1cm 0}, clip]{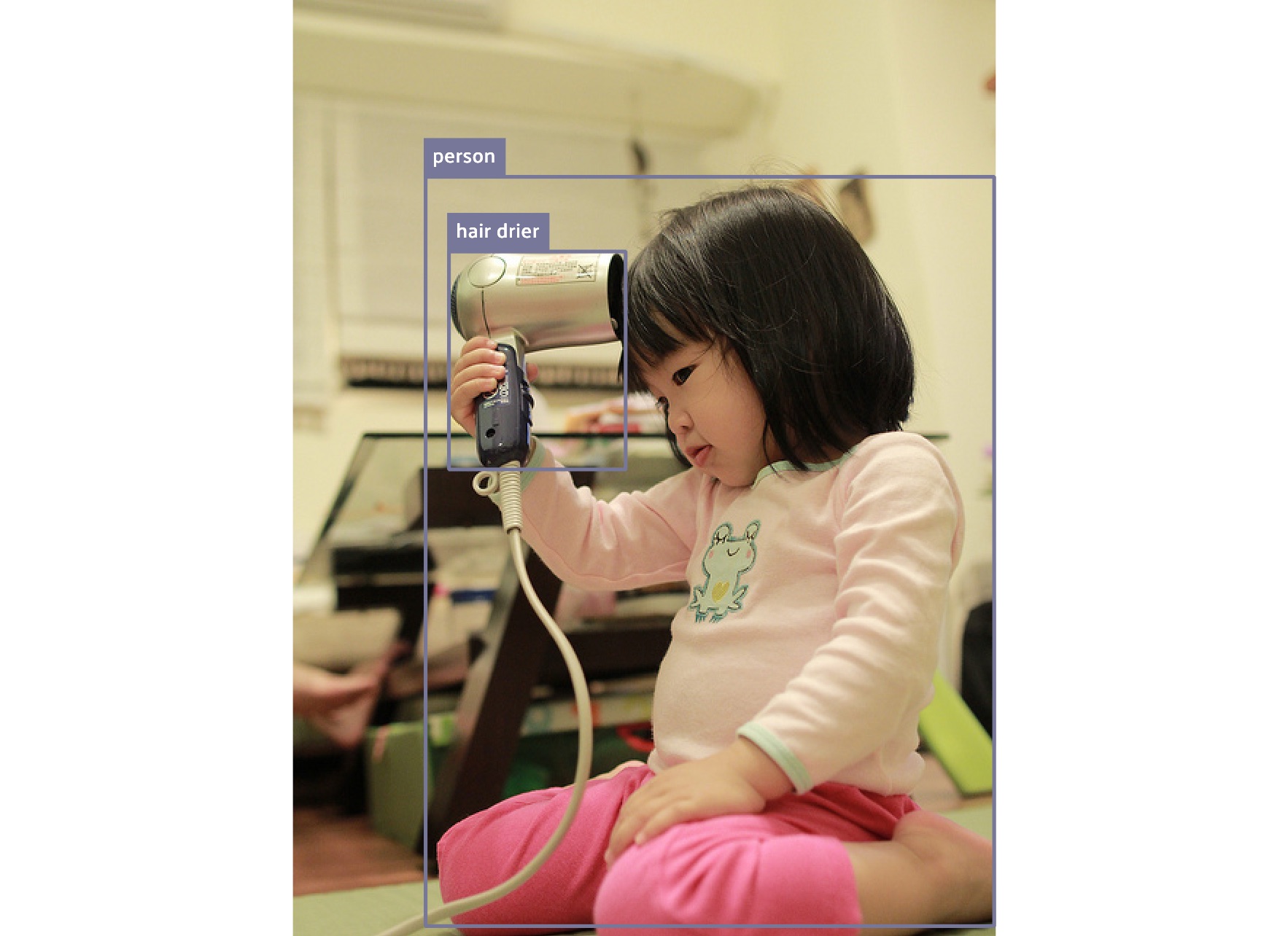}
	\end{minipage}%
	\begin{minipage}{0.25\textwidth}
		\includegraphics[width=0.975\textwidth,trim={14.1cm 0 14.1cm 0}, clip]{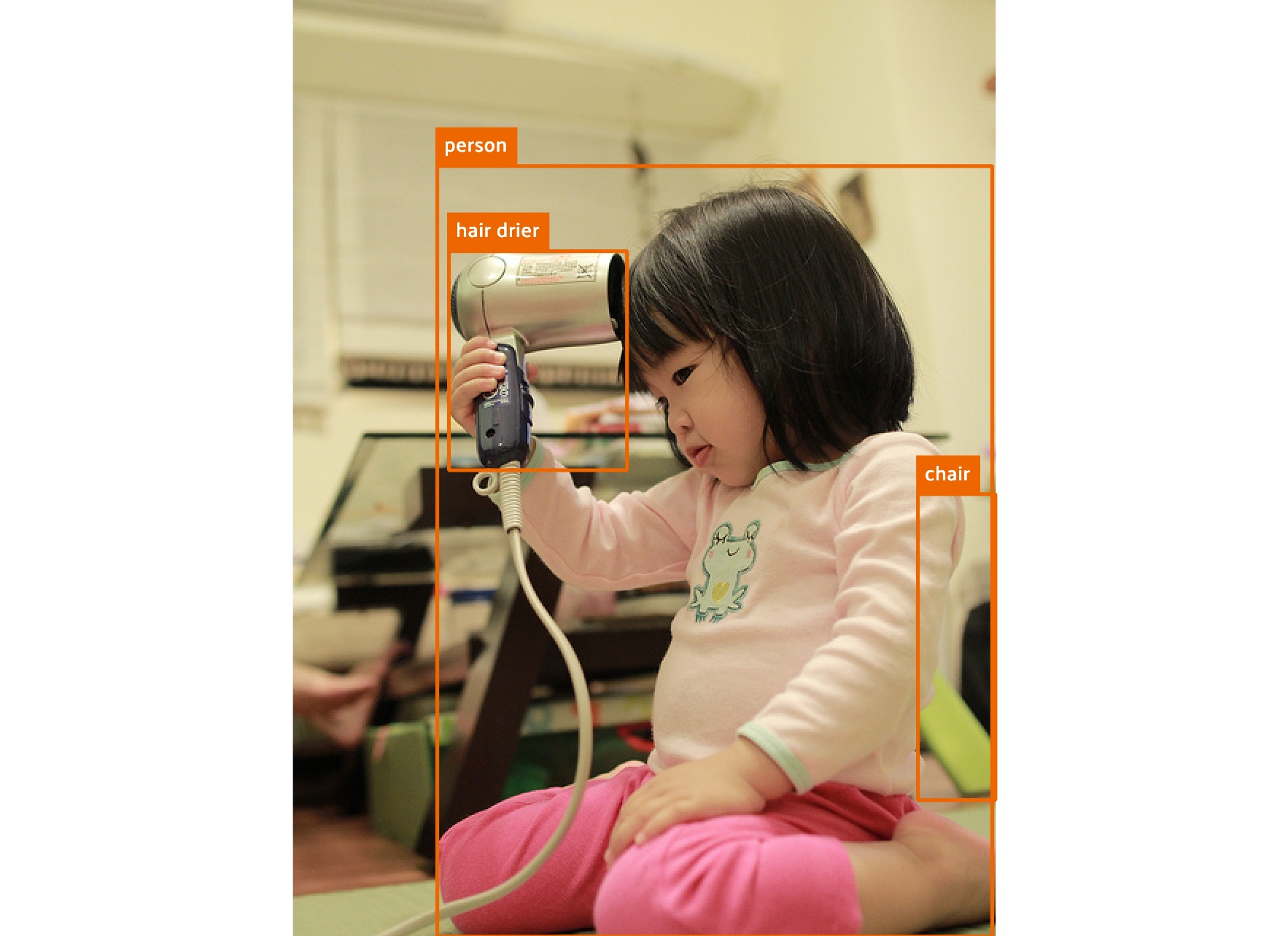}
	\end{minipage}%
	\begin{minipage}{0.25\textwidth}
		\includegraphics[width=0.975\textwidth,trim={14.1cm 0 14.1cm 0}, clip]{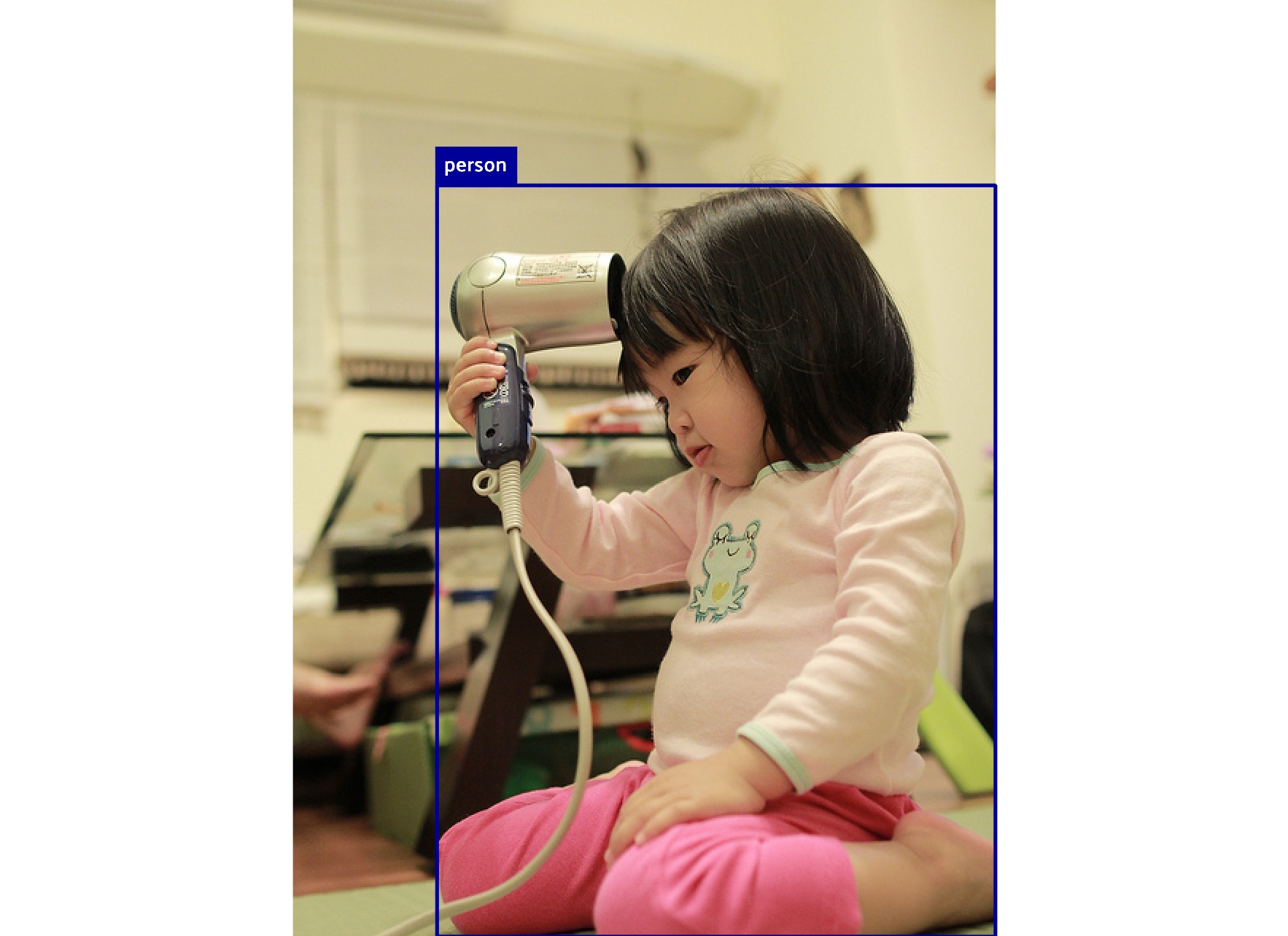}
	\end{minipage}%
	\begin{minipage}{0.25\textwidth}
		\includegraphics[width=0.975\textwidth,trim={14.1cm 0 14.1cm 0}, clip]{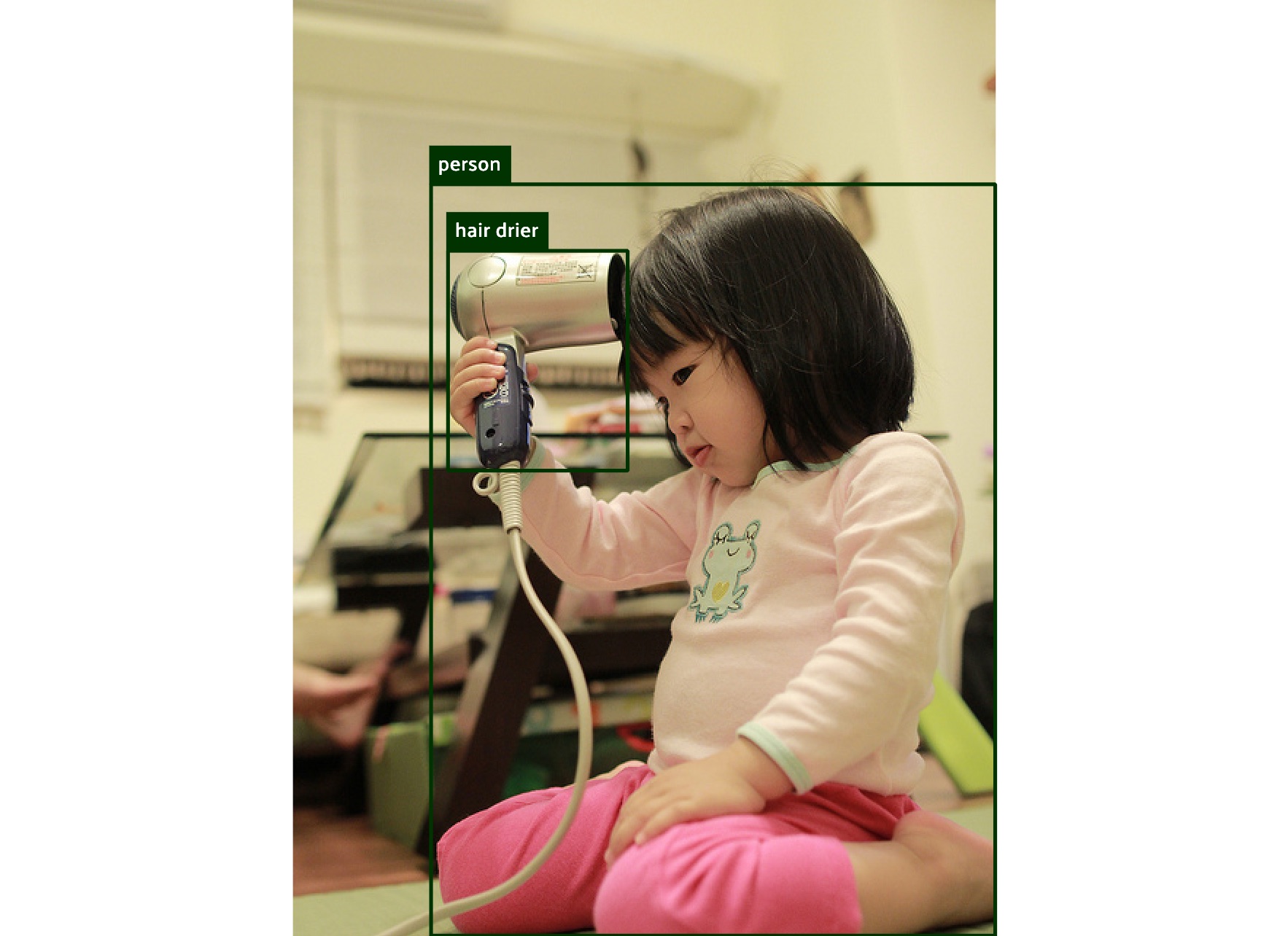}
	\end{minipage}%
	\\ 	\vspace{0.2em}
	\begin{minipage}{0.25\textwidth}
		\includegraphics[width=0.975\textwidth,trim={2.45cm 0 2.45cm 0}, clip]{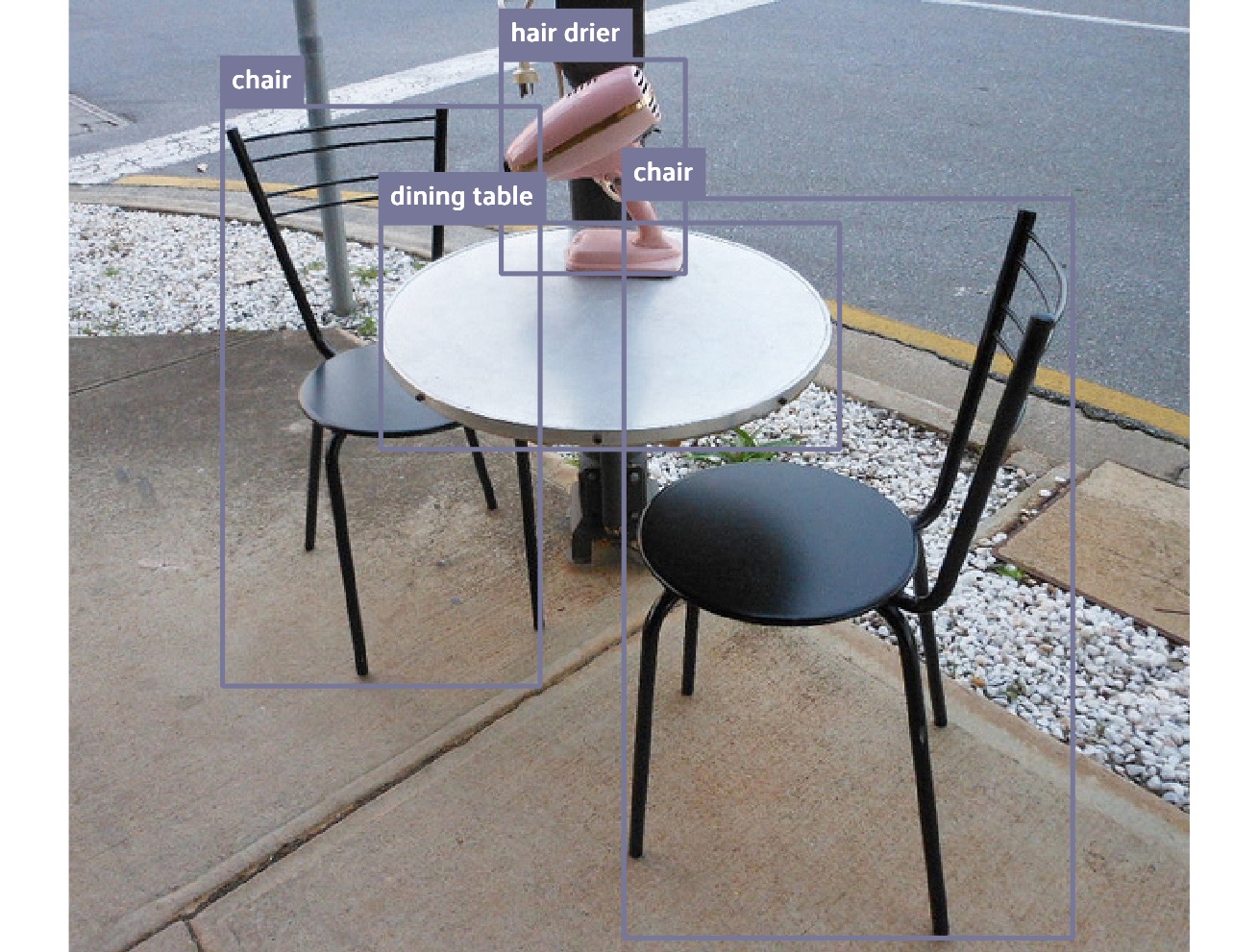}
	\end{minipage}%
	\begin{minipage}{0.25\textwidth}
		\includegraphics[width=0.975\textwidth,trim={2.45cm 0 2.45cm 0}, clip]{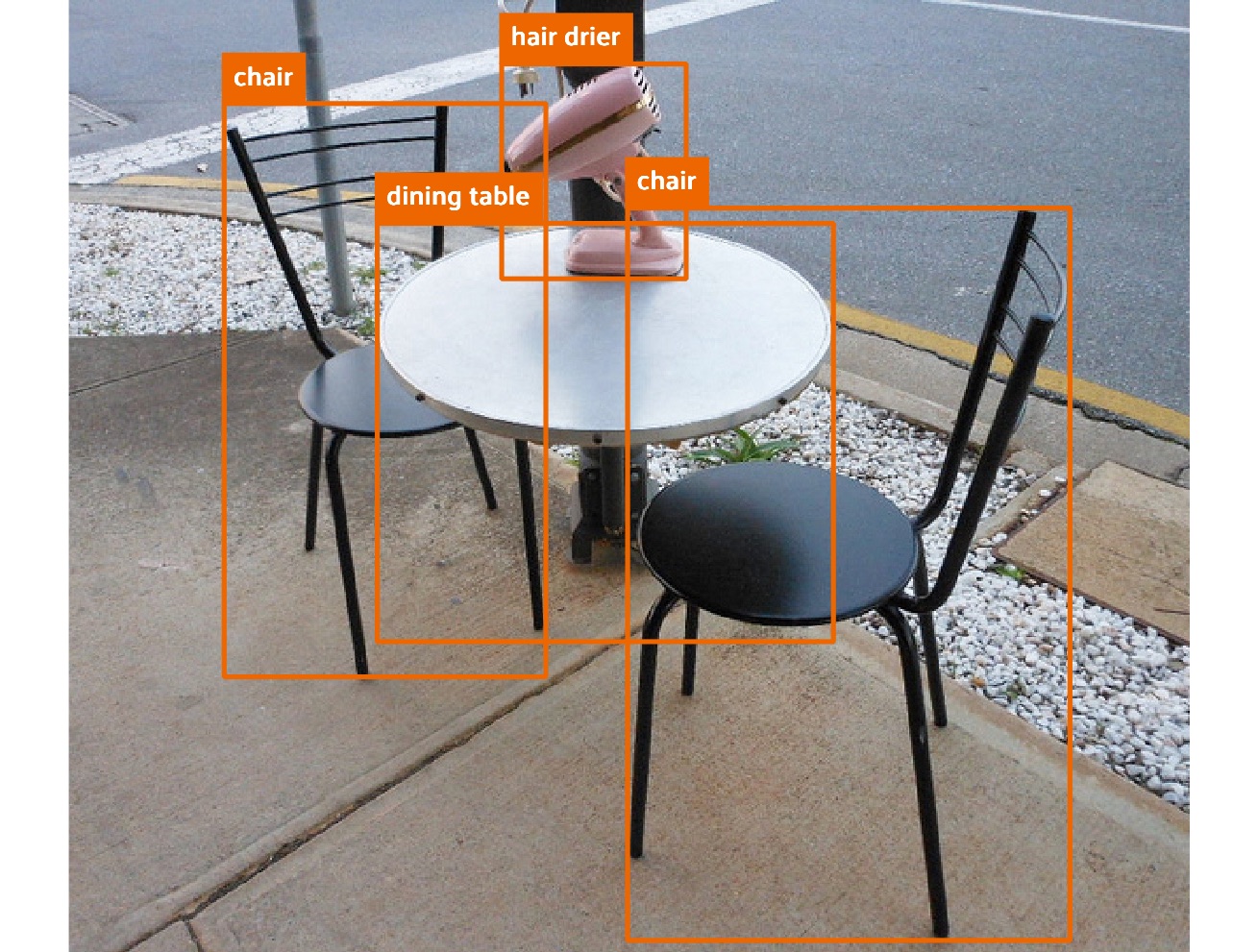}
	\end{minipage}%
	\begin{minipage}{0.25\textwidth}
		\includegraphics[width=0.975\textwidth,trim={2.45cm 0 2.45cm 0}, clip]{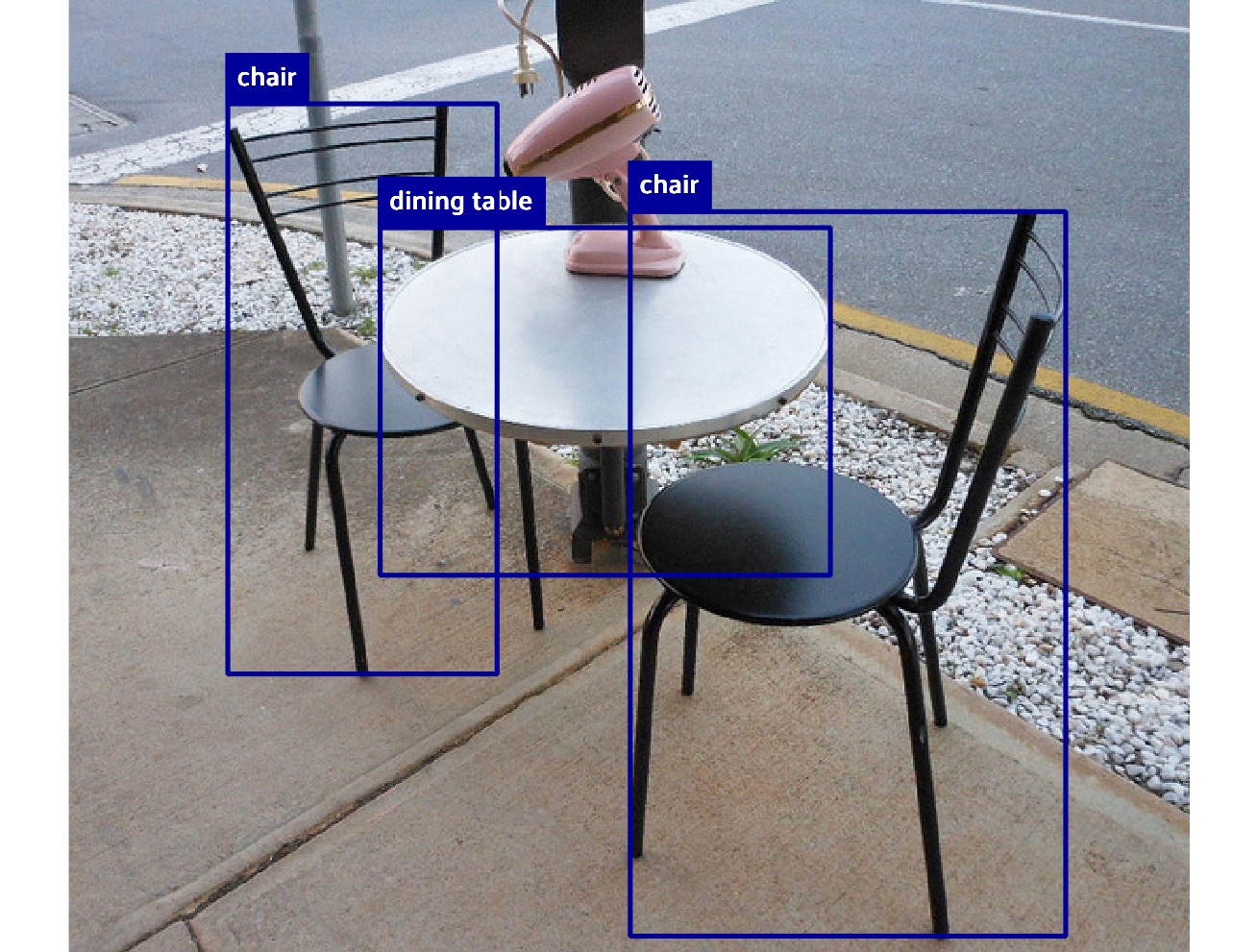}
	\end{minipage}%
	\begin{minipage}{0.25\textwidth}
		\includegraphics[width=0.975\textwidth,trim={2.45cm 0 2.45cm 0}, clip]{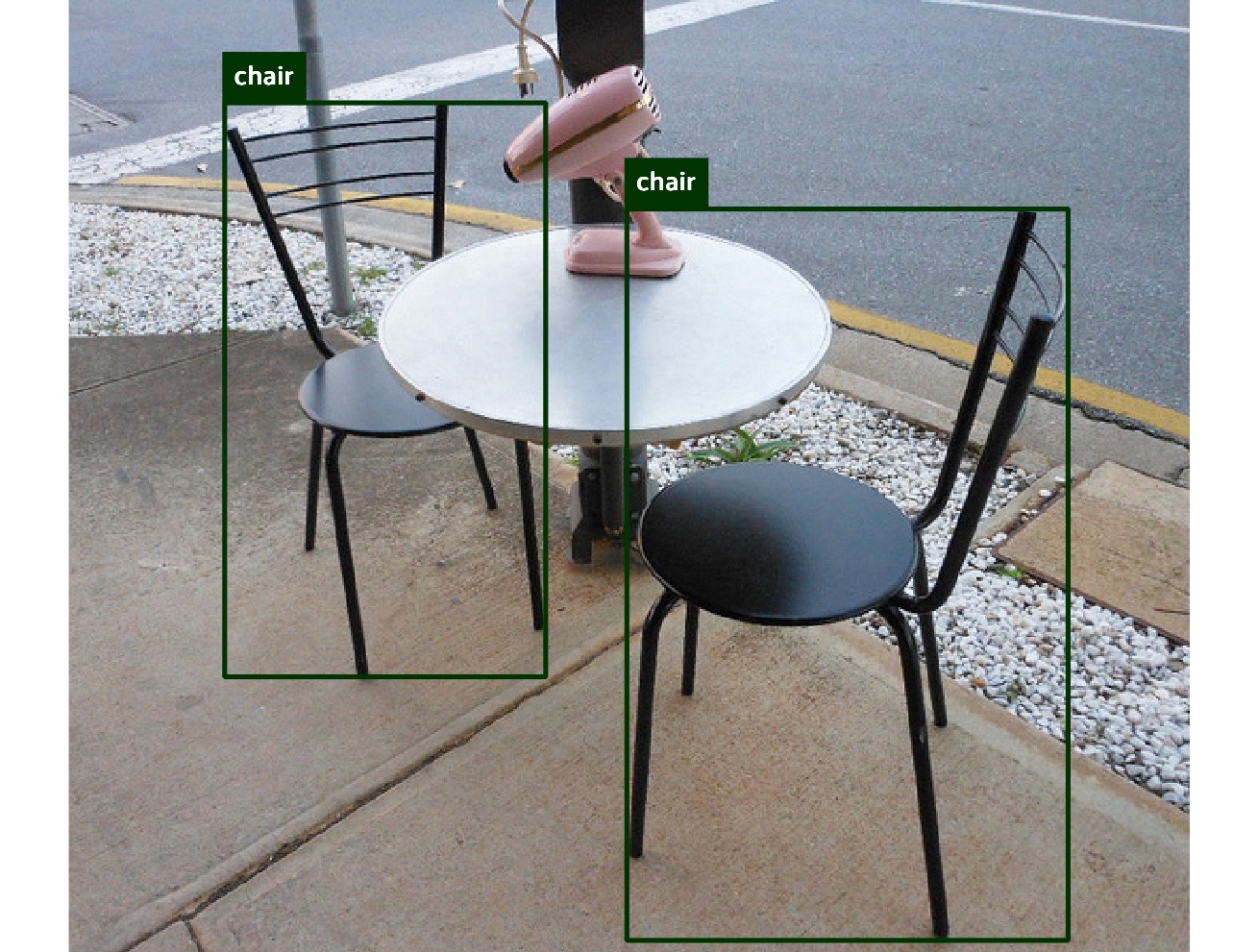}
	\end{minipage}%
	\vspace{-0.75em}
	\caption{{\bf Qualitative Comparison of Auto-Labels with Least Frequent Classes on VOC \& COCO}. 
		All images include least frequent train set class for VOC (sofa, top two rows) and COCO (hair drier, rows 3-4). 
		Label sources are (left to right) human (grey), YOLOW-0.2 (0.2 confidence threshold, orange), YOLOE-0.2 (blue), and GDINO-0.5 (green). 
		Visualizations generated using the FiftyOne Library \cite{moore2020fiftyone}.
	}
	\label{fig:qualvoc}
\end{figure*}

\begin{figure*}
	\centering
	\begin{minipage}{0.25\textwidth}
		\includegraphics[width=0.975\textwidth,trim={5.4cm 0 5.4cm 0}, clip]{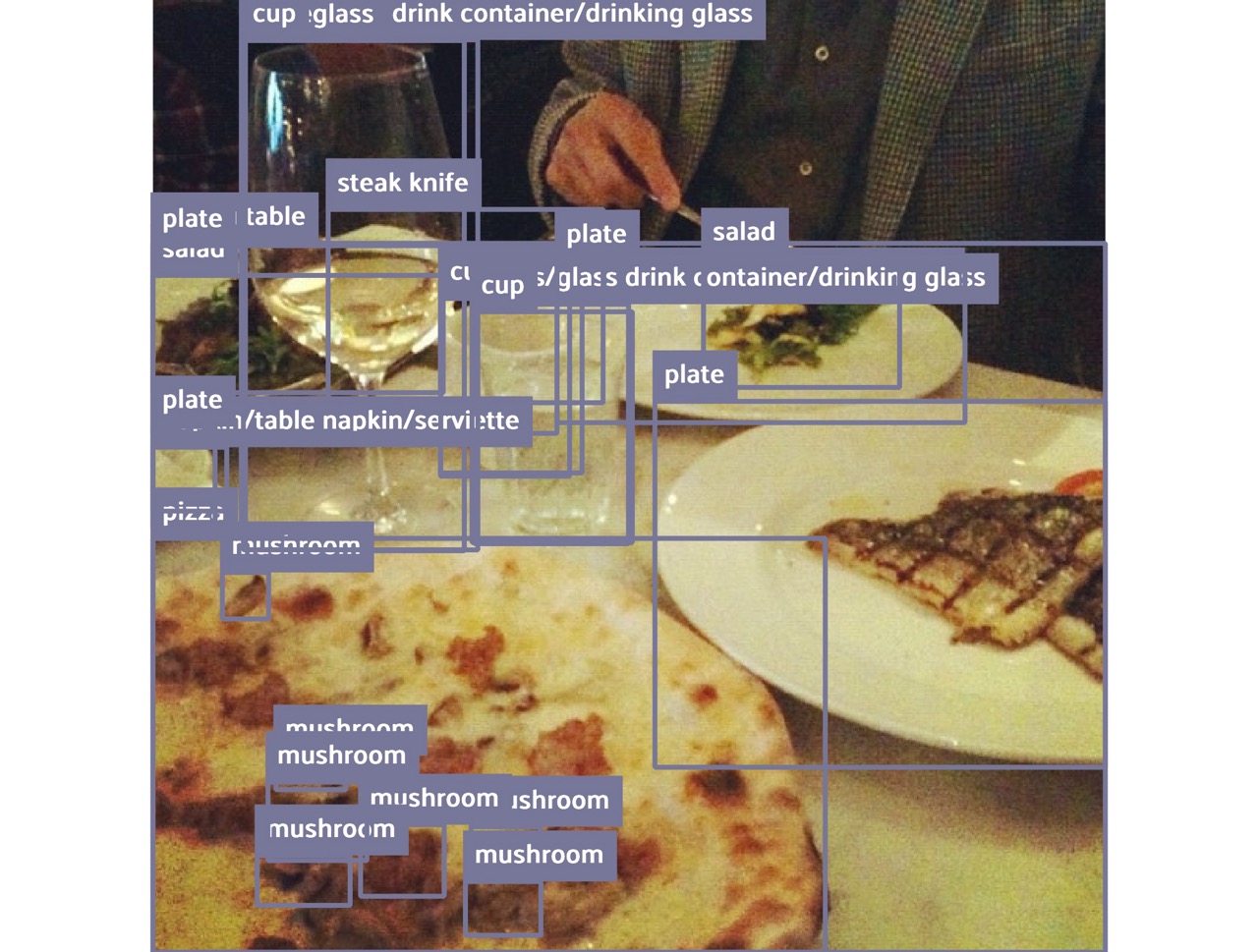}
	\end{minipage}%
	\begin{minipage}{0.25\textwidth}
		\includegraphics[width=0.975\textwidth,trim={5.4cm 0 5.4cm 0}, clip]{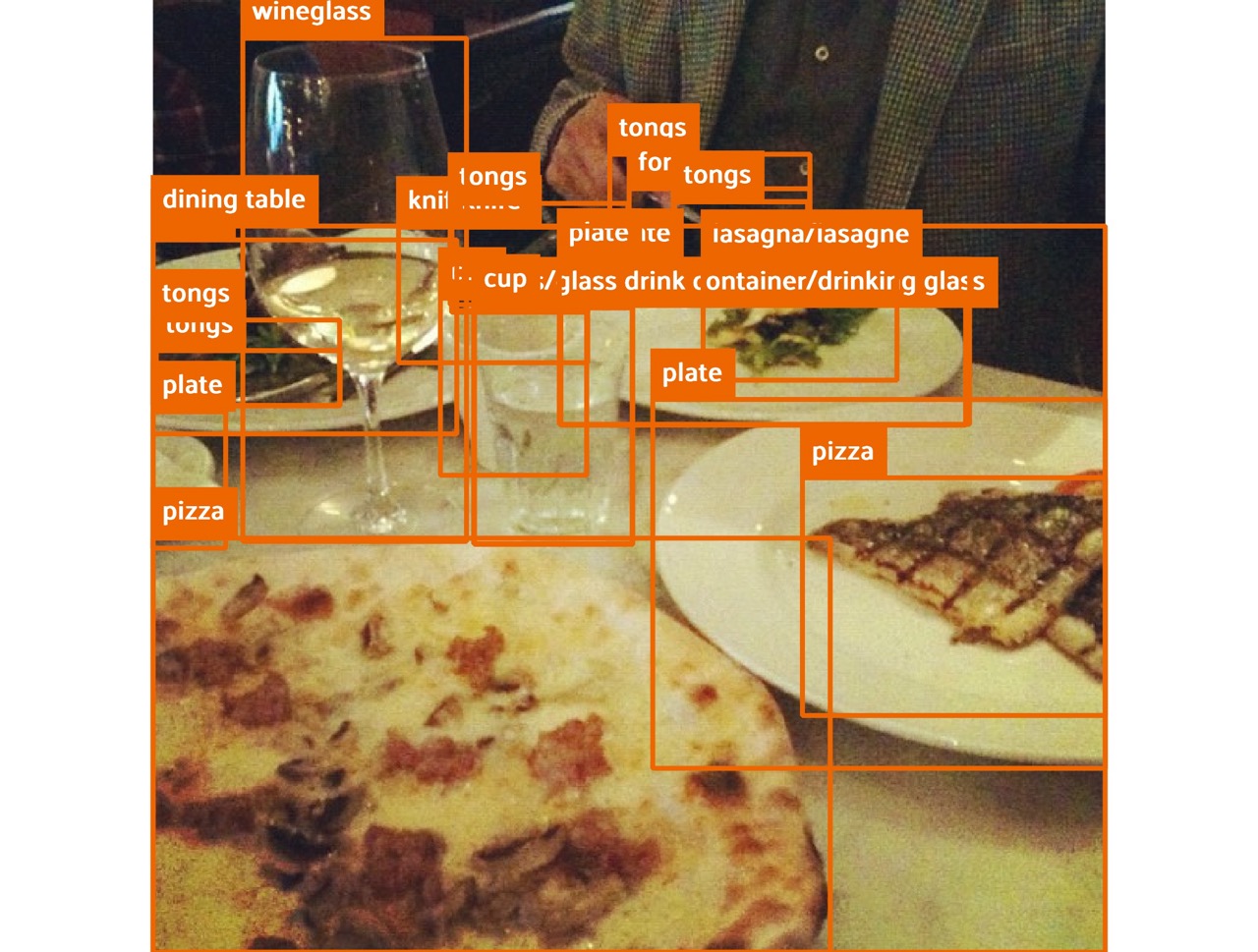}
	\end{minipage}%
	\begin{minipage}{0.25\textwidth}
		\includegraphics[width=0.975\textwidth,trim={5.4cm 0 5.4cm 0}, clip]{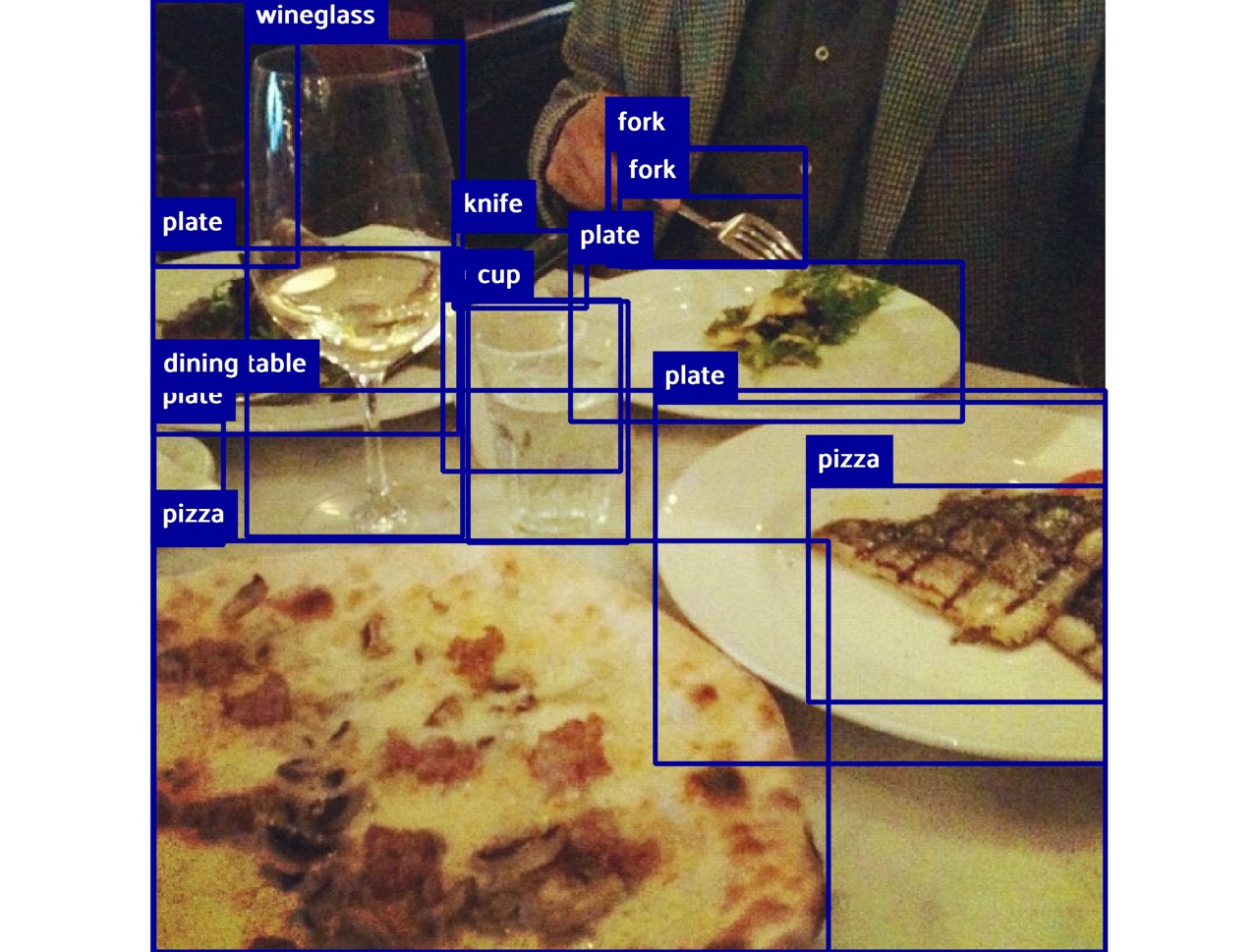}
	\end{minipage}%
	\begin{minipage}{0.25\textwidth}
		\includegraphics[width=0.975\textwidth,trim={5.4cm 0 5.4cm 0}, clip]{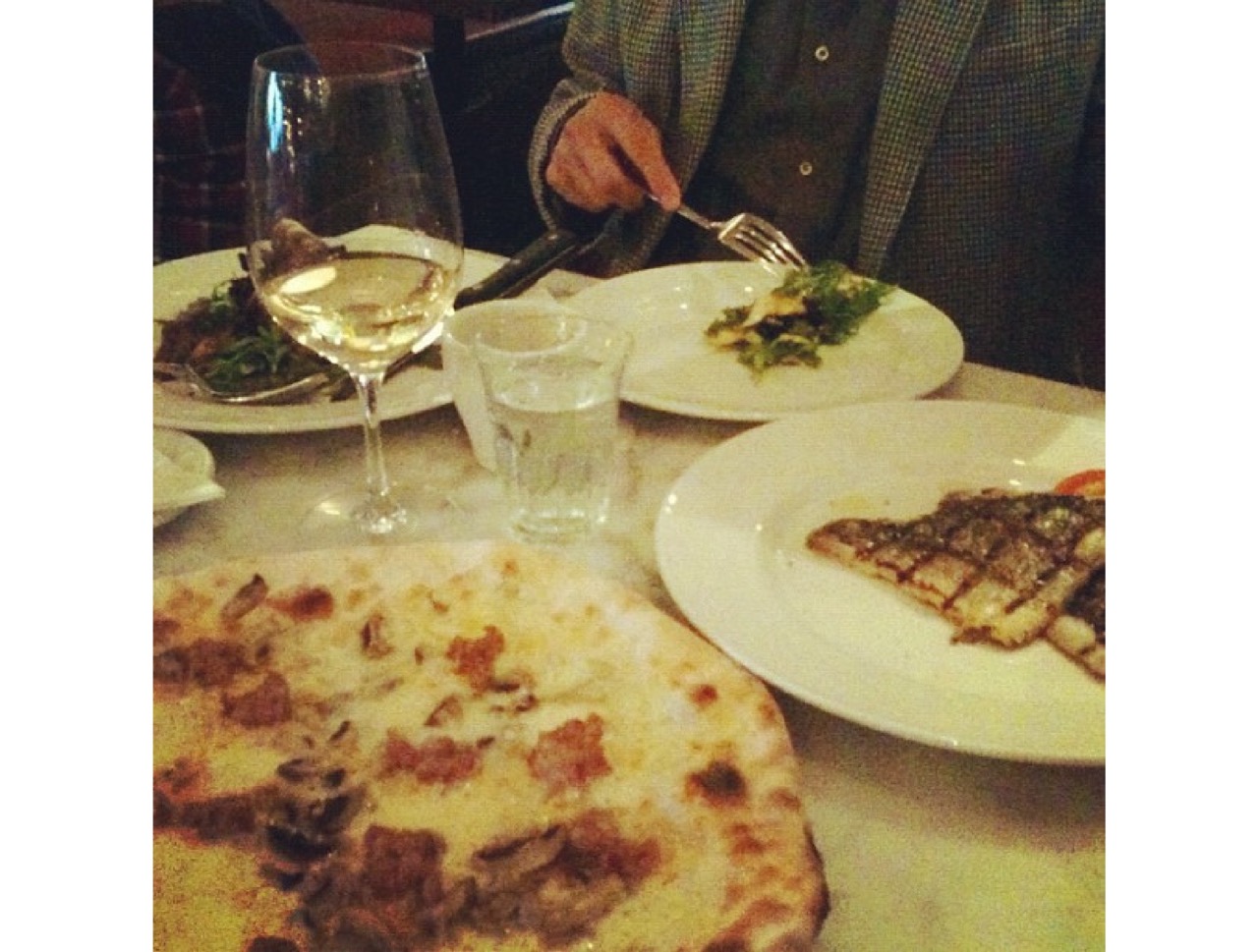}
	\end{minipage}%
	\\ 	\vspace{0.1em}
	\begin{minipage}{0.25\textwidth}
		\includegraphics[width=0.975\textwidth]{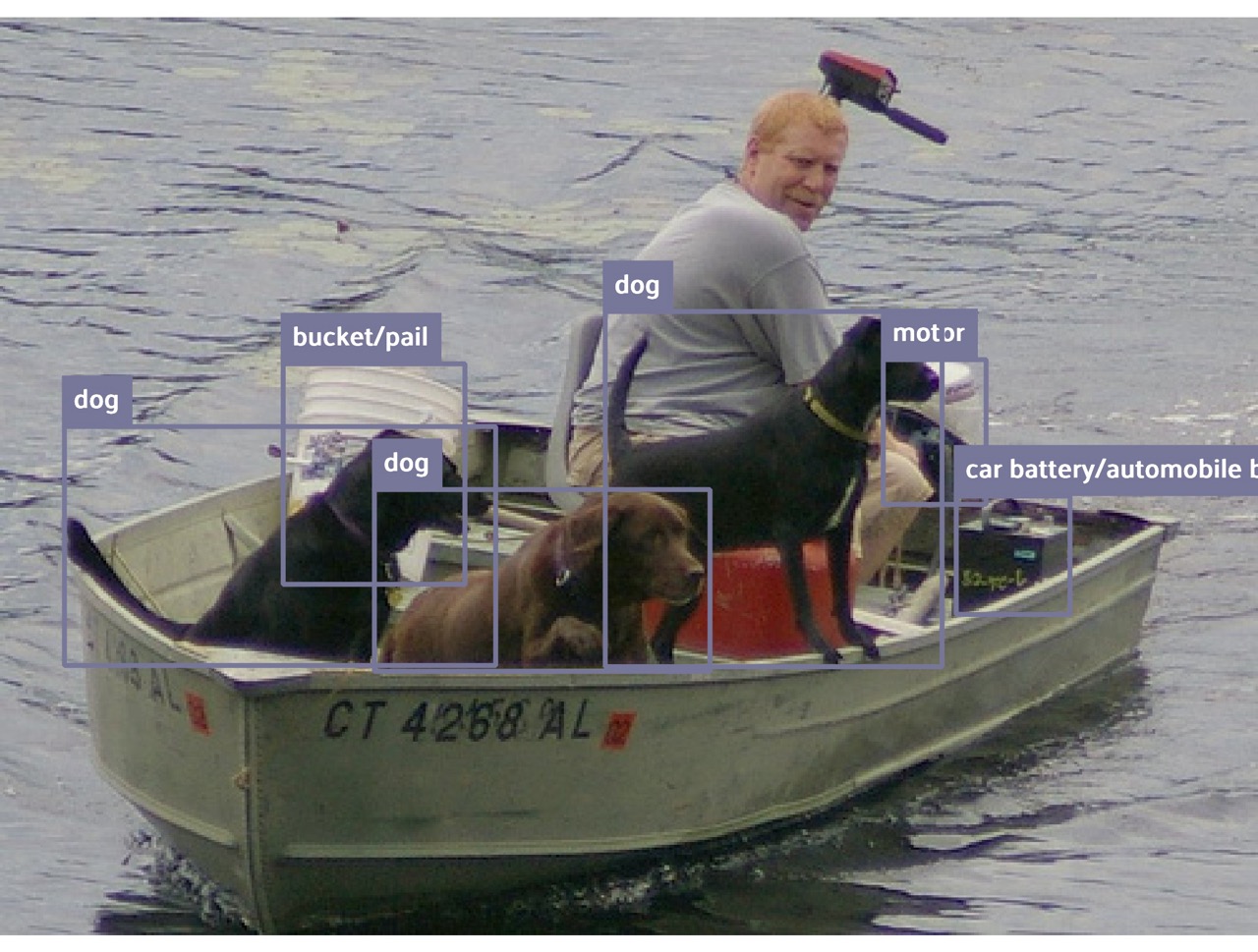}
	\end{minipage}%
	\begin{minipage}{0.25\textwidth}
		\includegraphics[width=0.975\textwidth]{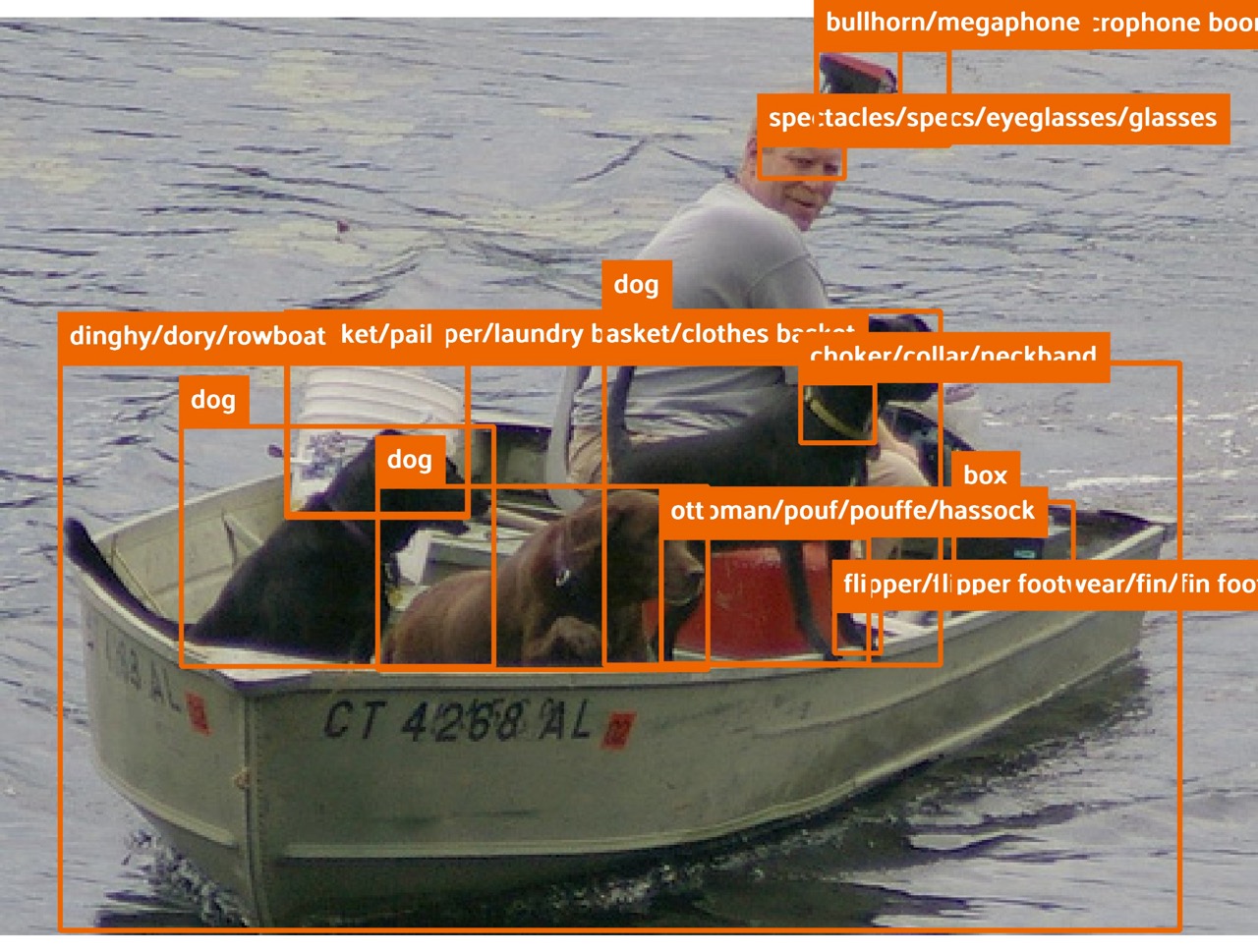}
	\end{minipage}%
	\begin{minipage}{0.25\textwidth}
		\includegraphics[width=0.975\textwidth]{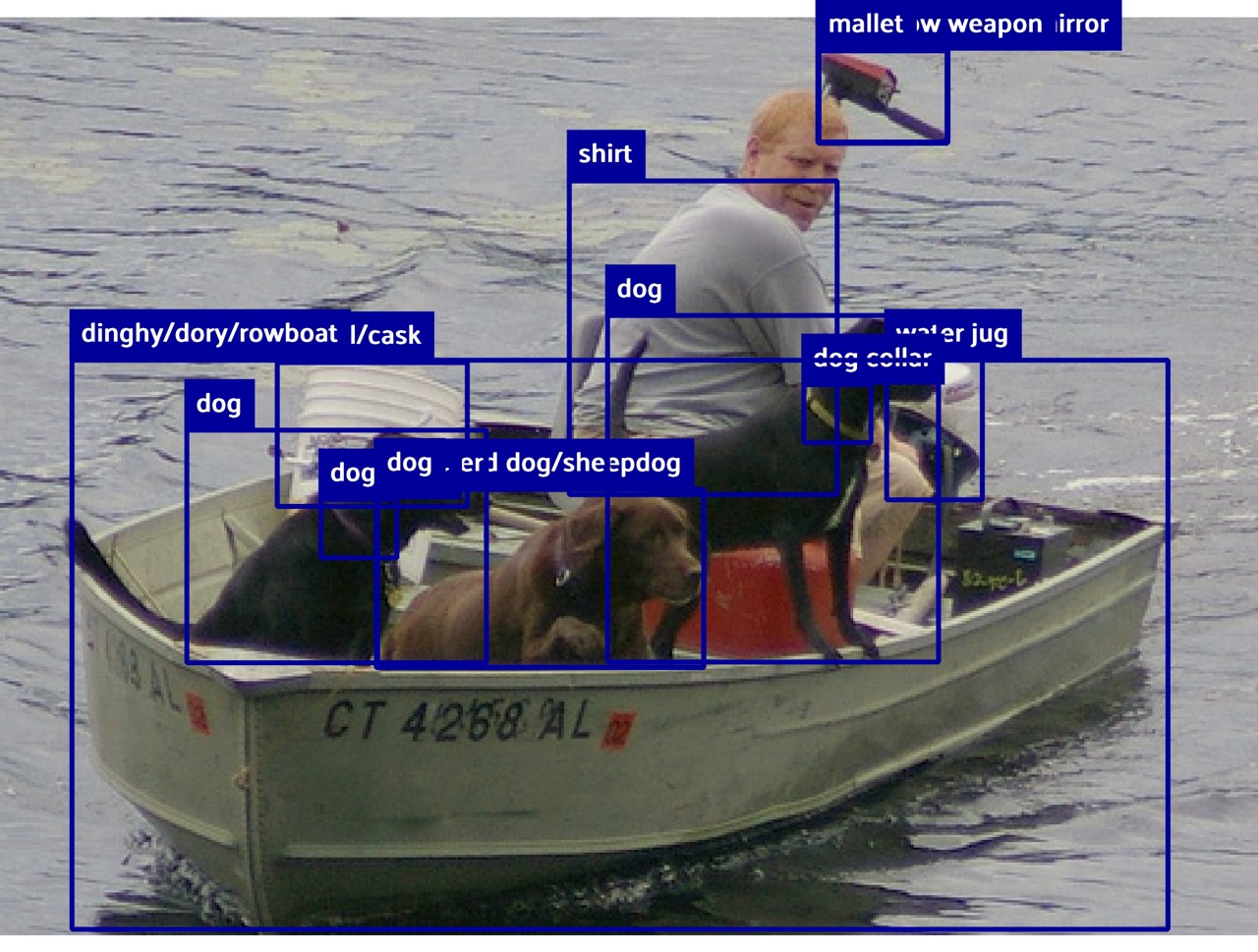}
	\end{minipage}%
	\begin{minipage}{0.25\textwidth}
		\includegraphics[width=0.975\textwidth]{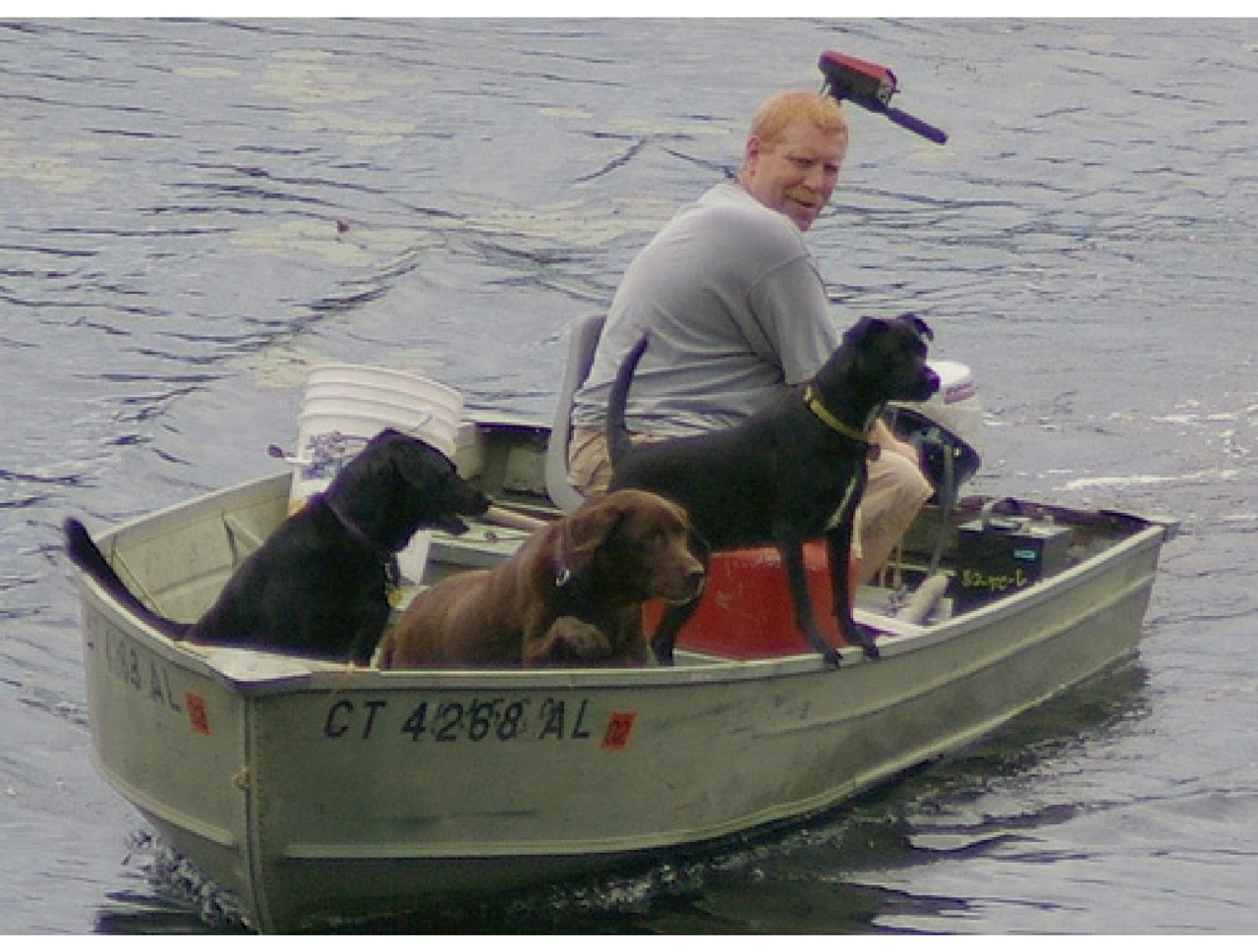}
	\end{minipage}%
	\\ 	\vspace{0.02em}
	\begin{minipage}{0.25\textwidth}
		\includegraphics[width=0.975\textwidth,trim={0 0 0 2.75cm}, clip]{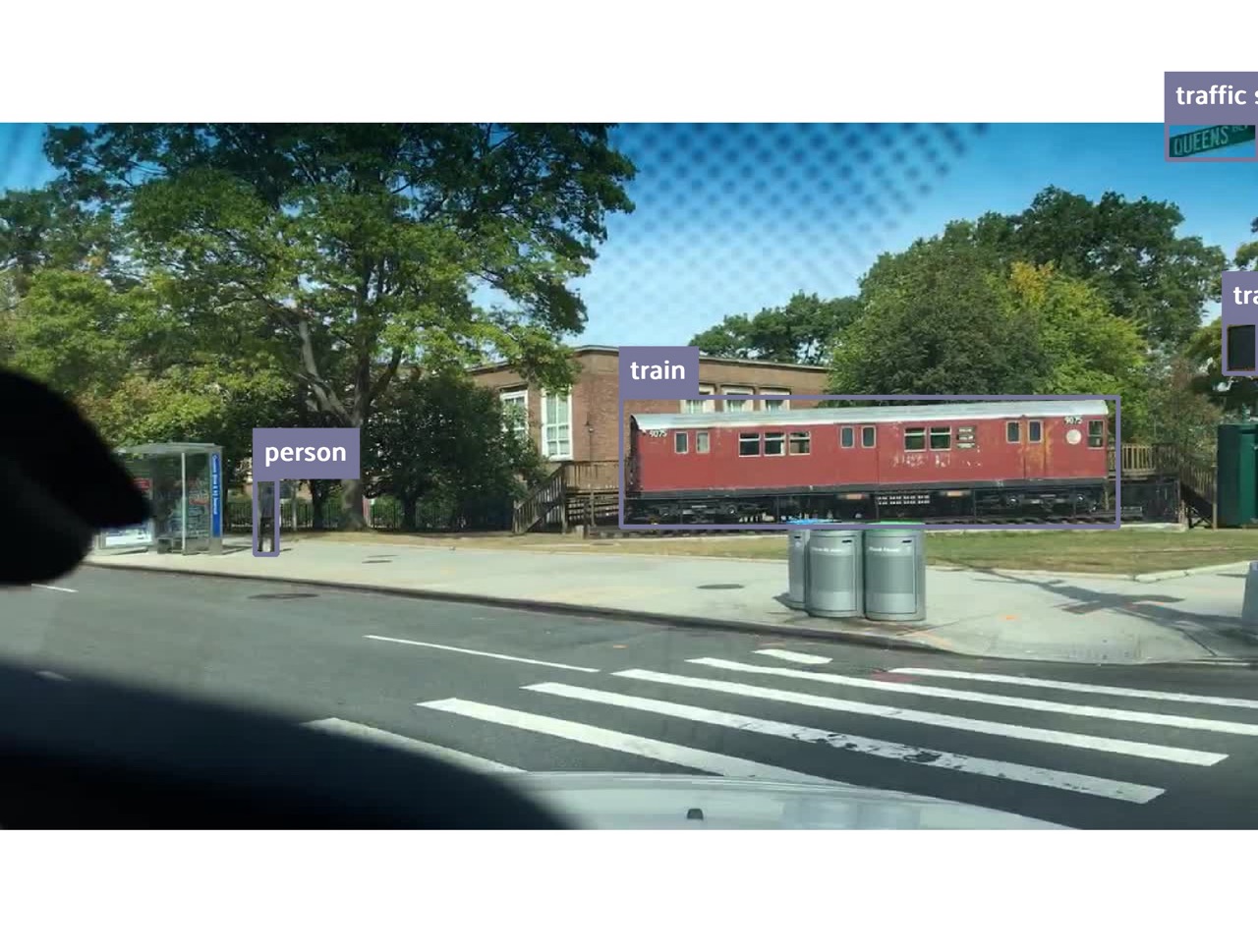}
	\end{minipage}%
	\begin{minipage}{0.25\textwidth}
		\includegraphics[width=0.975\textwidth,trim={0 0 0 2.75cm}, clip]{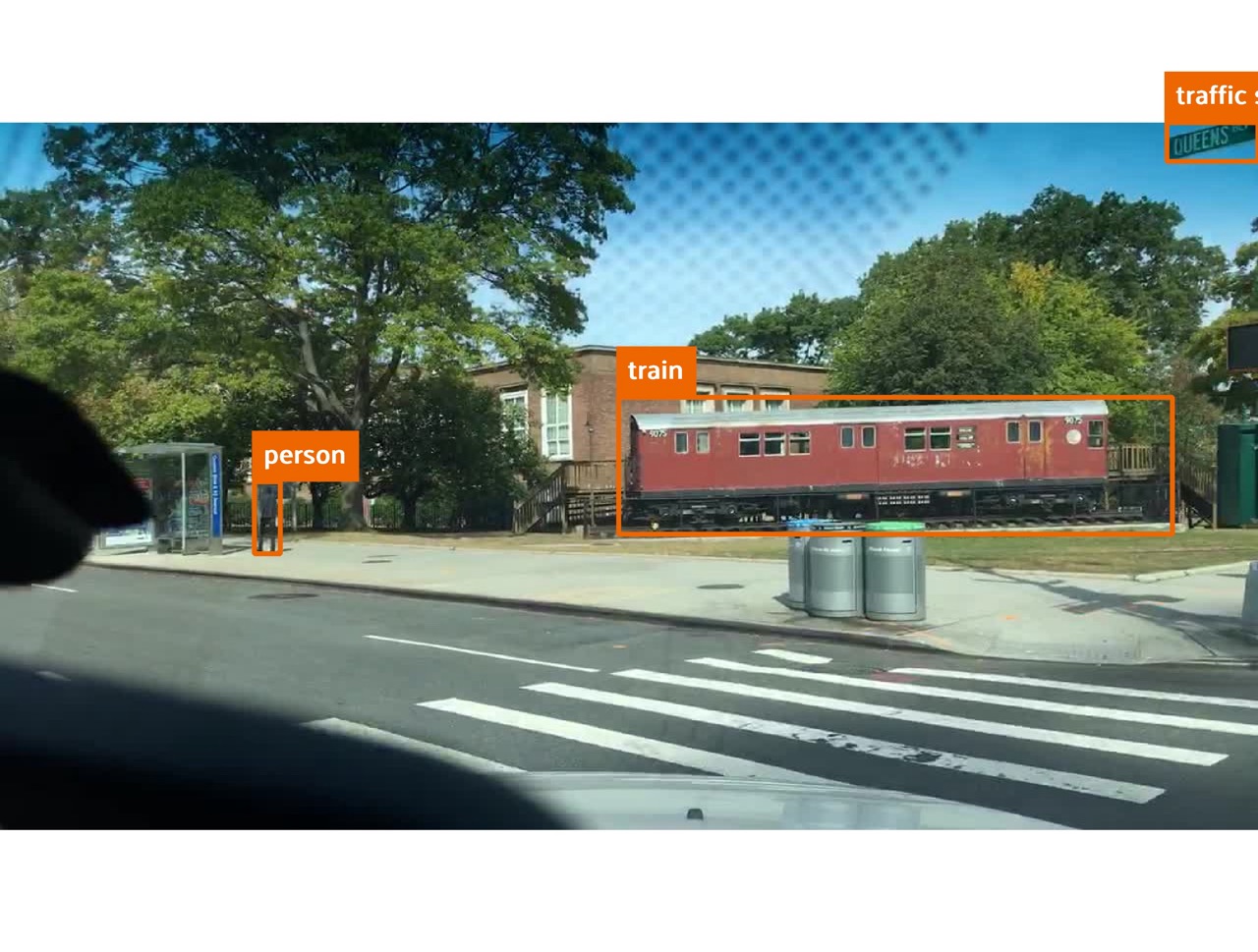}
	\end{minipage}%
	\begin{minipage}{0.25\textwidth}
		\includegraphics[width=0.975\textwidth,trim={0 0 0 2.75cm}, clip]{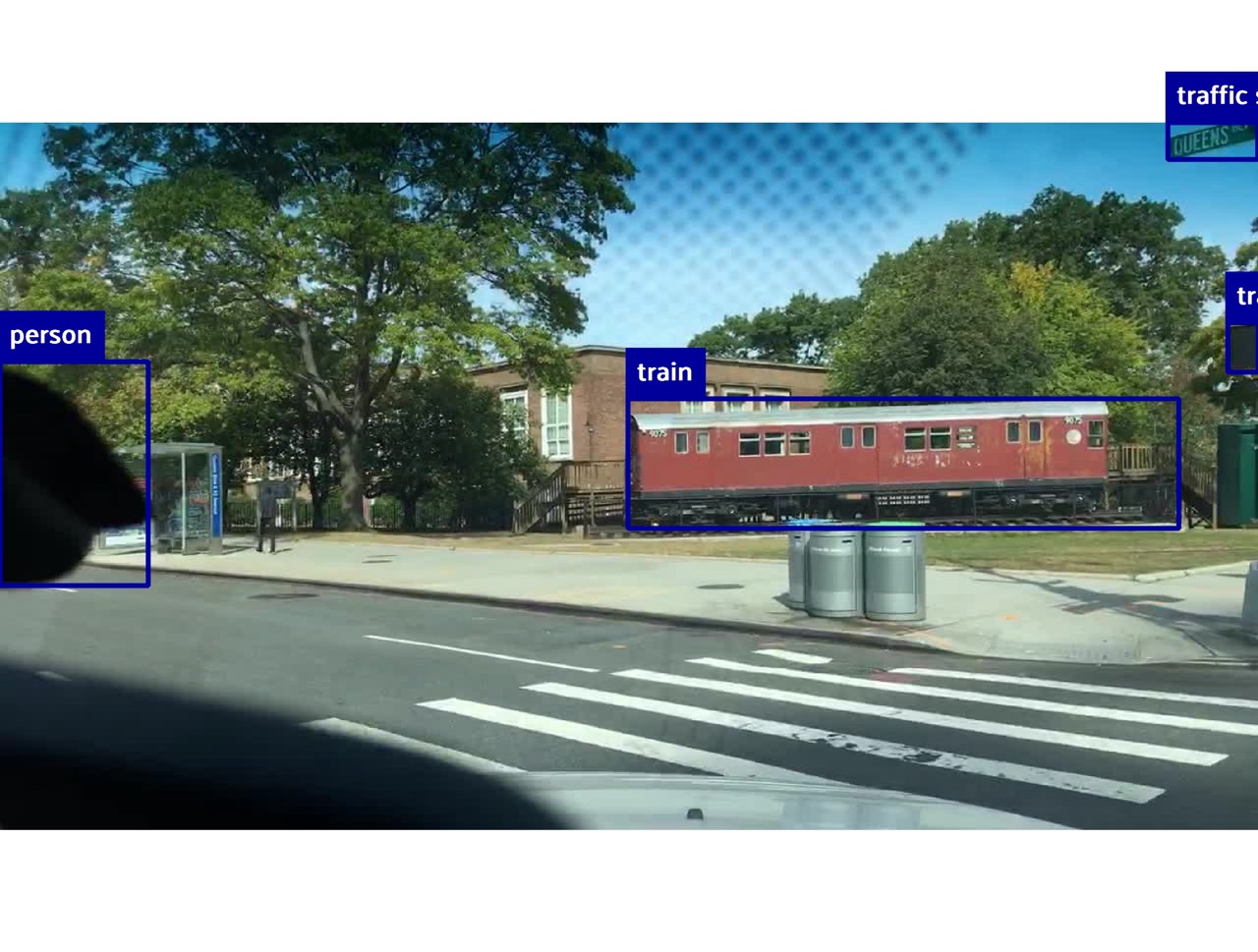}
	\end{minipage}%
	\begin{minipage}{0.25\textwidth}
		\includegraphics[width=0.975\textwidth,trim={0 0 0 2.75cm}, clip]{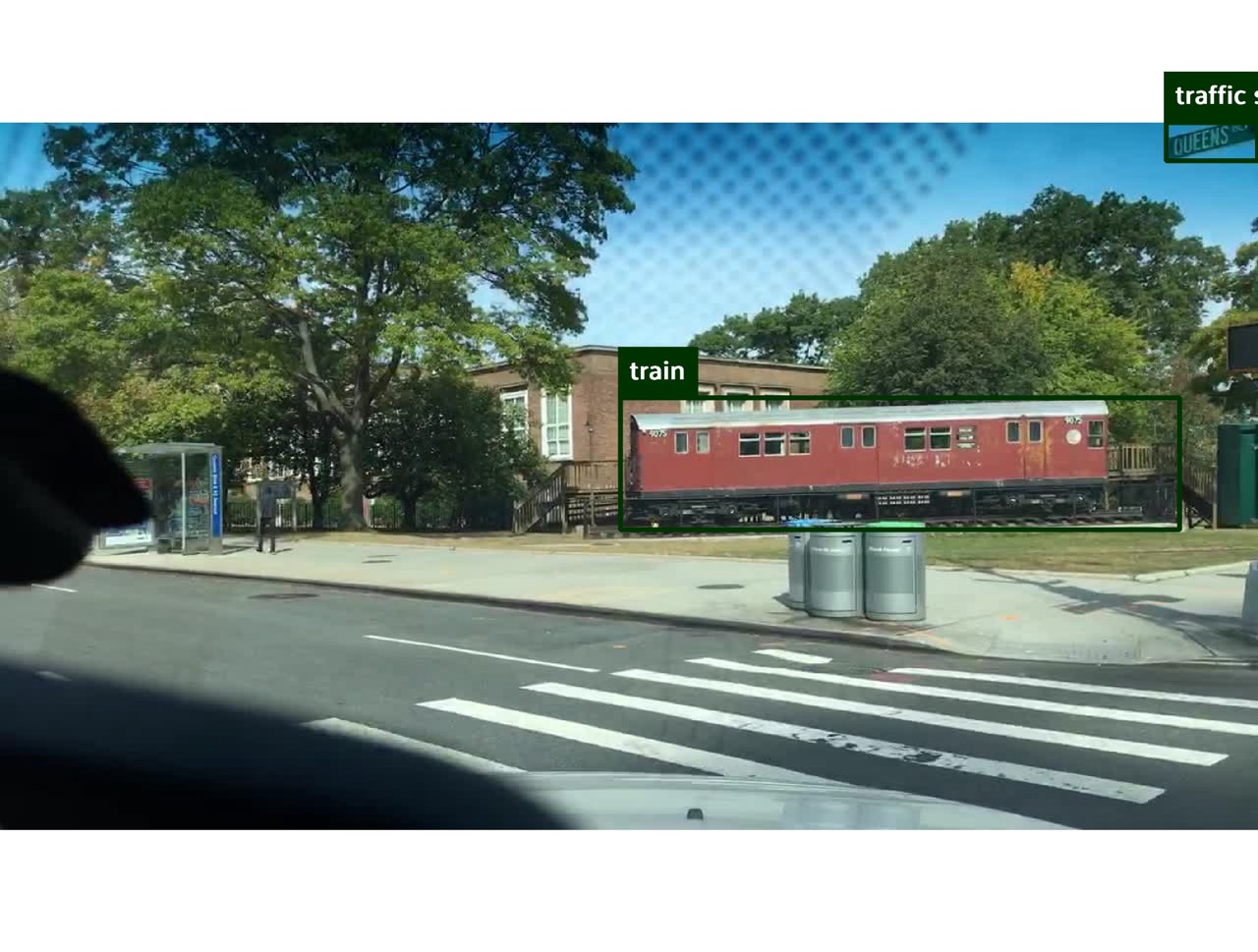}
	\end{minipage}%
	\\ 	\vspace{-0.9em}
	\begin{minipage}{0.25\textwidth}
		\includegraphics[width=0.975\textwidth,trim={0 0 0 8cm}, clip]{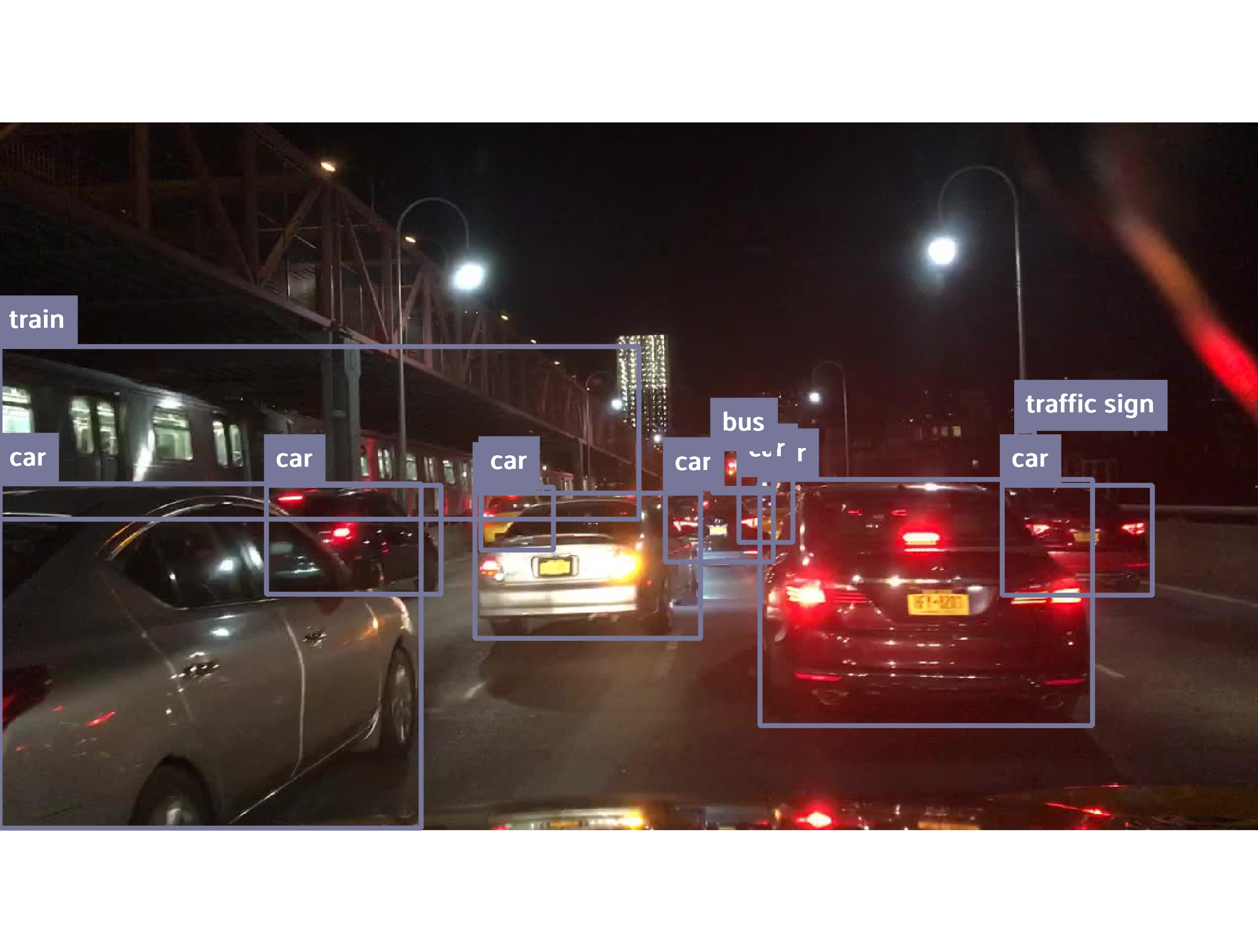}
	\end{minipage}%
	\begin{minipage}{0.25\textwidth}
		\includegraphics[width=0.975\textwidth,trim={0 0 0 8cm}, clip]{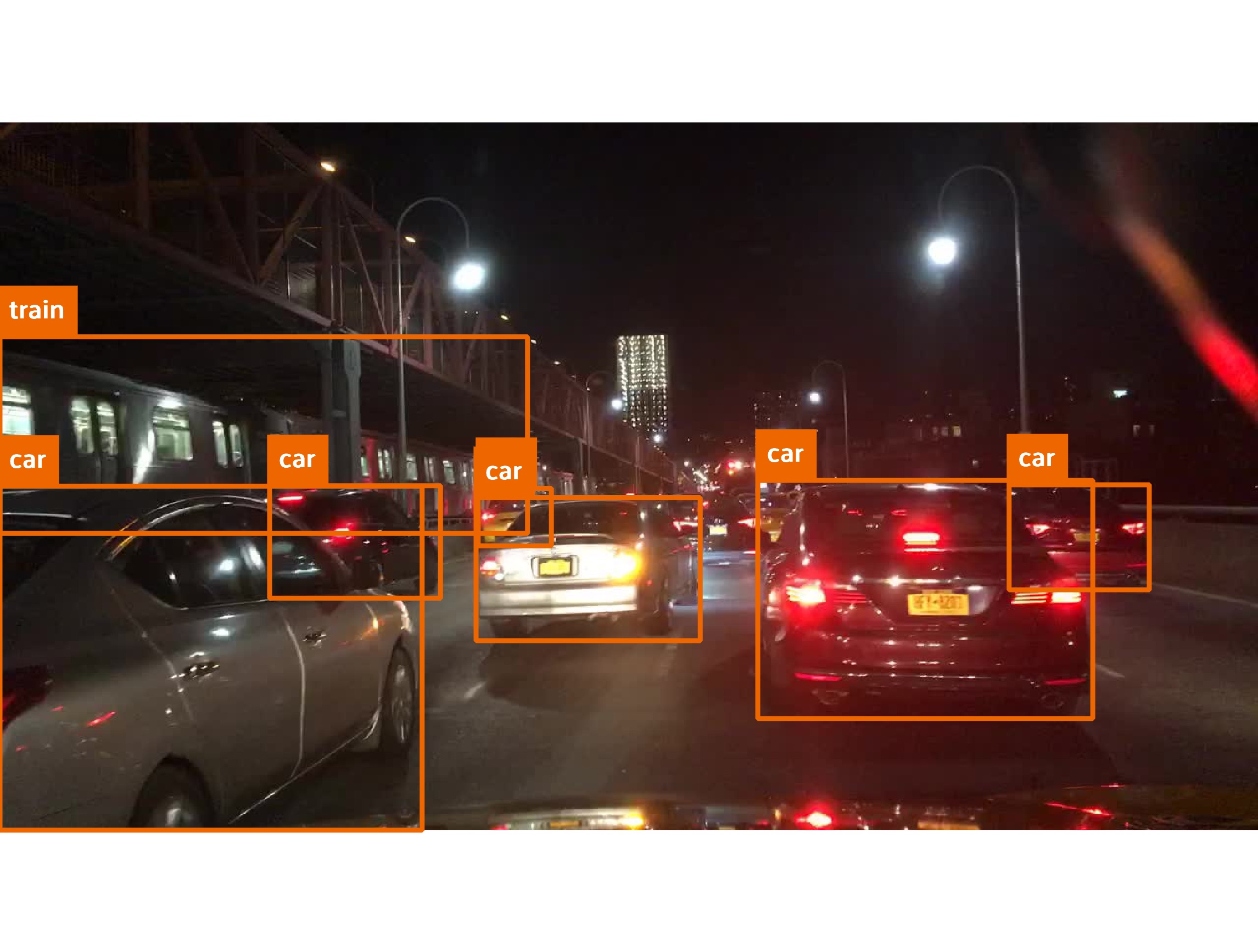}
	\end{minipage}%
	\begin{minipage}{0.25\textwidth}
		\includegraphics[width=0.975\textwidth,trim={0 0 0 8cm}, clip]{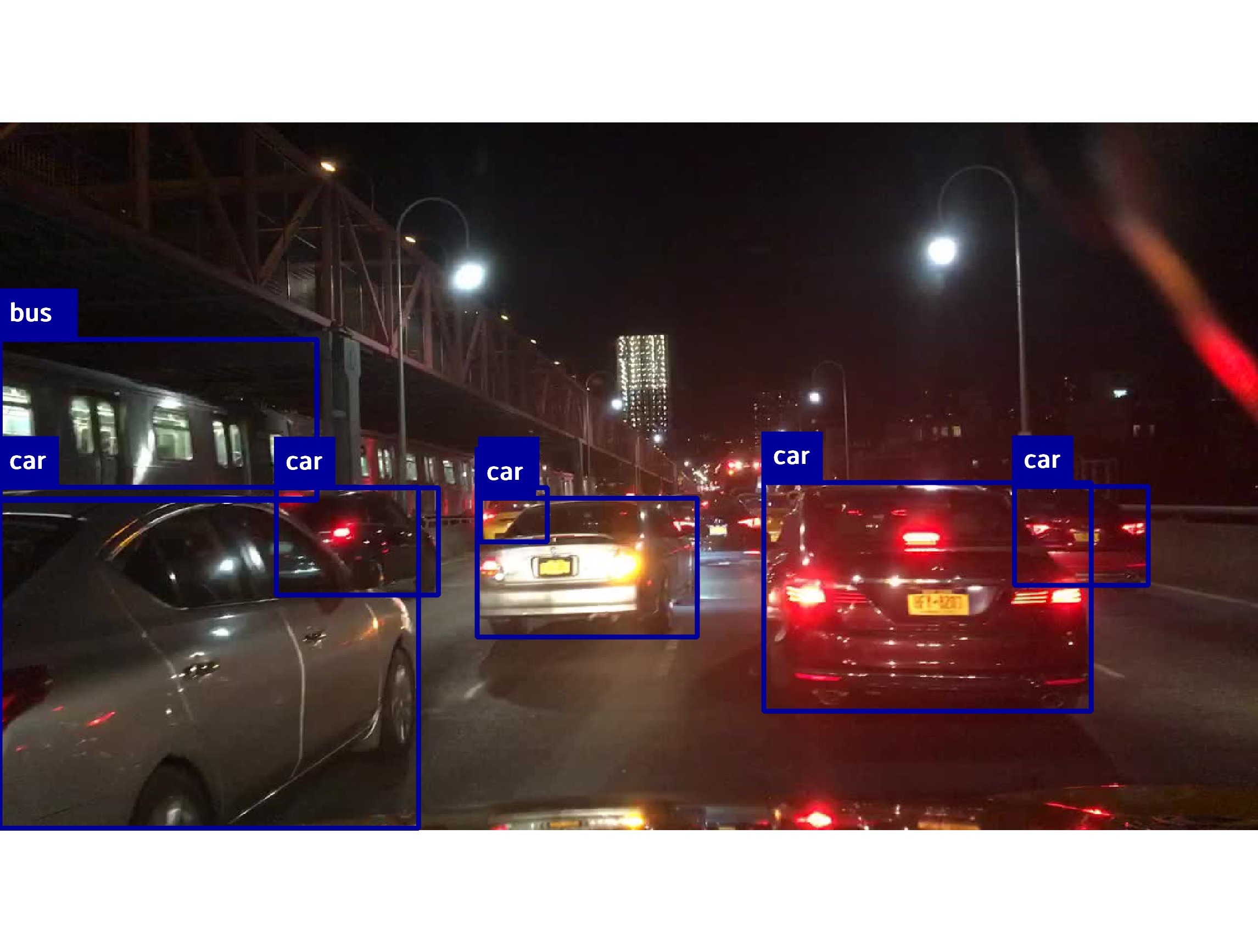}
	\end{minipage}%
	\begin{minipage}{0.25\textwidth}
		\includegraphics[width=0.975\textwidth,trim={0 0 0 8cm}, clip]{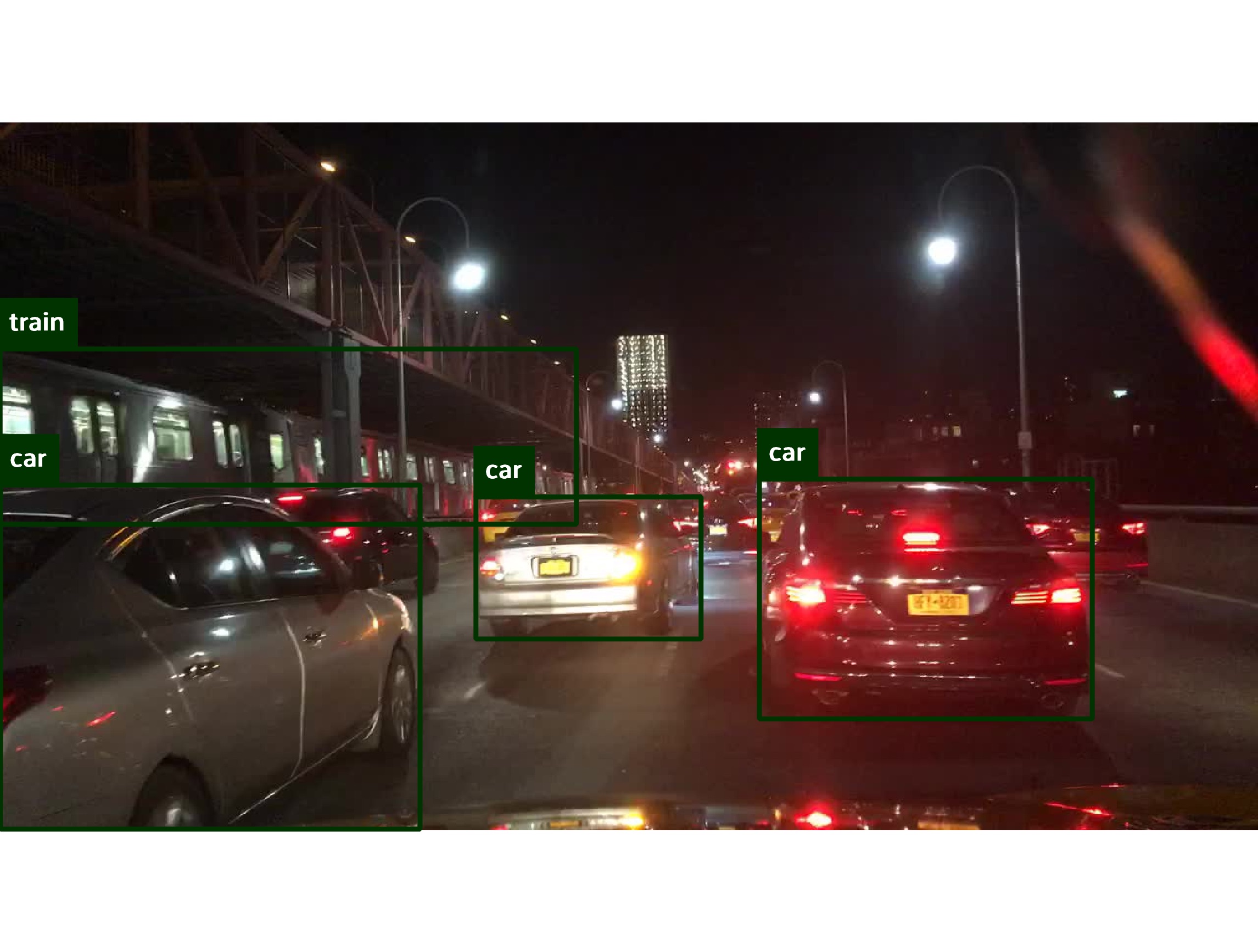}
	\end{minipage}%
	\vspace{-1.875em}
	\caption{{\bf Qualitative Comparison of Auto-Labels with Least Frequent Classes on LVIS \& BDD}. 
		All images include least frequent train set class for LVIS (steak knife \& car battery, top two rows) and BDD (train, rows 3-4). 
		We include two LVIS classes since each class has only one corresponding train image.
		Label sources are (left to right) human, YOLOW-0.2 (0.2 confidence threshold), YOLOE-0.2, and GDINO-0.5. 
		Due to architecture constraints, there are no GDINO labels on LVIS (see \cref{sec:data}), but we leave blank images for reference.
	}
	\label{fig:qualvis}
\end{figure*}

\subsection{Qualitative Auto-Labeling Evaluation}
We provide qualitative auto-labeling results across all datasets in \cref{fig:qualvoc}-\ref{fig:qualvis}.
All images include objects that are the least frequent class of their respective dataset (\cref{fig:class}), and we use a single confidence threshold setting ($\alpha$) for each AL model  (YOLOW-0.2, YOLOE-0.2, and GDINO-0.5).

From the {\bf VOC} examples, we find that auto-labeling includes a few objects missed by human labeling.
For example, the person behind the desk, the bottles attached to bicycles, and the tv monitor in \cref{fig:qualvoc}, top.
On the other hand, a few auto-labels have incorrect class labels, such as the dining table (row 2, column 2, $\alpha=0.2$) and cat (row 2, column 3, $\alpha=0.2$), but this is less common with a higher $\alpha$ setting (row 2, column 4, $\alpha=0.5$).
Nonetheless, our previous experiments show that setting a lower confidence threshold (increasing recall, even with such errors) measurably and consistently improves downstream model performance.

From the {\bf COCO} examples, we find that the infrequent hair drier class is accurately labeled by YOLOW, unlabeled by YOLOE, and labeled by GDINO when prominent but unlabeled when occurring in an abnormal outdoor context (\cref{fig:qualvoc}, rows 3-4).

From the {\bf LVIS} examples, which are labeled for 1,203 classes, AL includes objects that human labeling does not but with mixed accuracy.
For example, both AL models correctly add the pizza on a plate (\cref{fig:qualvis}, top) and the prominent rowboat (row 2).
On the other hand, YOLOW incorrectly labels a bullhorn, glasses, flipper, and ottoman (row 2, column 2) while YOLOE incorrectly labels a mallet and water jug (row 2, column 3).
For the very rare steak knife and car battery classes, AL models label the steak knife incorrectly as a knife (top) and label the car battery as a box or leave it unlabeled (row 2).
Notably, there are also examples where human labels are corrected by AL, such as the wine glass in the top row.
Given the number of mistakes with AL and human labels, we find that LVIS is by far the most challenging dataset to annotate.

From the {\bf BDD} examples, we find that AL is overall fairly accurate but leaves small or partially occluded objects in heavy traffic unlabeled.
In the daytime driving scene (\cref{fig:qualvis}, row 3), we find that YOLOW labels are more accurate than YOLOE, GDINO, \textit{or} human labels.
In the nighttime driving scene (row 4), the only inaccurate label is YOLOE labeling the train on the left as a bus, but AL leaves a few cars and a bus that are partially occluded by several rows of traffic unlabeled.
Given these unique challenges of the driving domain, we find that BDD is the second most challenging dataset to annotate.

\section{Conclusions}

We propose a general approach to auto-labeling data for object detection.
Specifically, we label data for a given application using previously-trained foundational models, which incorporate knowledge from massive amounts of data.
We then use the auto-labeled data to train lightweight inference models that are computationally efficient for practical application, e.g., real-time inference on deployed AV systems.
In this way, our approach trains detection models without conventional annotation, which is cost prohibitive at scale.

To establish best practices for auto-labeling, we conduct an exhaustive series of auto-labeling and downstream model training experiments across four unique datasets.
Notably, faulty configuration of auto-labeling degrades downstream model performance and a few of our findings are counter intuitive.
For example, the configurations with the highest precision relative to human labels result in the worst downstream model performance.
Furthermore, the models with the best performance do not necessarily have the closest resemblance to human labels.
Nonetheless, we find a single auto-labeling configuration that is reliable across a wide variety of applications (YOLOW-0.2, \cref{sec:bp}).

\setlength{\tabcolsep}{2.75pt}
\begin{table}
	\centering
	\caption{ {\bf Single Auto-Label Configuration vs. Human Labels}.
		YOLOW uses constant $\alpha=0.2$ confidence threshold.
		All mAP50 results use YOLO11n inference model for training and validation.
	}
	\vspace{-0.75em}
	\scriptsize
	\begin{tabular}{ |r | c cc c c |r r|}
		\hline
		\rowcolor{tableheader} \multicolumn{1}{| c |}{\bf Label} & \multicolumn{5}{ c |}{\bf Validation mAP50} & \multicolumn{2}{ c |}{\bf Total Label}  \\ 
		\rowcolor{tableheader} \multicolumn{1}{| c |}{\bf Source} & \multicolumn{1}{ c}{\bf VOC} & \multicolumn{1}{ c }{\bf COCO} & \multicolumn{1}{ c }{\bf LVIS} & \multicolumn{1}{ c }{\bf BDD} & \multicolumn{1}{ c |}{\bf Average} & \multicolumn{1}{ c }{\bf Cost} & \multicolumn{1}{ c |}{\bf Hours} \\ \hline
		Human	&	0.756	&	0.496	&	0.087	&	0.434	&	\bf 0.443	&	\$1,240,92.54 &	6,702.53	\\ \hline
		YOLOW-0.2	&	0.715	&	0.460	&	0.059	&	0.271	&	0.376	& \bf	\$1.18 &	\bf 1.27 	\\ \hline \hline
		\bf Difference	&	0.041	&	0.035	&	0.028	&	\bf 0.164	&	0.067	&	\$124,091.36 &	6,701.26	\\ \hline 
		\bf Ratio	&	1.057	&	1.077	&	1.484	&	\bf 1.605	&	1.178	&	105,105.49 &	5,279.60	\\ \hline
	\end{tabular}
	\label{tab:compare}
\end{table}

In regards to the viability of auto-labeling as a replacement for conventional labeling, we summarize a few key findings from our experiments in \cref{tab:compare}.
First, a lightweight YOLO11n model trained on YOLOW-0.2 labels achieves a mean average precision (mAP50) of 0.715, 0.460, 0.059, \& 0.271 on the VOC, COCO, LVIS, \& BDD validation sets respectively.
For comparison, the same model trained on standard labels achieves an mAP50 of 0.756, 0.496, 0.087, \& 0.434 on the same datasets.
Thus, auto-label-trained model performance is competitive on VOC \& COCO but less so on LVIS \& BDD.
However, our study also accounts for time and cost, and it is important to acknowledge that auto-labeling all the train sets takes 1.27 hours  and costs \$1.18 while human labeling takes 6,703 hours and using an annotation service currently costs \$124,092.54 (\cref{sec:cost}).
When jointly considering average performance and cost, the mAP50 per dollar spent is 0.319 for auto-labeling and 0.357$\times10^{-5}$ for standard annotation.

Given the competitive performance \textit{and} cost and time savings on VOC \& COCO, we find that auto-labeling is absolutely viable for these datasets and similar applications.
Furthermore, we show that if cost savings are redirected to accommodate a larger inference model, the net result is higher performance on these datasets (\cref{tab:alhl}).
On the other hand, for challenge applications closer to the LVIS \& BDD datasets, visual AI developers need to carefully consider the cost-performance trade offs.
Nonetheless, given its incredibly low cost, we believe auto-labeling data is the best starting point for most object detection applications.
Furthermore, our approach is broadly applicable, integrates directly with existing training frameworks, and will improve with future research advancements of foundation models.

{
	\small
	\bibliographystyle{ieeenat_fullname}
	\bibliography{refautolabel}
}

\end{document}